\documentclass[journal]{IEEEtran}

\usepackage[numbers]{natbib}
\usepackage{enumitem}
\usepackage{booktabs}
\usepackage{amsfonts}
\usepackage{amssymb}
\usepackage{amsmath}
\usepackage{mathtools}
\usepackage{algpseudocode, algorithm}
\usepackage{color}
\usepackage{multirow}
\usepackage{url}
\usepackage{graphicx}
\graphicspath{ {figures/} }

\DeclareMathOperator*{\argmin}{argmin}
\newtheorem{proposition}{Proposition}

\newtheorem{corollary}{Corollary}
\newtheorem{lemma}{Lemma}
\newtheorem{definition}{Definition}

\begin{document}
	
	\title{Geometric Interpretation of Running Nystr\"{o}m-Based Kernel Machines and Error Analysis}
	
	\author{Weida~Li,
		Mingxia~Liu,
		and~Daoqiang~Zhang
		
		\thanks{W. Li and D. Zhang are with the College of Computer Science and Technology, Nanjing University of Aeronautics and Astronautics, MIIT Key Laboratory of Pattern Analysis and Machine Intelligence, Nanjing 211106, China. M. Liu is with the Department of Radiology and BRIC, University of North Carolina at Chapel Hill, Chapel Hill, North Carolina 27599, USA.}
		\thanks{Corresponding authors: M.~Liu (mingxia\_liu@med.unc.edu) and D.~Zhang (dqzhang@nuaa.edu.cn).}
	}

	\if false
	\markboth{IEEE Transactions on Neural Networks and Learning Systems}%
	{W.~Li, M.~Liu, and D.~Zhang: IEEE Transactions on Neural Networks and Learning Systems}
	\fi
	
	\maketitle

	\begin{abstract}
		Recently, Nystr\"{o}m method has proved its prominence empirically and theoretically in speeding up the training of kernel machines while retaining satisfactory performances and accuracy. 
		So far, there are several different approaches proposed to exploit Nystr\"{o}m method in scaling up kernel machines.
		However, there is no comparative study over these approaches, and they were individually analyzed for specific types of kernel machines. 
		Therefore, it remains a question that the philosophy of which approach is more promising when it extends to other kernel machines.
		In this work, motivated by the column inclusion property of Gram matrices, we develop a new approach with a clear geometric interpretation for running Nystr\"{o}m-based kernel machines. 
		We show that the other two well-studied approaches can be equivalently transformed to be our proposed one. 
		Consequently, analysis established for the proposed approach also works for these two. 
		Particularly, our proposed approach makes it possible to develop approximation errors in a general setting.
		Besides, our analysis also manifests the relations among the aforementioned two approaches and another naive one.
		First, the analytical forms of the corresponding approximate solutions are only at odds with one term.
		Second, the naive approach can be implemented efficiently by sharing the same training procedure with others. 
		These analytical results lead to the conjecture that the naive approach can provide more accurate approximate solutions than the other two sophisticated approaches. 
		Since our analysis also offers ways for computing the accuracy of these approximate solutions, we run experiments with classification tasks to confirm our conjecture.
	\end{abstract}
	
	
	\IEEEpeerreviewmaketitle

	\section{Introduction}
	As a well theoretically developed realm in machine learning, kernel methods have already achieved certain success in a broad range of fields~\cite{steinwart2008support,shawe2004kernel}. 
	Specifically, the use of kernel functions allows for implicit non-linear transformations that map feature spaces into reproducing kernel Hilbert spaces (RKHSs), which makes kernel methods suitable for non-linear applications.
	However, memory and computation bottlenecks pop up when dealing with large-scale datasets. 
	To address this issue, much effort has been devoted to developing a variety of computationally efficient schemes \cite{fine2001efficient,rahimi2007random,le2013fastfood,avron2014subspace}.

	Among all competing scaling-up schemes, the Nystr\"{o}m method, first introduced to the machine learning community by \citet{williams2001using}, has demonstrated its efficiency in terms of memory and computation time. 
	So far, several different approaches that employ Nystr\"{o}m method have been proposed to scale up different types of kernel machines.
	Among various pioneering studies, \citet{williams2001using} suggested replacing the Gram matrix with the Nystr\"{o}m-based approximate one for kernel ridge regression (KRR). 
	Specifically, the training of KRR can be easily sped up by taking advantage of the low-rank decomposition of the approximate Gram matrix. 
	For convenience, this approach is named Gram matrix substitution approach (GSA) in this paper. 
	Inspired by an equivalent way to linearize kernel support vector machine (KSVM), \citet{lan2019scaling} proposed a low-rank linearization approach (LLA) that makes use of the low-rank structure of the Nystr\"{o}m-based approximate Gram matrix to linearize KSVM, after which efficient linear solvers can be utilized \cite{lin2008trust,keerthi2008sequential,hsieh2008dual}. 
	The idea of LLA was also adopted and studied in scaling up dictionary learning \cite{golts2016linearized}. 
	Besides, Nyst\"{o}m computational regularization (NCR) was developed for KRR that restricts feasible solutions to lie in the span of the selected landmarks in RKHS \cite{rudi2015less}. 
	Recently, NCR has also been extended to kernel principal component analysis (KPCA) \cite{sterge2020gain}. 
	However, previous studies generally analyze these three approaches separately, without considering their underlying relationships. 
	Therefore, it remains a question that the philosophy of which approach is more promising when it is used for other kernel machines.
	
	Even though approximation errors for each of these three approaches have been established, they are specific to certain types of kernel machines. 
	For instance, the prediction errors of GSA to KRR and KSVM were studied by \citet{cortes2010impact}. 
	LLA came with an approximation error for KSVM \cite{lan2019scaling}, but its established error is locally estimated from the best low-rank approximate solution rather than a non-approximate optimal one. 
	Generalization performance of NCR was developed specifically for KRR \cite{rudi2015less} and kernel classification \cite{jin2013improved} under carefully-imposed assumptions. 
	Therefore, a natural question is whether approximation error analysis can be done in a general setting for these approaches?

	In this paper, motivated by the column inclusion property of Gram matrices, we propose a subspace projection approach (SPA) for running Nystr\"{o}m-based kernel machines in general. 
	Unlike other studies that rely on RKHS~\cite{yang2012nystrom,jin2013improved,rudi2015less}, our analysis is based on the Hilbert space. 
	The main advantage of this simplification is its convenience in handling the geometry of data, which is instrumental in reaching conclusions of interest.   
	Specifically, aided by the setting of SPA, we first recast LLA into an equivalent optimization problem. 
	This equivalence quickly leads to the revelation that NCR is a specific case of LLA. 
	Thus, we will mainly focus on LLA.
	Then, we carefully study when SPA can serve as an alternative perspective for analyzing LLA.
	Our conclusion is that either a certain kind of sampling strategies used in the Nystr\"{o}m method or the representer theorem is enough to guarantee the equivalence between SPA and LLA. 
	One significant implication of it is that analysis developed for SPA also works for NCR and LLA. 
	In particular, we build up approximation errors (i.e., the accuracy of the computed approximate solutions) for SPA in a general setting.
	Moreover, the view of SPA also clearly demonstrates the relations between LLA (including NCR) and GSA.
	First, the analytical forms of the two computed approximate solutions only differ in one term.   
	Second, GSA can be implemented as efficiently as LLA (including NCR) by sharing the same training procedure. 
	Such an equivalent implementation for GSA does not add to computational cost.
	All these results lead to the conjecture that GSA can provide more accurate solutions than LLA (including NCR). 
	As provided by our analysis, the accuracy of the two corresponding approximate solutions can be exactly computed. 
	Therefore, we carry out experiments with classification tasks to support our conjecture.

	The contributions of this work can be summarized as:
	\begin{itemize}
		\item Our proposed SPA provides an alternative geometric interpretation for analyzing LLA. Meanwhile, we show that NCR is a specific case of LLA. In a nutshell, the mechanism behind LLA is that it projects all data in the new feature space before normally running kernel machines.
		
		\item We deduce an approximation error bound based on kernel machines in general for SPA (including LLA and NCR).
		
		\item The view of SPA reveals that the analytical forms of the computed approximate solutions from LLA and GSA only differ in one term. Also, GSA can be implemented as efficiently as LLA by sharing the same training procedure.
		
		\item Since our analytical framework provides ways for computing the accuracy of these approximate solutions, experiments with classification tasks are performed to verify our conjecture that GSA can provide more accurate solutions than LLA.
	\end{itemize}

	The rest of this paper is organized as follows. 
	Section~\ref{background} will review necessary requisite backgrounds. 
	In Section~\ref{Method}, we introduce our proposed SPA, study what it can provide for LLA, NCR and GSA, and also the sufficient conditions that lead to the equivalence between SPA and LLA.
	In Section~\ref{sec:experiments}, we perform experiments with classification tasks to support our conjecture about LLA (including NCR) and GSA. 
	Finally, this paper is concluded in Section~\ref{conclusion}.

	\section{Background}
	\label{background}
	\subsection{Notation}
	We focus on real Hilbert space $ \mathcal{H} $ with its endowed inner product $ \langle \cdot,\cdot \rangle_{\mathcal{H}} $, which serves as a new feature space when using a kernel function.
	In this paper, bold lower letters represent (column) vectors. 
	For instance, $ \mathbf{a} \in \mathbb{R}^{p} $ is a column vector and $ \mathbf{a}_{\mathcal{H}} $ is a vector in $ \mathcal{H} $.	
	Bold upper letters denote matrices or tuples of vectors. 
	For example, $ \mathbf{A}  $ is a matrix whereas $ \mathbf{A}_{\mathcal{H}} \in \mathcal{H}^{p} $ is a $ p $-tuple $ (\mathbf{a}_{\mathcal{H}}^{1},\mathbf{a}_{\mathcal{H}}^{2},\dots,\mathbf{a}_{\mathcal{H}}^{p}) $. 
	Non-bold letters are used to denote scalars or functions.  
	Given a matrix $ \mathbf{A} $, let $ \mathbf{a}_{i} $ be the $ i $-th column of $ \mathbf{A} $, $ \mathbf{A}^{\dagger} $ be its pseudo-inverse, and $ A_{ij} $ be its $ (i,j) $-th entry. 
	More definitions are listed in Table~\ref{tab:md}.
	Since $ (\mathbf{A}_{\mathcal{H}}\mathbf{B})\mathbf{C} = \mathbf{A}_{\mathcal{H}}(\mathbf{B}\mathbf{C}) $, writing $ \mathbf{A}_{\mathcal{H}}\mathbf{B}\mathbf{C} $ is without ambiguity. 
	Besides, it can be checked quickly that $ \langle \mathbf{Y}_{\mathcal{H}}\mathbf{A}, \mathbf{Z}_{\mathcal{H}}\mathbf{B} \rangle_{\mathcal{H}} = \mathbf{A}^{T}\langle \mathbf{Y}_{\mathcal{H}}, \mathbf{Z}_{\mathcal{H}} \rangle_{\mathcal{H}}\mathbf{B} $ and $ \langle \mathbf{Y}_{\mathcal{H}}, \mathbf{Z}_{\mathcal{H}} \rangle_{\mathcal{H}}^{T} = \langle \mathbf{Z}_{\mathcal{H}}, \mathbf{Y}_{\mathcal{H}} \rangle_{\mathcal{H}} $.
	
	\begin{table}
		\small
		\caption{Mathematical definitions used in this paper}
		\label{tab:md}
		\begin{center}
			\begin{small}
				\begin{tabular}{ll}
					\toprule
					Notation & Definition\\
					\midrule
					$ \Phi: \mathbb{R}^{d} \mapsto \mathcal{H} $ & A feature map\\
					$ \mathbf{Y}_{\mathcal{H}} = \mathbf{A}_{\mathcal{H}}\mathbf{B} $  & $ \mathbf{y}^{i}_{\mathcal{H}} = \sum_{j}b_{ji}\mathbf{a}^{j}_{\mathcal{H}} $ for all $ i $\\
					$ \mathbf{Z} = \langle \mathbf{A}_{\mathcal{H}}, \mathbf{B}_{\mathcal{H}} \rangle_{\mathcal{H}} $ & $ Z_{ij} = \langle \mathbf{a}^{i}_{\mathcal{H}}, \mathbf{b}^{j}_{\mathcal{H}} \rangle_{\mathcal{H}} $ for all $ i,j $ \\ $ \mathbf{Z}_{\mathcal{H}} =  \mathbf{A}_{\mathcal{H}}+\mathbf{B}_{\mathcal{H}} $ & $ \mathbf{z}^{i}_{\mathcal{H}} = \mathbf{a}^{i}_{\mathcal{H}} + \mathbf{b}^{i}_{\mathcal{H}} $ for all $ i $\\
					$ \mathbf{Y}_{\mathcal{H}} = \alpha\mathbf{A}_{\mathcal{H}} $ & $ \mathbf{y}^{i}_{\mathcal{H}} = \alpha\mathbf{a}^{i}_{\mathcal{H}} $ for all $ i $, and $ \alpha \in \mathbb{R} $ \\
					$ \mathrm{span}(\mathbf{A}_{\mathcal{H}}) $ & $ \{ \sum_{i}\alpha_{i}\mathbf{a}_{\mathcal{H}}^{i}\,:\, \alpha_{i} \in \mathbb{R} \} $ \\  $ \mathbf{A}_{\mathcal{H}} = \Phi(\mathbf{A}) $ & $ \mathbf{a}_{\mathcal{H}}^{i} = \Phi(\mathbf{a}_{i}) $ for all $ i $\\
					$ \|\mathbf{A}\|_{2} $ & Spectral norm of the matrix $ \mathbf{A} $\\
					$ \|\mathbf{A}\|_{*} $ & Trace norm of the matrix $ \mathbf{A} $\\
					$ \|\mathbf{A}\|_{F} $ & Frobenius norm of the matrix $ \mathbf{A} $\\
					$ \| \mathbf{A}_{\mathcal{H}} \|_{\mathcal{H}S}$ & Hilbert-Schmidt norm $ (\sum_{i}\| \mathbf{a}^{i}_{\mathcal{H}} \|_{\mathcal{H}}^{2})^{\frac{1}{2}} $\\
					$ \| \mathbf{A}_{\mathcal{H}} \|_{op} $ & Operator norm $ \sup_{\| \boldsymbol\alpha \|_{F}=1}\| \mathbf{A}_{\mathcal{H}}\boldsymbol\alpha \|_{\mathcal{H}} $\\
					\bottomrule
				\end{tabular}
			\end{small}
		\end{center}
	\end{table}	
	
	Let $ \mathbf{X} \in \mathbb{R}^{d\times n} $ denote a set of training data, where $ d $ and $ n $ refer to the number of features and data points, respectively. 
	$ \Phi:\mathbb{R}^{d} \mapsto \mathcal{H} $ denotes a feature map of a selected kernel function. 
	The advantages of using Hilbert space rather than RKHS will be demonstrated in our analysis. 
	Let $ \mathbf{X}_{\mathcal{H}} = \Phi(\mathbf{X}) \in \mathcal{H}^{n} $ and $ \mathbf{K} = \langle \mathbf{X}_{\mathcal{H}}, \mathbf{X}_{\mathcal{H}} \rangle_{\mathcal{H}} \in \mathbb{R}^{n\times n} $.

	\subsection{Kernel Machines}
	In this paper, we consider a general form of kernel machines as follows:
	\begin{equation} \label{op:kernel models}
		\begin{aligned}
			\argmin_{f\in\mathcal{H}} \hat{\mathcal{R}}(f,\mathbf{X}_{\mathcal{H}}) =& \cfrac{1}{n}\sum_{i=1}^{n}\mathcal{L}(\langle f, \mathbf{x}_{\mathcal{H}}^{i} \rangle_{\mathcal{H}}, y_{i}) + \Omega(\|f\|_{\mathcal{H}}^{2})\\
			\text{ subject to }& f \in \mathrm{span}(\mathbf{X}_{\mathcal{H}})
		\end{aligned}
	\end{equation} where $ \hat{\mathcal{R}} $ is an objective function, $ \mathcal{L} $ is a loss function, $ \mathbf{y} $ is a vector of labels, and $ \Omega : [0,+\infty] \mapsto [-\infty,+\infty] $ is a regularizing function. 
	If $ \mathcal{H} $ is assumed to be a reproducing kernel Hilbert space, then $ \langle f, \mathbf{x}_{\mathcal{H}}^{i} \rangle_{\mathcal{H}} = f(\mathbf{x}_{i}) $, and thus we treat $ f $ as a function rather than a vector in $ \mathcal{H} $. 
	The constraint $ f \in \mathrm{span}(\mathbf{X}_{\mathcal{H}}) $ mainly results from the representer theorem. 
	Note that the sufficient conditions leading to the representer theorem vary \cite{Schlkopf2001AGR,dinuzzo2012representer,yu2013characterizing}.

	The merit of this constraint is that it makes the problem~\eqref{op:kernel models} solvable. 
	That is, the problem~\eqref{op:kernel models} is equivalent to
	\begin{equation} \label{op:solvable models}
		\begin{gathered}
			\argmin_{\boldsymbol\alpha \in \mathbb{R}^{n}} \cfrac{1}{n}\sum_{i=1}^{n}\mathcal{L}(\boldsymbol\alpha^{T}\mathbf{k}_{i}, y_{i}) + \Omega(\boldsymbol\alpha^{T}\mathbf{K}\boldsymbol\alpha)
		\end{gathered}
	\end{equation}
	with $ f=\mathbf{X}_{\mathcal{H}}\boldsymbol\alpha $. 
	Henceforth, one can obtain an optimal solution to the problem~\eqref{op:kernel models} through optimizing the problem~\eqref{op:solvable models}.

	\subsection{Nystr\"{o}m Method}
	Without scalable techniques, in general, the running time for optimizing the problem~\eqref{op:solvable models} is $ \mathcal{O}(n^{3}) $, which is quite computationally expensive. 
	Fortunately, the Nystr\"{o}m method is able to reduce the running time significantly.
	The main idea of the Nystr\"{o}m method is to generate a small set of landmarks $ \mathbf{C}_{\mathcal{H}} \in \mathcal{H}^{m} $ ($ m \ll n) $ to efficiently ``represent'' the training data by, e.g., approximating $ \mathbf{K} $ by $ \widetilde{\mathbf{K}} $ which is cheaper to calculate. 
	So far, the sampling strategies for generating $ \mathbf{C}_{\mathcal{H}} $ have been extensively studied \cite{drineas2005nystrom,drineas2012fast,wang2019scalable}, and there is a wide range of choices \cite{gittens2016revisiting,oglic2017nystrom,pourkamali2018randomized,wang2019scalable}.
	On the other hand, there are already several well-studied Nystr\"{o}m methods for using $ \mathbf{C}_{\mathcal{H}} $ to obtain $ \widetilde{\mathbf{K}} $ \cite{li2014large,wang2013improving,lim2015double}. 
	As a recent advance, \citet{lim2018multi} proposed a multi-scale Nystr\"{o}m method that further shapes $ \mathbf{C}_{\mathcal{H}} $ into a multi-layer structure so that a good balance between approximation and running time can be achieved while increasing $ m $. 
	Generally, $ \widetilde{\mathbf{K}} $ admits the form $ \widetilde{\mathbf{K}} = \mathbf{K}_{nm}\mathbf{M}\mathbf{M}^{T}\mathbf{K}_{mn} $ where $ \mathbf{K}_{nm} = \langle \mathbf{X}_{\mathcal{H}}, \mathbf{C}_{\mathcal{H}} \rangle_{\mathcal{H}} $, $ \mathbf{K}_{mn} = \mathbf{K}_{nm}^{T} $, $ \mathbf{M} \in \mathbb{R}^{m\times s} $ is a method-dependent variable and $ s\leq m \ll n $.

	Note that the form $ \widetilde{\mathbf{K}} = \mathbf{G}^{T}\mathbf{G} $ with $ \mathbf{G} = \mathbf{M}^{T}\mathbf{K}_{mn} \in \mathbb{R}^{s\times n} $ and $ s\leq m\ll n $ is the key that enables both GSA and LLA to scale up kernel machines.
	For example, GSA can reduce the training time from $ \mathcal{O}(n^{3}) $ to $ O(nms) $ for KRR \cite{williams2001using}.


	\subsection{Gram Matrix Substitution Approach (GSA)}
	As suggested by \citet{williams2001using}, one way to make use of $ \mathbf{C}_{\mathcal{H}} $ is to merely replace $ \mathbf{K} $ by $ \widetilde{\mathbf{K}} = \mathbf{G}^{T}\mathbf{G} $ with $ \mathbf{G} = \mathbf{M}^{T}\mathbf{K}_{mn} $ in the problem~\eqref{op:solvable models}, leading to the following
	\begin{equation} \label{op:approximate_solvable models}
		\begin{gathered}
			\argmin_{\boldsymbol\alpha \in \mathbb{R}^{n}} \cfrac{1}{n}\sum_{i=1}^{n}\mathcal{L}(\boldsymbol\alpha^{T}\widetilde{\mathbf{k}}_{i}, y_{i}) + \Omega(\boldsymbol\alpha^{T}\widetilde{\mathbf{K}}\boldsymbol\alpha) .
		\end{gathered}
	\end{equation}
	Then, if $ \hat{\boldsymbol\alpha} $ is optimal to the problem~\eqref{op:approximate_solvable models}, $ f^{\mathrm{GSA}} = \mathbf{X}_{\mathcal{H}}\hat{\boldsymbol\alpha} $ is an approximate solution computed by using GSA.
	Notably, the optimization can be accelerated by taking advantage of the low-rank decomposition $ \widetilde{\mathbf{K}} = \mathbf{G}^{T}\mathbf{G} $.
	For example, for each data point $ \mathbf{z} \in \mathbb{R}^{d} $ and each scalar $ c > 0 $, $ \widetilde{\mathbf{K}}\mathbf{z} $ be implemented as $ \mathbf{G}^{T}(\mathbf{G}\mathbf{z}) $, or aided by the Woodbury formula, $ (\widetilde{\mathbf{K}}+c\mathbf{I})^{-1}\mathbf{z} $ be equally replaced by $ \frac{1}{c}(\mathbf{z}-\mathbf{G}^{T}(\mathbf{G}\mathbf{G}^{T}+c\mathbf{I})^{-1}(\mathbf{G}\mathbf{z})) $. 
	Note that the replacement provided by the Woodbury formula reduces the running time from $ \mathcal{O}(n^{3}) $ to $ \mathcal{O}(ns^{2}) $ where $ s\ll n $. 
	However, such an approach would be inconvenient to apply when the optimization procedure is given as a black box, which is often the case.

	\subsection{Low-rank Linearization Approach (LLA)} \label{sub:lla}
	Since the inspiration for LLA is an equivalent linearization of KSVM, it was merely studied and analyzed for KSVM when proposed. However, its mechanism can be described for kernel machines in general. 
	Specifically, the goal of the LLA approach is to find a finite-dimensional approximate feature map for $ \Phi:\mathbb{R}^{d}\mapsto \mathcal{H} $ by looking into the approximate Gram matrix $ \widetilde{\mathbf{K}} $. 
	Since $ \widetilde{\mathbf{K}} = \mathbf{G}^{T}\mathbf{G} $ with $ \mathbf{G} = \mathbf{M}^{T}\mathbf{K}_{mn} $, LLA treats $ \mathbf{G} $ as the sought-after mapped training data in the approximate feature space. 
	Specifically, let $ \mathbf{z} \in \mathbb{R}^{d} $ be a data point, the map $ \widetilde{\Phi}: \mathbf{z} \mapsto \mathbf{M}^{T}\langle \mathbf{C}_{\mathcal{H}}, \Phi(\mathbf{z}) \rangle_{\mathcal{H}} $ exactly maps  $ \mathbf{X} $ into $ \mathbf{G} $ in a column-by-column manner, and thus is considered as the desired approximate feature map. 
	In a nutshell, LLA first maps all data by using the finite-dimensional approximate feature map $ \widetilde{\Phi} $ so as to linearize the kernel-based optimization problem. Precisely,
	LLA is trying to solve a linearized version of the problem~\eqref{op:kernel models} as follows,
	\begin{equation} \label{op:solvableLLA}
		\begin{gathered}
			\argmin_{\mathbf{w} \in \mathbb{R}^{s}} \hat{\mathcal{R}}(\mathbf{w},\mathbf{G}) = \cfrac{1}{n}\sum_{i=1}^{n}\mathcal{L}(\mathbf{w}^{T}\mathbf{g}_{i}, y_{i}) + \Omega(\|\mathbf{w}\|_{F}^{2}) ,
		\end{gathered}
	\end{equation}
	which can be solved much more efficiently by using well-developed linear solvers since $ s\ll n $. 
	If $ \hat{\mathbf{w}} $ is an optimal solution to the problem~\eqref{op:solvableLLA}, then for each data point $ \mathbf{z} \in \mathbb{R}^{d} $, the corresponding prediction is $ \hat{\mathbf{w}}^{T}\mathbf{M}^{T}\langle \mathbf{C}_{\mathcal{H}}, \Phi(\mathbf{z}) \rangle_{\mathcal{H}} $. 
	Specifically, LLA suggests using standard Nystr\"{o}m method \cite{williams2001using} to form $ \mathbf{G} $, in which the approximate Gram matrix is $ \widetilde{\mathbf{K}}^{\mathrm{std}} = \mathbf{K}_{nm}\mathbf{K}_{mm}^{\dagger}\mathbf{K}_{mn} $ where $ \mathbf{K}_{mm} = \langle \mathbf{C}_{\mathcal{H}}, \mathbf{C}_{\mathcal{H}} \rangle_{\mathcal{H}} $. 
	
	

	\subsection{Nystr\"{o}m Computational Constraint (NCR)}
	As another approach, NCR aims to scale up kernel machines by replacing the constraint in the problem~\eqref{op:kernel models} by $ f \in \mathrm{span}(\mathbf{C}_{\mathcal{H}}) $ \cite{rudi2015less,jin2013improved,sterge2020gain}. 
	However, a noticeable drawback of this approach is that it can not be straightforwardly 
	generalized to other types of kernel machines, since each related study developed an analytical optimal solution exclusively for a certain kind of kernel machine.

	\section{Proposed Approach}
	\label{Method}
	\subsection{Motivation and Modeling}
	Our main inspiration is Observation 7.1.10 in \cite{horn2012matrix}, which leads to the following proposition. 
	We offer a more straightforward proof that provides a clear geometric interpretation. 
	\begin{proposition}[Column Inclusion Property of Gram Matrices] \label{prop:CIP}
		Let $ \mathbf{k} = \langle \mathbf{X}_{H}, \Phi(\mathbf{z}) \rangle_{\mathcal{H}} $, where $ \mathbf{z} \in \mathbb{R}^{d} $ is an unseen data point, then there exists $ \boldsymbol\beta \in \mathbb{R}^{n} $ such that $ \mathbf{k} = \mathbf{K}\boldsymbol\beta $. 
	\end{proposition}
	\emph{Proof.} 
	Let $ \mathbf{S}_{\mathcal{H}} $ be an orthogonal basis of $ \mathrm{span}(\mathbf{X}_{\mathcal{H}}) $, which can be obtained by performing Gram-Schmidt process on $ \mathbf{X}_{\mathcal{H}} $. Let $ \mathbf{z}_{\mathcal{H}} = \Phi(\mathbf{z}) $, then it holds that 
	\begin{equation}
		\begin{aligned}
			\langle \mathbf{X}_{\mathcal{H}}, \mathbf{z}_{\mathcal{H}} \rangle_{\mathcal{H}} =&\langle \mathbf{S}_{\mathcal{H}}\langle \mathbf{S}_{\mathcal{H}}, \mathbf{X}_{\mathcal{H}} \rangle_{\mathcal{H}}, \mathbf{z}_{\mathcal{H}} \rangle_{\mathcal{H}} \\
			=& \langle \mathbf{X}_{\mathcal{H}}, \mathbf{S}_{\mathcal{H}} \rangle_{\mathcal{H}}\langle \mathbf{S}_{\mathcal{H}}, \mathbf{z}_{\mathcal{H}} \rangle_{\mathcal{H}} \\
			=& \langle \mathbf{X}_{\mathcal{H}}, \mathbf{S}_{\mathcal{H}}\langle \mathbf{S}_{\mathcal{H}}, \mathbf{z}_{\mathcal{H}} \rangle_{\mathcal{H}} \rangle_{\mathcal{H}} .
		\end{aligned}
	\end{equation}
	Since $ \mathrm{span}(\mathbf{S}_{\mathcal{H}}) = \mathrm{span}(\mathbf{X}_{\mathcal{H}}) $, there exists $ \boldsymbol\beta \in \mathbb{R}^{n} $ such that $ \mathbf{X}_{\mathcal{H}}\boldsymbol\beta = \mathbf{S}_{\mathcal{H}}\langle \mathbf{S}_{\mathcal{H}}, \mathbf{z}_{\mathcal{H}} \rangle_{\mathcal{H}} $. Consequently, $ \mathbf{k} = \mathbf{K}\boldsymbol\beta $. \hfill $ \Box $

	Note that in our proof, $ \mathbf{S}_{\mathcal{H}}\langle \mathbf{S}_{\mathcal{H}}, \mathbf{z}_{\mathcal{H}} \rangle_{\mathcal{H}} $ simply projects the unseen data point $ \mathbf{z}_{\mathcal{H}} $ onto $ \mathrm{span}(\mathbf{X}_{\mathcal{H}}) $.
	In other words, the constraint $ f \in \mathrm{span}(\mathbf{X}_{\mathcal{H}}) $ implicitly projects all data, including the mapped training data $ \mathbf{X}_{\mathcal{H}} $ and each mapped unseen data point $ \mathbf{z}_{\mathcal{H}} $, onto a finite-dimensional subspace $ \mathrm{span}(\mathbf{X}_{\mathcal{H}}) $ before training and test.

	Motivated by Proposition~\ref{prop:CIP}, a natural way to make kernel machines scalable by using Nystr\"{o}m method is to use a set of landmarks $ \mathbf{C}_{\mathcal{H}} $ to first learn a meaningful orthogonal basis $ \mathbf{B}_{\mathcal{H}} \in \mathcal{H}^{s} $ (which means $ \langle \mathbf{B}_{\mathcal{H}}, \mathbf{B}_{\mathcal{H}} \rangle_{\mathcal{H}} = \mathbf{I} $) with the constraint $ \mathbf{B}_{\mathcal{H}} = \mathbf{C}_{\mathcal{H}}\mathbf{A} $. 
	Here, $ \mathbf{A} \in \mathbb{R}^{m\times s} $ is a learning variable and $ s \leq m $ denotes the dimension of the targeted subspace. 
	When a learned $ \mathbf{B}_{\mathcal{H}} $ is given, our proposed approach is to project all data onto $ \mathrm{span}(\mathbf{B}_{\mathcal{H}}) $ before training and test. 
	In this way, the projected training data will be $ \widetilde{\mathbf{X}}_{\mathcal{H}} = \mathbf{B}_{\mathcal{H}}\langle \mathbf{B}_{\mathcal{H}}, \mathbf{X}_{\mathcal{H}} \rangle_{\mathcal{H}} $, and the problem~\eqref{op:kernel models} becomes
	\begin{equation} \label{op:projected kernel models}
		\begin{aligned}
			\argmin_{f\in\mathcal{H}} \hat{\mathcal{R}}(f,\widetilde{\mathbf{X}}_{\mathcal{H}}) =& \cfrac{1}{n}\sum_{i=1}^{n}\mathcal{L}(\langle f, \widetilde{\mathbf{x}}_{\mathcal{H}}^{i} \rangle_{\mathcal{H}}, y_{i}) + \Omega(\|f\|_{\mathcal{H}}^{2})\\
			\text{ subject to }& f \in \mathrm{span}(\widetilde{\mathbf{X}}_{\mathcal{H}}),
		\end{aligned}
	\end{equation}
	which is called subspace projection approach (SPA) in this paper. 
	Note that the problem above is equivalent to the problem~\eqref{op:approximate_solvable models} with $ f = \widetilde{\mathbf{X}}_{\mathcal{H}}\boldsymbol\alpha $. 
	In the following, we will show how NCR and LLA can be equivalently transformed to be the problem~\eqref{op:projected kernel models}.

	It is worth mentioning that each optimal solution to the problem~\eqref{op:projected kernel models} will automatically project all unseen data onto $ \mathrm{span}(\widetilde{\mathbf{X}}_{\mathcal{H}}) $ before predicting.
	Specifically,
	given an unseen data point $ \mathbf{z} \in \mathbb{R}^{d} $ and let $ \mathbf{z}_{\mathcal{H}} = \Phi(\mathbf{z}) $, since $ \widetilde{\mathbf{X}}_{\mathcal{H}} = \mathbf{B}_{\mathcal{H}}\langle \mathbf{B}_{\mathcal{H}}, \widetilde{\mathbf{X}}_{\mathcal{H}} \rangle_{\mathcal{H}} $,
	there is
	\begin{equation}
		\begin{aligned}
			\langle \widetilde{\mathbf{X}}_{\mathcal{H}}, \mathbf{z}_{\mathcal{H}} \rangle_{\mathcal{H}} =& \langle \mathbf{B}_{\mathcal{H}}\langle \mathbf{B}_{\mathcal{H}}, \widetilde{\mathbf{X}}_{\mathcal{H}} \rangle_{\mathcal{H}}, \mathbf{z}_{\mathcal{H}} \rangle_{\mathcal{H}} \\
			=& \langle \widetilde{\mathbf{X}}_{\mathcal{H}}, \mathbf{B}_{\mathcal{H}} \rangle_{\mathcal{H}}\langle \mathbf{B}_{\mathcal{H}}, \mathbf{z}_{\mathcal{H}} \rangle_{\mathcal{H}} \\
			=& \langle \widetilde{\mathbf{X}}_{\mathcal{H}}, \mathbf{B}_{\mathcal{H}}\langle \mathbf{B}_{\mathcal{H}}, \mathbf{z}_{\mathcal{H}} \rangle_{\mathcal{H}} \rangle_{\mathcal{H}}.
		\end{aligned}
	\end{equation}
	In this equality, $ \mathbf{B}_{\mathcal{H}}\langle \mathbf{B}_{\mathcal{H}}, \mathbf{z}_{\mathcal{H}} \rangle_{\mathcal{H}} $ is the projection of $ \mathbf{z}_{\mathcal{H}} $ onto $ \mathrm{span}(\mathbf{B}_{\mathcal{H}}) $.
	So, there is no need to explicitly project unseen data onto $ \mathrm{span}(\mathbf{B}_{\mathcal{H}}) $.

	\subsection{Further Justification of the Use of $ \mathbf{B}_{\mathcal{H}} $} 
	\label{sub:justification}
	The use of orthogonal basis $ \mathbf{B}_{\mathcal{H}} $ is consistent with the aforementioned useful form $ \widetilde{\mathbf{K}} = \mathbf{K}_{nm}\mathbf{MM}^{T}\mathbf{K}_{mn} $.
	Precisely,
	given the projected training data $ \widetilde{\mathbf{X}}_{\mathcal{H}} $, the expected approximate Gram matrix is $ \widetilde{\mathbf{K}} = \langle \widetilde{\mathbf{X}}_{\mathcal{H}}, \widetilde{\mathbf{X}}_{\mathcal{H}} \rangle_{\mathcal{H}} = \langle \mathbf{X}_{\mathcal{H}},\mathbf{B}_{\mathcal{H}} \rangle_{\mathcal{H}}\langle \mathbf{B}_{\mathcal{H}}, \mathbf{X}_{\mathcal{H}} \rangle_{\mathcal{H}} = \mathbf{K}_{nm}\mathbf{AA}^{T}\mathbf{K}_{mn} $. 
	Note that $ \mathbf{M} $ is a method-dependent variable when selecting Nystr\"{o}m methods to form $ \widetilde{\mathbf{K}} $, whereas $ \mathbf{A} $ is a learning variable when searching a meaningful orthogonal basis $ \mathbf{B}_{\mathcal{H}} $ with the constraint $ \mathbf{B}_{\mathcal{H}} = \mathbf{C}_{\mathcal{H}}\mathbf{A} $. 
	Since all our results will be based on the use of $ \mathbf{B}_{\mathcal{H}} $, we take $ \mathbf{A} = \mathbf{M} $ in the following.
	Notably, it has been demonstrated that for standard Nystr\"{o}m method \cite{williams2001using}, one-shot Nystr\"{o}m method \cite{fowlkes2004spectral}, double-shot Nystr\"{o}m method \cite{lim2015double}, and multi-scale Nystr\"{o}m method \cite{lim2018multi}, the corresponding approximate Gram matrix $ \widetilde{\mathbf{K}} $ exactly admits the form $ \widetilde{\mathbf{K}} = \langle \mathbf{X}_{\mathcal{H}}, \mathbf{B}_{\mathcal{H}} \rangle_{\mathcal{H}}\langle \mathbf{B}_{\mathcal{H}}, \mathbf{X}_{\mathcal{H}} \rangle_{\mathcal{H}} $ where $ \mathbf{B}_{\mathcal{H}} = \mathbf{C}_{\mathcal{H}}\mathbf{A} $ and $ \langle \mathbf{B}_{\mathcal{H}},\mathbf{B}_{\mathcal{H}} \rangle_{\mathcal{H}} = \mathbf{I} $.
	More details can be found in \cite{lim2015double,lim2018multi}.

	Meanwhile, the use of $ \mathbf{B}_{\mathcal{H}} $ can be justified by its relations with how accurate the corresponding approximate Gram matrix $ \widetilde{\mathbf{K}} = \langle \widetilde{\mathbf{X}}_{\mathcal{H}}, \widetilde{\mathbf{X}}_{\mathcal{H}} \rangle_{\mathcal{H}} $ is, relations that are depicted by the following two lemmas. Particularly, they hold only with the assumption $ \langle \mathbf{B}_{\mathcal{H}}, \mathbf{B}_{\mathcal{H}} \rangle_{\mathcal{H}} = \mathbf{I} $.
	\begin{lemma} \label{lem:relation between K and B}
		\begin{gather}
			\| \mathbf{K} - \widetilde{\mathbf{K}} \|_{*} = \|\mathbf{X}_{\mathcal{H}}-\mathbf{B}_{\mathcal{H}}\langle \mathbf{B}_{\mathcal{H}}, \mathbf{X}_{\mathcal{H}} \rangle_{\mathcal{H}}\|_{\mathcal{H}S}^{2} , \label{eq:k=B1}\\  
			\| \mathbf{K} - \widetilde{\mathbf{K}} \|_{2} = \|\mathbf{X}_{\mathcal{H}}-\mathbf{B}_{\mathcal{H}}\langle \mathbf{B}_{\mathcal{H}}, \mathbf{X}_{\mathcal{H}} \rangle_{\mathcal{H}}\|_{op}^{2} . \label{eq:k=B2}
		\end{gather}
	\end{lemma}
	When the Hilbert space $ \mathcal{H} $ is assumed to be finite-dimensional, the operator norm $ \|\cdot\|_{op} $ and the Hilbert-Schmidt norm $ \|\cdot\|_{\mathcal{H}S} $ will reduce to the spectral norm $ \|\cdot\|_{2} $ and the Frobenius norm $ \|\cdot\|_{F} $, respectively. 
	Basically, Lemma~\ref{lem:relation between K and B} indicates that learning a meaningful orthogonal basis $ \mathbf{B}_{\mathcal{H}} $ can be exactly equivalent to searching a good approximation $ \widetilde{\mathbf{K}} $.

	\begin{lemma} \label{lem:reconstruction error}
		Given two data points $ \mathbf{p}, \mathbf{q} \in \mathbb{R}^{d} $, let $ \widetilde{\mathbf{p}}_{\mathcal{H}} = \mathbf{B}_{\mathcal{H}}\langle \mathbf{B}_{\mathcal{H}}, \Phi(\mathbf{p}) \rangle_{\mathcal{H}} $ and $ \widetilde{\mathbf{q}}_{\mathcal{H}} = \mathbf{B}_{\mathcal{H}}\langle \mathbf{B}_{\mathcal{H}}, \Phi(\mathbf{q}) \rangle_{\mathcal{H}} $, then the reconstruction error $ |\langle \widetilde{\mathbf{p}}_{\mathcal{H}}, \widetilde{\mathbf{q}}_{\mathcal{H}} \rangle_{\mathcal{H}} - \langle \Phi(\mathbf{p}), \Phi(\mathbf{q}) \rangle_{\mathcal{H}}| $ is $ 0 $ if either $ \Phi(\mathbf{p}) $ or $ \Phi(\mathbf{q}) $ belongs to $ \mathrm{span}(\mathbf{B}_{\mathcal{H}}) $.
	\end{lemma}
	
	This result generalizes Proposition 2 in the previous study \cite{zhang2010clustered}. 
	To be specific, the proposition there proves that if the set of landmarks $ \mathbf{C}_{\mathcal{H}} $ contains two training data points from $ \mathbf{X}_{\mathcal{H}} $, say $ \mathbf{x}_{\mathcal{H}}^{i} $ and $ \mathbf{x}_{\mathcal{H}}^{j} $,
	then $ \widetilde{K}_{ij} = K_{ij} $. 
	Here, $ \widetilde{\mathbf{K}} = \mathbf{K}_{nm}\mathbf{K}_{mm}^{\dagger}\mathbf{K}_{mn} $ where $ \mathbf{K}_{mm} = \langle \mathbf{C}_{\mathcal{H}}, \mathbf{C}_{\mathcal{H}} \rangle_{\mathcal{H}} $, which is the result of using standard Nystr\"{o}m method to form $ \widetilde{\mathbf{K}} $. 
	This fact is straightforward by using Lemma~\ref{lem:reconstruction error}.
	Note that the corresponding embedded orthogonal basis $ \mathbf{B}_{\mathcal{H}}^{\mathrm{std}} \in \mathcal{H}^{s} $ is the one spanning $ \mathrm{span}(\mathbf{C}_{\mathcal{H}}) $ \cite{lim2015double}.
	Particularly, $ \langle \widetilde{\mathbf{X}}_{\mathcal{H}}^{\mathrm{std}}, \widetilde{\mathbf{X}}_{\mathcal{H}}^{\mathrm{std}} \rangle_{\mathcal{H}} = \mathbf{K}_{nm}\mathbf{K}_{mm}^{\dagger}\mathbf{K}_{mn} $ where $ \mathbf{X}_{\mathcal{H}}^{\mathrm{std}} = \mathbf{B}_{\mathcal{H}}^{\mathrm{std}}\langle \mathbf{B}_{\mathcal{H}}^{\mathrm{std}}, \mathbf{X}_{\mathcal{H}} \rangle_{\mathcal{H}} $. 	
	If $ \mathbf{C}_{\mathcal{H}} $ contains $ \mathbf{x}_{\mathcal{H}}^{i} $ and $ \mathbf{x}_{\mathcal{H}}^{j} $, then the two data points belong to $ \mathrm{span}(\mathbf{B}_{\mathcal{H}}^{\mathrm{std}}) = \mathrm{span}(\mathbf{C}_{\mathcal{H}}) $. By Lemme~\ref{lem:reconstruction error}, there is $  \widetilde{K}_{ij} = \langle \widetilde{\mathbf{x}}_{\mathcal{H}}^{i}, \widetilde{\mathbf{x}}_{\mathcal{H}}^{j} \rangle_{\mathcal{H}} = \langle \mathbf{x}_{\mathcal{H}}^{i}, \mathbf{x}_{\mathcal{H}}^{j} \rangle_{\mathcal{H}} = K_{ij} $.	 
	After all, Lemma~\ref{lem:reconstruction error} suggests that the closer the training data $ \mathbf{X}_{\mathcal{H}} $ are to $ \mathrm{span}(\mathbf{B}_{\mathcal{H}}) $, the smaller the reconstruction errors will be.

	\subsection{NCR: A Specific Case of LLA}
	To explore the relationships among LLA, NCR and SPA, we introduce the following problem when a learned orthogonal basis $ \mathbf{B}_{\mathcal{H}} $ is provided, which will be shown is exactly an optimization problem of LLA that searches solutions directly in $ \mathcal{H} $.
	\begin{equation} \label{op:recastLLA}
		\begin{aligned}
			\argmin_{f\in\mathcal{H}} \hat{\mathcal{R}}(f, \mathbf{X}_{\mathcal{H}}) = & \cfrac{1}{n}\sum_{i=1}^{n}\mathcal{L}(\langle f, \mathbf{x}_{\mathcal{H}}^{i} \rangle_{\mathcal{H}}, y_{i}) + \Omega(\|f\|_{\mathcal{H}}^{2})\\
			\text{ subject to } & f \in \mathrm{span}(\mathbf{B}_{\mathcal{H}}).
		\end{aligned}
	\end{equation}
	The problem above is exactly equivalent to the problem~\eqref{op:solvableLLA} with $ f = \mathbf{B}_{\mathcal{H}}\mathbf{w} $ by noticing that $ \langle \mathbf{B}_{\mathcal{H}}, \mathbf{X}_{\mathcal{H}} \rangle_{\mathcal{H}} = \mathbf{A}^{T}\mathbf{K}_{mn} = \mathbf{G} $. 
	Note that $ \mathbf{A}=\mathbf{M} $ in our settings as stated in Subsection~\ref{sub:justification}.
	Therefore,
	if $ \hat{\mathbf{w}} $ is an optimal solution to the problem~\eqref{op:solvableLLA}, which is LLA, the solution $ \hat{f} = \mathbf{B}_{\mathcal{H}}\hat{\mathbf{w}} $ is optimal to the problem~\eqref{op:recastLLA}. Then, we verify that $ \hat{f} $ and LLA share the same prediction for each data point $ \mathbf{z} \in \mathbb{R}^{d} $. 
	As provided in Subsection~\ref{sub:lla}, the prediction from LLA is $ \hat{\mathbf{w}}^{T}\mathbf{A}^{T}\langle \mathbf{C}_{\mathcal{H}}, \Phi(\mathbf{z}) \rangle_{\mathcal{H}} $. 
	This result is exactly the same as the prediction $ \langle \hat{f},\Phi(\mathbf{z}) \rangle_{\mathcal{H}} = \hat{\mathbf{w}}^{T}\langle \mathbf{B}_{\mathcal{H}}, \Phi(\mathbf{z}) \rangle_{\mathcal{H}} $ when using $ \hat{f} $.
	Thus, we conclude that the problem~\eqref{op:recastLLA} is an alternative optimization problem for LLA.
	The significance of this result is that an approximate solution generated from LLA can be expressed as $ \mathbf{B}_{\mathcal{H}}\hat{\mathbf{w}} \in \mathcal{H} $. And it is the key to show that NCR is a specific case of LLA in the following.
	
	\begin{proposition} \label{prop:NCR<LLA}
		Suppose standard Nystr\"{o}m method is chosen to form $ \widetilde{\mathbf{K}} $, which indicates $ \mathrm{span}(\mathbf{B}_{\mathcal{H}}) = \mathrm{span}(\mathbf{C}_{\mathcal{H}}) $. Then, NCR is exactly LLA.
	\end{proposition}
	
	The above proposition is straightforward by noting that the problem~\eqref{op:recastLLA} becomes NCR when the constraint is replaced by $ \mathrm{span}(\mathbf{C}_{\mathcal{H}}) $. 
	Since $ \mathbf{B}_{\mathcal{H}} $ can vary for LLA, NCR is just a specific case of LLA.

	Particularly, Proposition~\ref{prop:NCR<LLA} connects two existing analytical optimal solutions for KRR. 
	To be precise, the loss function and regularizing function for KRR are $ \mathcal{L}^{\mathrm{KRR}}(\langle f, \mathbf{x}_{\mathcal{H}}^{i} \rangle_{\mathcal{H}},y_{i}) = (\langle f, \mathbf{x}_{\mathcal{H}}^{i} \rangle_{\mathcal{H}}-y_{i})^{2} $ and $ \Omega^{\mathrm{KRR}}(\| f \|_{\mathcal{H}}^{2}) = \lambda\| f \|_{\mathcal{H}}^{2}  $ where $ \lambda > 0 $, respectively. 
	As provided by \citet{rudi2015less}, the analytical optimal solution for NCR to KRR is
	\begin{equation}
		\begin{gathered}
			\mathbf{C}_{\mathcal{H}} (\mathbf{K}_{mn}\mathbf{K}_{nm}+\lambda_{0}\mathbf{K}_{mm})^{\dagger}\mathbf{K}_{mn}\mathbf{y}
		\end{gathered}
	\end{equation}
	where $ \mathbf{K}_{mm} = \langle \mathbf{C}_{\mathcal{H}}, \mathbf{C}_{\mathcal{H}} \rangle_{\mathcal{H}} $ and $ \lambda_{0} = n\lambda $. 
	By contrast, when using LLA with standard Nystr\"{o}m method, the optimal solution to the problem~\eqref{op:solvableLLA} in terms of KRR is $ \hat{\mathbf{w}}^{\mathrm{KRR}} = (\mathbf{G}\mathbf{G}^{T}+\lambda_{0}\mathbf{I})^{-1}\mathbf{G}\mathbf{y} $ where $ \mathbf{G} \gets (\mathbf{V}\boldsymbol\Sigma^{-1})^{T}\mathbf{K}_{mn} $. 
	Here, $ \mathbf{V} $ and $ \boldsymbol\Sigma $ are from the spectral decomposition $ \mathbf{K}_{mm} = \mathbf{V}\boldsymbol\Sigma^{2}\mathbf{V}^{T} $, where all diagonal entries in $ \boldsymbol\Sigma $ are positive.
	Meanwhile, the orthogonal basis induced by standard Nystr\"{o}m method is $ \mathbf{B}_{\mathcal{H}}^{\mathrm{std}} = \mathbf{C}_{\mathcal{H}}\mathbf{V}\boldsymbol\Sigma^{-1} $.
	Then, the optimal solution for LLA to KRR is
	\begin{equation}
		\begin{gathered}
			\mathbf{B}_{\mathcal{H}}^{\mathrm{std}}\hat{\mathbf{w}}^{\mathrm{KRR}}=\mathbf{B}_{\mathcal{H}}^{\mathrm{std}}(\mathbf{G}\mathbf{G}^{T}+\lambda_{0}\mathbf{I})^{-1}\mathbf{G}\mathbf{y} .
		\end{gathered}
	\end{equation}
	Since the optimal solution to the problem~\eqref{op:solvableLLA} in terms of linear ridge regression is unique, the equivalence between the problems~\eqref{op:solvableLLA} and~\eqref{op:recastLLA} immediately leads to the following corollary, which will be verified directly in Appendix~\ref{app:NCR2LLA}.
	\begin{corollary} \label{cor:NCR2LLA}
		For KRR, the two analytical optimal solutions generated by using NCR, and LLA with $ \mathbf{B}_{\mathcal{H}}^{\mathrm{std}} $ are exactly the same, i.e.,
		\begin{equation}
			\begin{gathered}
				\mathbf{C}_{\mathcal{H}} (\mathbf{K}_{mn}\mathbf{K}_{nm}+\lambda_{0}\mathbf{K}_{mm})^{\dagger}\mathbf{K}_{mn}\mathbf{y} = \mathbf{B}_{\mathcal{H}}^{\mathrm{std}}\hat{\mathbf{w}}^{\mathrm{KRR}} .
			\end{gathered}
		\end{equation}
	\end{corollary}

	\subsection{Implications from Equivalence between SPA and LLA}
	Our next question is whether SPA is equivalent to LLA. 
	If so, what will be provided by such an equivalence? 
	Before moving forward, it will be helpful to lay down the following definitions. 
	\begin{definition}
		LLA and SPA are said to be strongly equivalent if LLA and SPA share the same set of optimal solutions.
	\end{definition}
	\begin{definition}
		LLA and SPA are said to be weakly equivalent if whenever $ f \in \mathrm{span}(\mathbf{B}_{\mathcal{H}}) $ is optimal to LLA, the projection of $ f $ onto $ \mathrm{span}(\widetilde{\mathbf{X}}_{\mathcal{H}}) $ is optimal to both LLA and SPA.
	\end{definition}
	
	Note that the weak equivalence implies that each optimal solution to SPA is also optimal to LLA.
	Before studying the sufficient conditions for these two types of equivalence, we first assume the strong equivalence holds and explore what it can provide.
	Let a learned orthogonal basis $ \mathbf{B}_{\mathcal{H}} $ be given, and $ \hat{\mathbf{w}} $ be an optimal solution to the problem~\eqref{op:solvableLLA}. 
	Then, an immediate result is that there must exist an optimal solution $ \hat{\boldsymbol\alpha} $ to the problem~\eqref{op:approximate_solvable models} such that
	\begin{equation} \label{eq:xa=bw}
		\begin{gathered}
			\widetilde{\mathbf{X}}_{\mathcal{H}}\hat{\boldsymbol\alpha} = \mathbf{B}_{\mathcal{H}}\hat{\mathbf{w}} .
		\end{gathered}
	\end{equation}
	
	This equality indicates that the optimal solution sought by LLA, which is $ f^{\mathrm{LLA}} = \mathbf{B}_{\mathcal{H}}\hat{\mathbf{w}} $, can be equally expressed as $ f^{\mathrm{LLA}} = \widetilde{\mathbf{X}}_{\mathcal{H}}\hat{\boldsymbol\alpha} $, which is optimal to SPA~\eqref{op:projected kernel models}. 
	There are three significant messages conveyed by the equality~\eqref{eq:xa=bw}.

	\emph{First}, it provides a more convenient way for analyzing how accurate $ f^{\mathrm{LLA}} $ is.
	Precisely, let $ \widetilde{\boldsymbol\alpha} $ be an optimal solution to the problem~\eqref{op:solvable models}, then $ f^{*} = \mathbf{X}_{\mathcal{H}}\widetilde{\boldsymbol\alpha} $ is a non-approximate optimal solution to the kernel machine~\eqref{op:kernel models}. 
	An approximation error bound for LLA (including NCR) in a general setting can be easily developed.
	Also, the approximation error $ \| f^{\mathrm{LLA}}-f^{*} \|_{\mathcal{H}} $ can be explicitly computed. 
	These results are summarized in the following proposition.
	
	\begin{proposition} \label{prop:approximation error}
		Assume the strong equivalence between LLA and SPA holds. Then, for kernel machines in general, the approximation error for LLA (including NCR) satisfies
		\begin{gather}
			\left\lVert f^{\mathrm{LLA}}-f^{*} \right\rVert_{\mathcal{H}}^{2}  = \hat{\boldsymbol\alpha}^{T}\widetilde{\mathbf{K}}\hat{\boldsymbol\alpha} + \widetilde{\boldsymbol\alpha}^{T}\mathbf{K}\widetilde{\boldsymbol\alpha} - 2\hat{\boldsymbol\alpha}^{T}\widetilde{\mathbf{K}}\widetilde{\boldsymbol\alpha} ,\label{eq:LLA-f*}\\
			\|f^{\mathrm{LLA}}-f^{*}\|_{\mathcal{H}} \leq \|\mathbf{K}-\widetilde{\mathbf{K}}\|_{2}^{\frac{1}{2}}\|\widetilde{\boldsymbol\alpha}\|_{F} + \|\mathbf{K}\|_{2}^{\frac{1}{2}}\|\hat{\boldsymbol\alpha}-\widetilde{\boldsymbol\alpha}\|_{F} . \label{ieq:LLA}
		\end{gather}
	\end{proposition}
	
	The approximation error bound~\eqref{ieq:LLA} shows that the accuracy of the approximate solutions computed through LLA is mainly determined by 1) the Gram matrix approximation error $ \| \mathbf{K} - \widetilde{\mathbf{K}} \|_{2}^{\frac{1}{2}} $ and 2) the continuity of the kernel machine $ \| \hat{\boldsymbol\alpha} - \widetilde{\boldsymbol\alpha} \|_{F} $, which is machine-dependent. Particularly, \citet{cortes2010impact} has proved that for KRR, $ \|\hat{\boldsymbol\alpha}-\widetilde{\boldsymbol\alpha}\|_{F} \leq \mathcal{O}(\|\mathbf{K}-\widetilde{\mathbf{K}}\|_{2}) $; for KSVM, $ \|\hat{\boldsymbol\alpha}-\widetilde{\boldsymbol\alpha}\|_{F} \leq \mathcal{O}(\|\mathbf{K}-\widetilde{\mathbf{K}}\|_{2}^{\frac{1}{4}}) $. 
	Combining these results, we have the following corollary.
	
	\begin{corollary}
		Suppose the strong equivalence between LLA and SPA holds, there is
		\begin{gather}
			\| f^{\mathrm{LLA}}-f^{*} \|_{\mathcal{H}} \leq \mathcal{O}(\| \mathbf{K}-\widetilde{\mathbf{K}} \|_{2}^{\frac{1}{2}}) \text{ for KRR, }\\
			\| f^{\mathrm{LLA}}-f^{*} \|_{\mathcal{H}} \leq \mathcal{O}(\| \mathbf{K}-\widetilde{\mathbf{K}} \|_{2}^{\frac{1}{4}}) \text{ for KSVM. } \label{ieq:ksvm}
		\end{gather}
	\end{corollary}

	\emph{Second}, the difference between LLA and GSA is clear. Note that $ f^{\mathrm{GSA}} = \mathbf{X}_{\mathcal{H}}\hat{\boldsymbol\alpha} $ is an approximate solution looked for by GSA. By comparing $ f^{\mathrm{LLA}} = \widetilde{\mathbf{X}}_{\mathcal{H}}\hat{\boldsymbol\alpha} $ with $ f^{\mathrm{GSA}} = \mathbf{X}_{\mathcal{H}}\hat{\boldsymbol\alpha} $, it is expected that GSA can provide more accurate solutions than LLA. 
	Because in comparison with the non-approximate optimal solution $ f^{*} =  \mathbf{X}_{\mathcal{H}}\widetilde{\boldsymbol\alpha} $ to the problem~\eqref{op:kernel models}, there are two approximate items in $ f^{\mathrm{LLA}} = \widetilde{\mathbf{X}}_{\mathcal{H}}\hat{\boldsymbol\alpha} $ while there is only one in $ f^{\mathrm{GSA}} = \mathbf{X}_{\mathcal{H}}\hat{\boldsymbol\alpha} $. Note that the approximation error $ \| f^{\mathrm{GSA}} - f^{*} \|_{\mathcal{H}} $ for GSA can be computed by
	\begin{equation} 
		\begin{gathered} \label{eq:GSA-f*}
			\left\lVert f^{\mathrm{GSA}}-f^{*} \right\rVert_{\mathcal{H}}^{2} = (\widetilde{\boldsymbol\alpha}-\hat{\boldsymbol\alpha})^{T}\mathbf{K}(\widetilde{\boldsymbol\alpha}-\hat{\boldsymbol\alpha}).
		\end{gathered}
	\end{equation}
	Aided by the equalities~\eqref{eq:LLA-f*} and~\eqref{eq:GSA-f*}, we run experiments with classification tasks to verify our conjecture that GSA can provide more accurate solutions than LLA. 
	
	\emph{In addition}, the equality~\eqref{eq:xa=bw} suggests that $ \hat{\boldsymbol\alpha} $ can be computed from $ \hat{\mathbf{w}} $.
	As suggested by other studies \cite{lan2019scaling,jin2013improved}, the reason to abandon GSA is because $ \hat{\boldsymbol\alpha} $ cannot be calculated efficiently when the related solvers are used as a black box. 
	By contrast, $ \hat{\mathbf{w}} $ can be easily obtained by using efficient linear solvers. 
	However, the way of computing $ \hat{\boldsymbol\alpha} $ from $ \hat{\mathbf{w}} $ is as computationally efficient as calculating $ \hat{\mathbf{w}} $.
	
	\begin{corollary} \label{cor: LLA2GSA}
		If SPA and LLA are strongly equivalent, and $\hat{\mathbf{w}}$ is optimal to the problem~\eqref{op:solvableLLA}, then
		\begin{equation} \label{step:LLA2GSA}
			\begin{gathered}
				\hat{\boldsymbol\alpha} \gets \mathbf{G}^{\dagger}\hat{\mathbf{w}} \text{ with } \mathbf{G} = \mathbf{A}^{T}\mathbf{K}_{mn} = \langle \mathbf{B}_{\mathcal{H}},\mathbf{X}_{\mathcal{H}} \rangle_{\mathcal{H}}
			\end{gathered}
		\end{equation}
		is an optimal solution to the problem~\eqref{op:approximate_solvable models}.
	\end{corollary}
	\emph{Proof.} Since $ \mathbf{B}_{\mathcal{H}}\mathbf{G}\hat{\boldsymbol\alpha} = \widetilde{\mathbf{X}}_{\mathcal{H}}\hat{\boldsymbol\alpha} = \mathbf{B}_{\mathcal{H}}\hat{\mathbf{w}} $ and $ \mathbf{B}_{\mathcal{H}} $ is linearly independent, there is the equality $ \mathbf{G}\hat{\boldsymbol\alpha} = \hat{\mathbf{w}} $, which indicates $ \hat{\boldsymbol\alpha} = \mathbf{G}^{\dagger}\hat{\mathbf{w}} $. \hfill $ \Box $
	
	Note that the running time for yielding $ \mathbf{G} $ is $ \mathcal{O}(nms) $ when standard Nystr\"{o}m method is taken, and the time for calculating $ \mathbf{G}^{\dagger} $ is always $ \mathcal{O}(ns^{2}) $ where $ s\leq m $. Therefore, the step~\eqref{step:LLA2GSA} does not add to computational cost while it enables us to use GSA as efficiently as LLA.

	Even though Corollary~\ref{cor: LLA2GSA} is based on the strong equivalence between LLA and SPA. 
	In fact, it holds even if there is only weak equivalence between LLA and SPA.
	
	\begin{proposition} \label{prop:w2a}
		Suppose the weak equivalence between LLA and SPA holds. If $ \hat{\mathbf{w}} $ is optimal to the problem~\eqref{op:solvableLLA}, $ \mathbf{G}^{\dagger}\hat{\mathbf{w}} $ is optimal to the problem~\eqref{op:approximate_solvable models}.
	\end{proposition}	
	To conclude, LLA and GSA can be implemented to share the same training procedure, as summarized in Algorithm~\ref{alg:general}.
	
	\begin{algorithm}[!t]
		\small
		\caption{Algorithms for running LLA (including NCR) or GSA} 
		\label{alg:general}
		\begin{algorithmic}
			\Statex {\bfseries Training phase}
			\Statex {\bfseries Input:} Data $ \mathbf{X}_{\mathcal{H}} = \Phi(\mathbf{X}) $, a vector of labels $ \mathbf{y} $, the number of landmarks $ m $, the targeted low-dimension $ s \leq m $.
			\Statex ~~~1) Generate a set of landmarks $ \mathbf{C}_{\mathcal{H}} \in \mathcal{H}^{m} $ by a specific sampling strategy \cite{kumar2012sampling,sun2015review,gittens2016revisiting}.
			\Statex ~~~2) Compute $ \mathbf{K}_{nm}^{T} = \mathbf{K}_{mn} = \langle \mathbf{C}_{\mathcal{H}}, \mathbf{X}_{\mathcal{H}} \rangle_{\mathcal{H}} $ and $ \mathbf{K}_{mm} = \langle \mathbf{C}_{\mathcal{H}}, \mathbf{C}_{\mathcal{H}} \rangle_{\mathcal{H}} $.
			\Statex ~~~3) Obtain $ \mathbf{A} \in \mathbb{R}^{m\times s} $ that satisfies $ \mathbf{A}^{T}\mathbf{K}_{mm}\mathbf{A} = \mathbf{I} $ by using a specific Nystr\"{o}m method \cite{lim2015double,lim2018multi};
			\Statex ~~~4) Compute $ \mathbf{G} \gets \mathbf{A}^{T}\mathbf{K}_{mn} $;
			\Statex ~~~5) Get $ \hat{\mathbf{w}} $ by using an efficient linear solver upon $ \mathbf{G} $.
			\Statex {\bfseries Output}: $ \hat{\mathbf{w}} $.
			\\\hrulefill
			\Statex {\bfseries Testing phase} (LLA: Low-rank linearization approach; GSA: Gram matrix substitution approach)
			\Statex {\bfseries Input:} A test data point $ \mathbf{z} $.
			\Statex ~~~~{\bf If implementing LLA: } $ \mathbf{t} \gets \mathbf{A}^{T}\langle \mathbf{C}_{\mathcal{H}}, \Phi(\mathbf{z}) \rangle_{\mathcal{H}} $.
			\Statex ~~~~{\bf If implementing GSA: } $ \mathbf{t} \gets (\mathbf{G}^{\dagger})^{T}\langle \mathbf{X}_{\mathcal{H}}, \Phi(\mathbf{z}) \rangle_{\mathcal{H}} $.
			\Statex \textbf{Prediction:} $ \hat{\mathbf{w}}^{T}\mathbf{t} $.
		\end{algorithmic}
	\end{algorithm}
	
	When there is only weak equivalence, though the computed approximate solution $ f^{\mathrm{LLA}} $ may not satisfies the equality~\eqref{eq:xa=bw}, the projection of $ f^{\mathrm{LLA}} $ onto $ \mathrm{span}(\widetilde{\mathbf{X}}_{\mathcal{H}}) $ always meets the equality~\eqref{eq:xa=bw}. 
	Therefore, all the analysis above still works when using the projection of $ f^{\mathrm{LLA}} $ instead. 
	Moreover, since the weak equivalence implies that each optimal solution to SPA is also optimal to LLA, SPA always serves as an alternative perspective for LLA (including NCR).

	\subsection{Sufficient Conditions for the Equivalence}
	The remaining question is when the two types of equivalence hold.
	First, note that in the problem~\eqref{op:recastLLA}, it holds that $ \langle f, \mathbf{x}_{\mathcal{H}}^{i} \rangle_{\mathcal{H}} = \langle f, \widetilde{\mathbf{x}}_{\mathcal{H}}^{i} \rangle_{\mathcal{H}} $ for each $ 1\leq i \leq n $ and each feasible solution $ f \in \mathrm{span}(\mathbf{B}_{\mathcal{H}}) $. 
	This result is due to
	\begin{equation}
		\begin{gathered}
			\langle \mathbf{B}_{\mathcal{H}}, \widetilde{\mathbf{X}}_{\mathcal{H}} \rangle_{\mathcal{H}} = \langle \mathbf{B}_{\mathcal{H}}, \mathbf{B}_{\mathcal{H}}\langle \mathbf{B}_{\mathcal{H}}, \mathbf{X}_{\mathcal{H}} \rangle_{\mathcal{H}} \rangle_{\mathcal{H}} \\= \langle \mathbf{B}_{\mathcal{H}},\mathbf{B}_{\mathcal{H}} \rangle_{\mathcal{H}}\langle \mathbf{B}_{\mathcal{H}}, \mathbf{X}_{\mathcal{H}} \rangle_{\mathcal{H}} = \langle \mathbf{B}_{\mathcal{H}}, \mathbf{X}_{\mathcal{H}} \rangle_{\mathcal{H}} .
		\end{gathered}
	\end{equation}
	Therefore, the problem~\eqref{op:recastLLA} is exactly equivalent to
	\begin{equation} \label{op:recastLLA2}
		\begin{aligned}
			\argmin_{f\in\mathcal{H}} \hat{\mathcal{R}}(f, \widetilde{\mathbf{X}}_{\mathcal{H}}) =& \cfrac{1}{n}\sum_{i=1}^{n}\mathcal{L}(\langle f, \widetilde{\mathbf{x}}_{\mathcal{H}}^{i} \rangle_{\mathcal{H}}, y_{i}) + \Omega(\|f\|_{\mathcal{H}}^{2})\\
			\text{ subject to }& f \in \mathrm{span}(\mathbf{B}_{\mathcal{H}}).
		\end{aligned}
	\end{equation}
	where $ \mathbf{B}_{\mathcal{H}} $ is a learned orthogonal basis.
	One can observe that the only difference between the problems~\eqref{op:projected kernel models} and~\eqref{op:recastLLA2} lies on the constraint. 
	Therefore, it is obvious that the strong equivalence will hold when there is
	\begin{equation} \label{eq:span(X)=span(B)}
		\begin{gathered}
			\mathrm{span}(\widetilde{\mathbf{X}}_{\mathcal{H}}) = \mathrm{span}(\mathbf{B}_{\mathcal{H}}) .
		\end{gathered}
	\end{equation}
	Since $ \widetilde{\mathbf{X}} = \mathbf{B}_{\mathcal{H}}\langle \mathbf{B}_{\mathcal{H}}, \mathbf{X}_{\mathcal{H}} \rangle_{\mathcal{H}} $, we already have $ \mathrm{span}(\widetilde{\mathbf{X}}_{\mathcal{H}}) \subseteq \mathrm{span}(\mathbf{B}_{\mathcal{H}}) $. However, the other direction does not necessarily hold. 
	For example, when the set of landmarks $ \mathbf{C}_{\mathcal{H}} $ contains some points that are not located inside $ \mathrm{span}(\mathbf{X}_{\mathcal{H}}) $. 
	To the best of our knowledge, this could only happen when non-kernel k-means clustering sampling strategies are used \cite{zhang2010clustered,pourkamali2018randomized}. 
	For other sampling strategies, $ \mathbf{C}_{\mathcal{H}} $ comes with an associated condition $ \mathbf{C}_{\mathcal{H}} = \mathbf{X}_{\mathcal{H}}\mathbf{P} $ where $ \mathbf{P} \in \mathbb{R}^{n\times m} $ is a sampling matrix. 
	This equality definitely asserts that $ \mathrm{span}(\widetilde{\mathbf{X}}) = \mathrm{span}(\mathbf{B}_{\mathcal{H}}) $.
	\begin{proposition} \label{prop:C=XP}
		If $ \mathbf{C}_{\mathcal{H}} = \mathbf{X}_{\mathcal{H}}\mathbf{P} $, it holds that $ \mathrm{span}(\widetilde{\mathbf{X}}_{\mathcal{H}}) = \mathrm{span}(\mathbf{B}_{\mathcal{H}}) $. Henceforth, there is strong equivalence between SPA and LLA.
	\end{proposition}
	
	Even if $ \mathrm{span}(\widetilde{\mathbf{X}}) $ is a proper subset of $ \mathrm{span}(\mathbf{B}) $, the two types of equivalence are closely related to the representer theorem. 
	In particular, we need to categorize the representer theorem. 
	For kernel machine~\eqref{op:kernel models}, each solution $ f \in \mathcal{H} $
	can be uniquely decomposed as $ f = f_{r} + f_{n} $ such that $ f_{r} \in \mathrm{span}(\mathbf{X}_{\mathcal{H}}) $ and $ f_{n} \in \mathrm{span}(\mathbf{X}_{\mathcal{H}})^{\perp} $ (the complement of $ \mathrm{span}(\mathbf{X}_{\mathcal{H}}) $). There are two specific types of the representer theorem.
	\begin{definition}
		The objective of kernel machine~\eqref{op:kernel models} is said to satisfy the strong representer theorem if whenever a solution $ f \in \mathcal{H} $ comes with $ f_{n} \not= \mathbf{0} $, there is $ \hat{\mathcal{R}}(f_{r}) < \hat{\mathcal{R}}(f) $.
	\end{definition}
	\begin{definition}
		The objective of kernel machine~\eqref{op:kernel models} is said to satisfy the weak representer theorem if there is $ \hat{\mathcal{R}}(f_{r}) \leq \hat{\mathcal{R}}(f) $ for each solution $ f \in \mathcal{H} $.
	\end{definition}
	
	The difference between the strong and weak type of the representer theorem is that the weak one does not exclude the case when there exists an optimal solution that is not located inside $ \mathrm{span}(\mathbf{X}_{\mathcal{H}}) $. 
	Previous studies have already provided some sufficient conditions for these two types of the representer theorem \cite{Schlkopf2001AGR,dinuzzo2012representer,yu2013characterizing}. 
	For example, if the regularizing function $ \Omega:[0,+\infty] \mapsto [-\infty,+\infty] $ is strictly increasing, then the objective of kernel machine~\eqref{op:kernel models} satisfies the strong representer theorem; if $ \Omega $ is non-decreasing, then the objective of kernel machine~\eqref{op:kernel models} satisfies the weak representer theorem instead.
	With this categorization, the equivalence between SPA and LLA can be characterized by the following proposition.

	\begin{proposition} \label{prop:reprst to equiv}
		If the objective of kernel machine~\eqref{op:kernel models} satisfies the strong (respectively, weak) representer theorem, then SPA and LLA are strongly (respectively, weakly) equivalent.
	\end{proposition}

	With the proposition above, it is easy to see that some well-established kernel machines, e.g., KRR, KSVM, KPCA, etc., satisfy the strong equivalence, since their optimization problems can be expressed into the form~\eqref{op:kernel models} where
	$ \Omega $ is strictly increasing \cite{dinuzzo2012representer}.

	To sum up, the strong (respectively, weak) representer theorem leads to strong (respectively, weak) equivalence between SPA and LLA.
	Alternatively, for most sampling strategies, the associated condition $ \mathbf{C}_{\mathcal{H}} = \mathbf{X}_{\mathcal{H}}\mathbf{P} $ alone is sufficient to ensure the strong equivalence between LLA and SPA. Therefore, we conclude that the mechanism behind LLA is that it projects all data in the new feature space onto $ \mathrm{span}(\mathbf{B}_{\mathcal{H}}) $ before normally running kernel machines.

	\begin{figure*}[h]
		\centering
		\begin{tabular}{@{\hskip 0ex}r@{\hskip 0ex}cccc}
			& {\small Gaussian Sampling} & {\small Uniform Sampling} & {\small Leverage score Sampling} & {\small K-Means Clustering Sampling}
			\\
			\multirow{2}{*}[0.5ex]{\rotatebox[origin=c]{90}{usps}}
			& \includegraphics[width=0.225\linewidth]{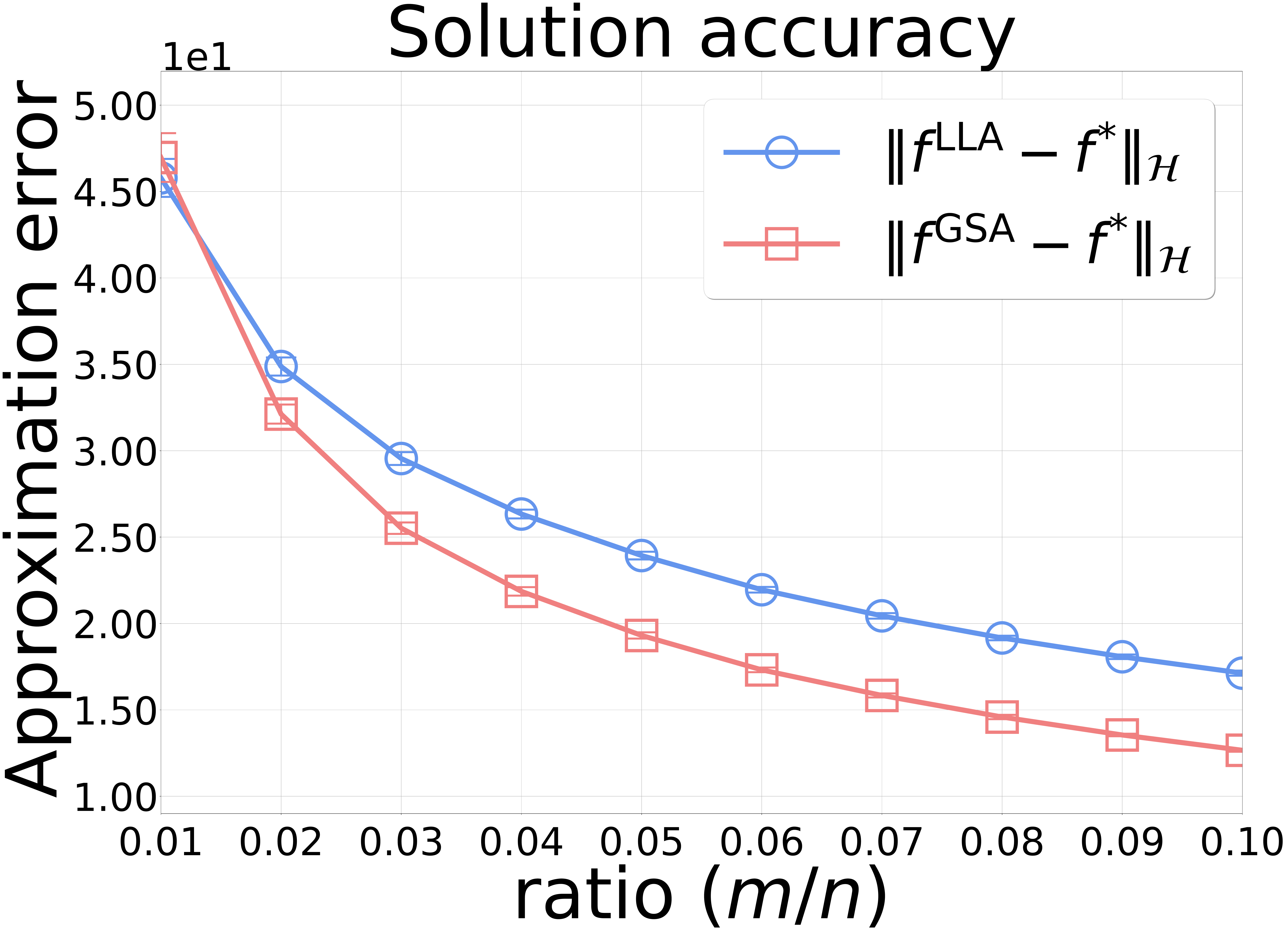}
			& \includegraphics[width=0.225\linewidth]{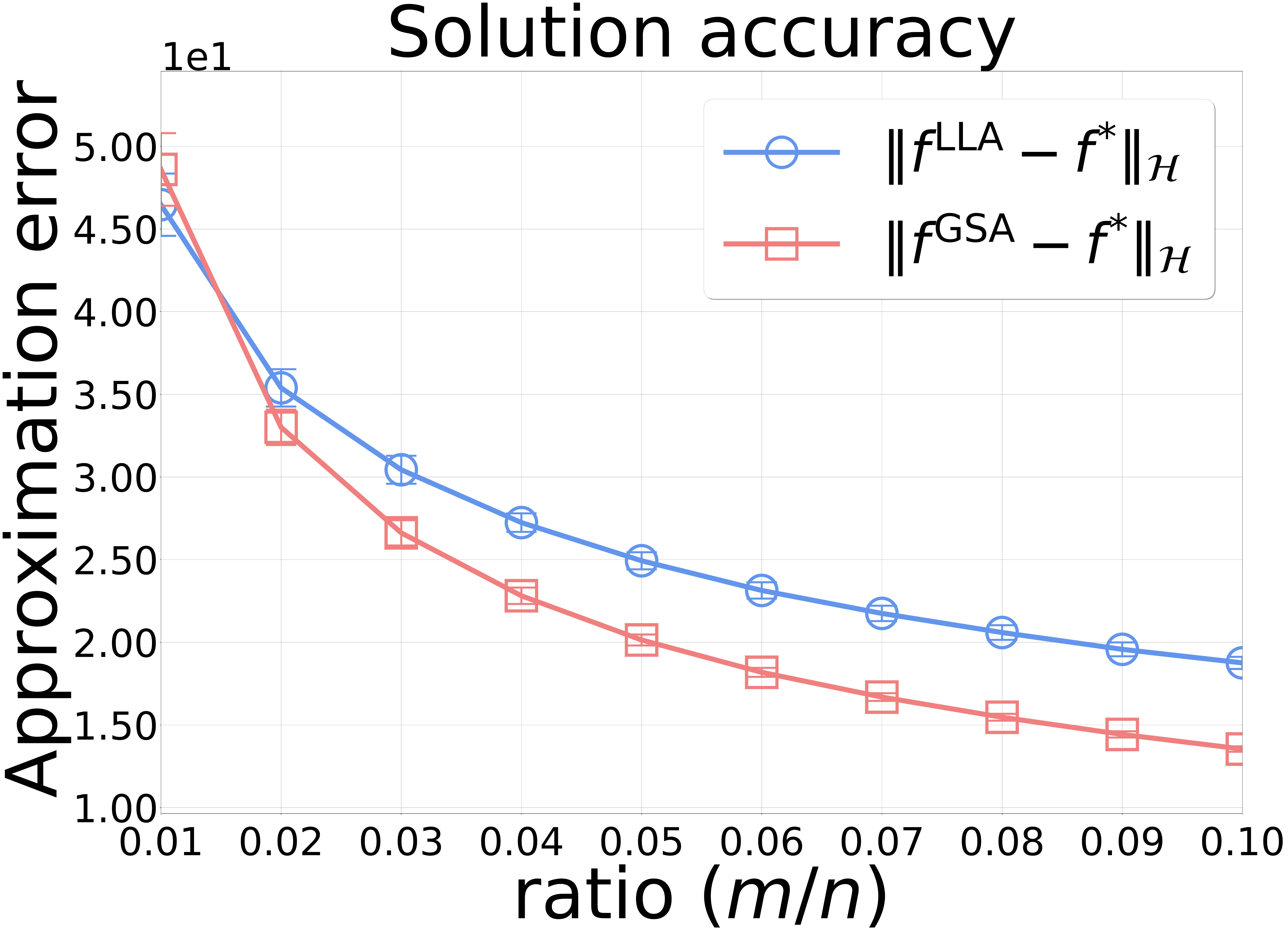}
			& \includegraphics[width=0.225\linewidth]{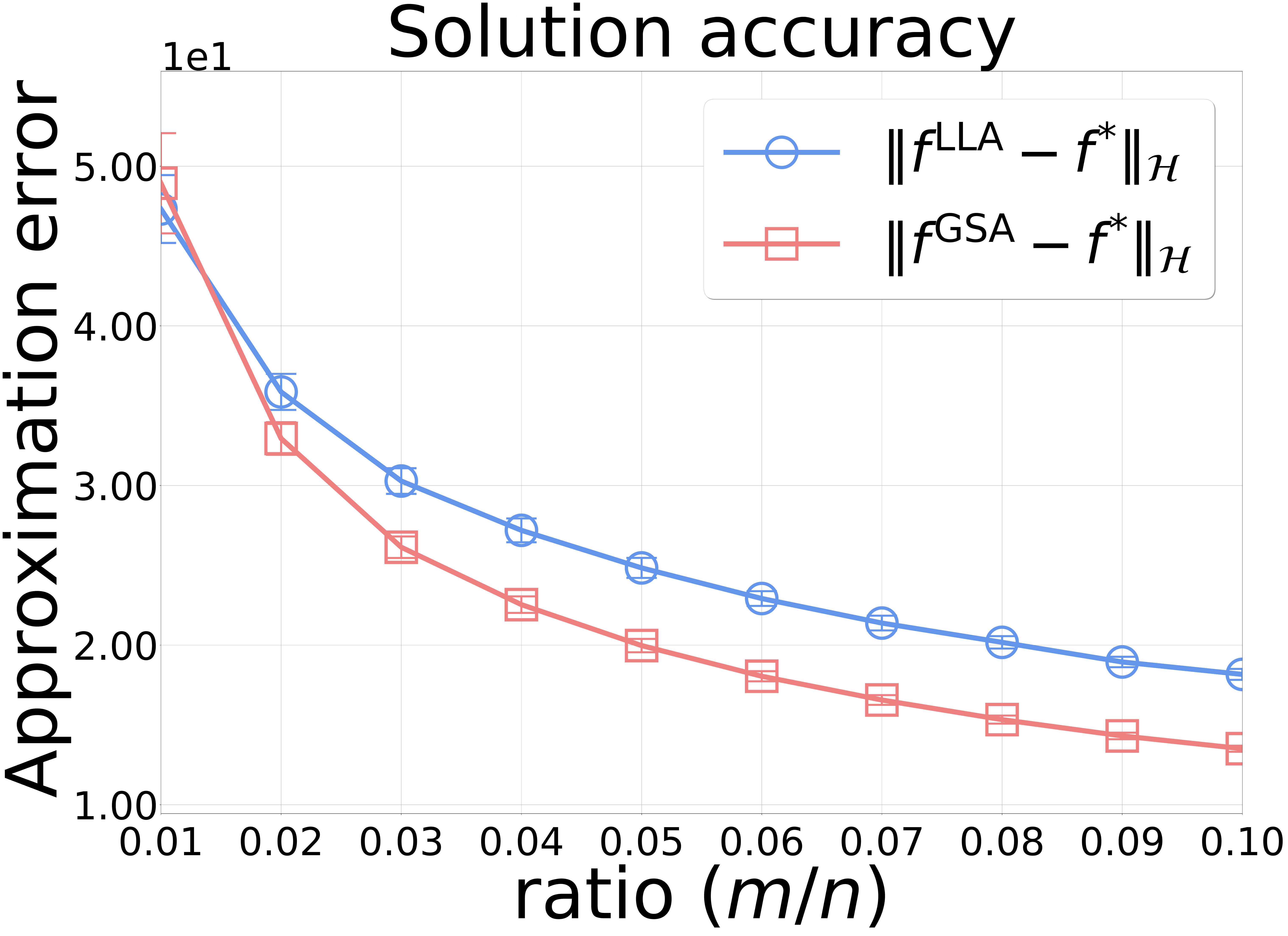}
			& \includegraphics[width=0.225\linewidth]{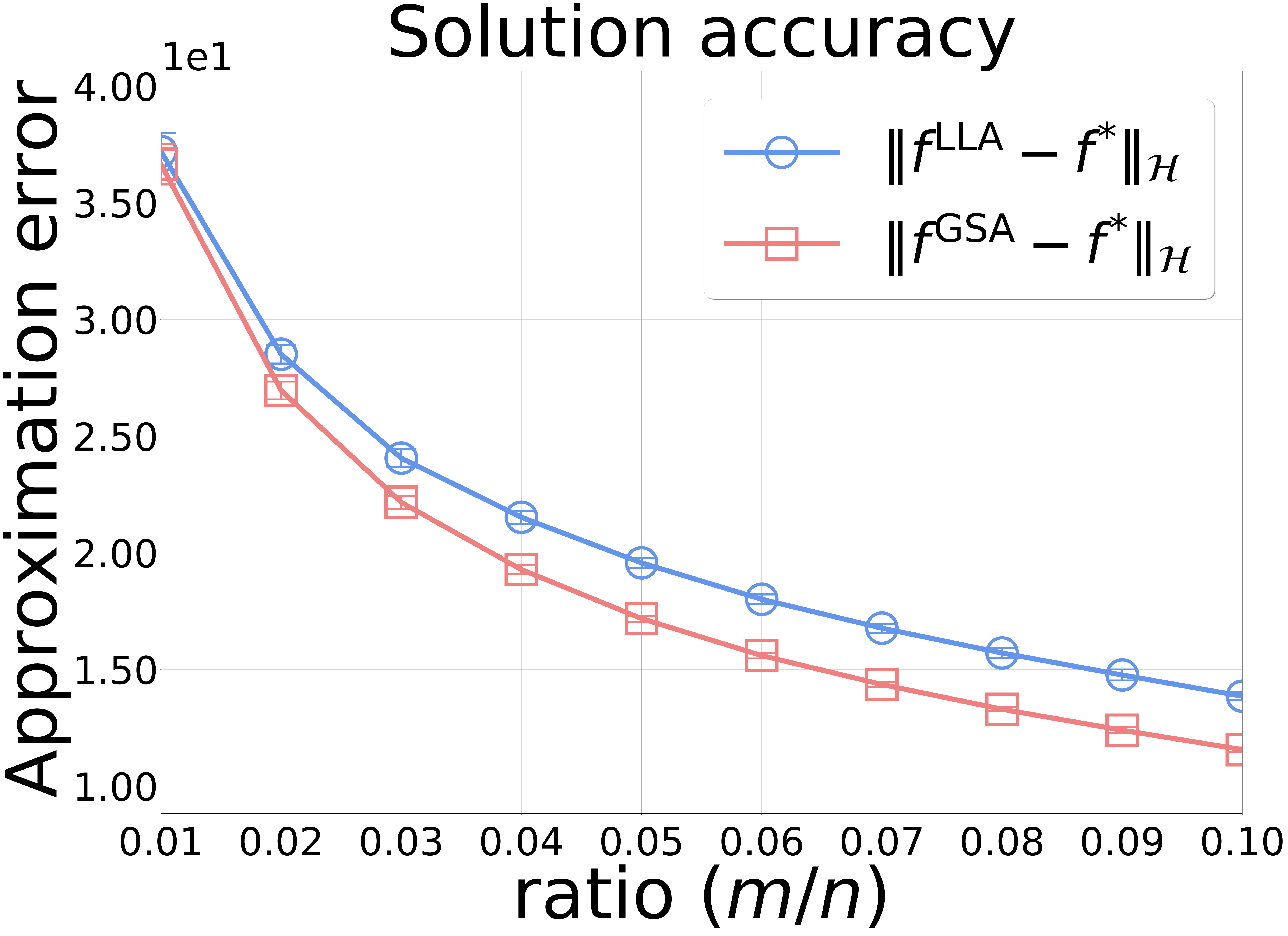} \\
			& \includegraphics[width=0.225\linewidth]{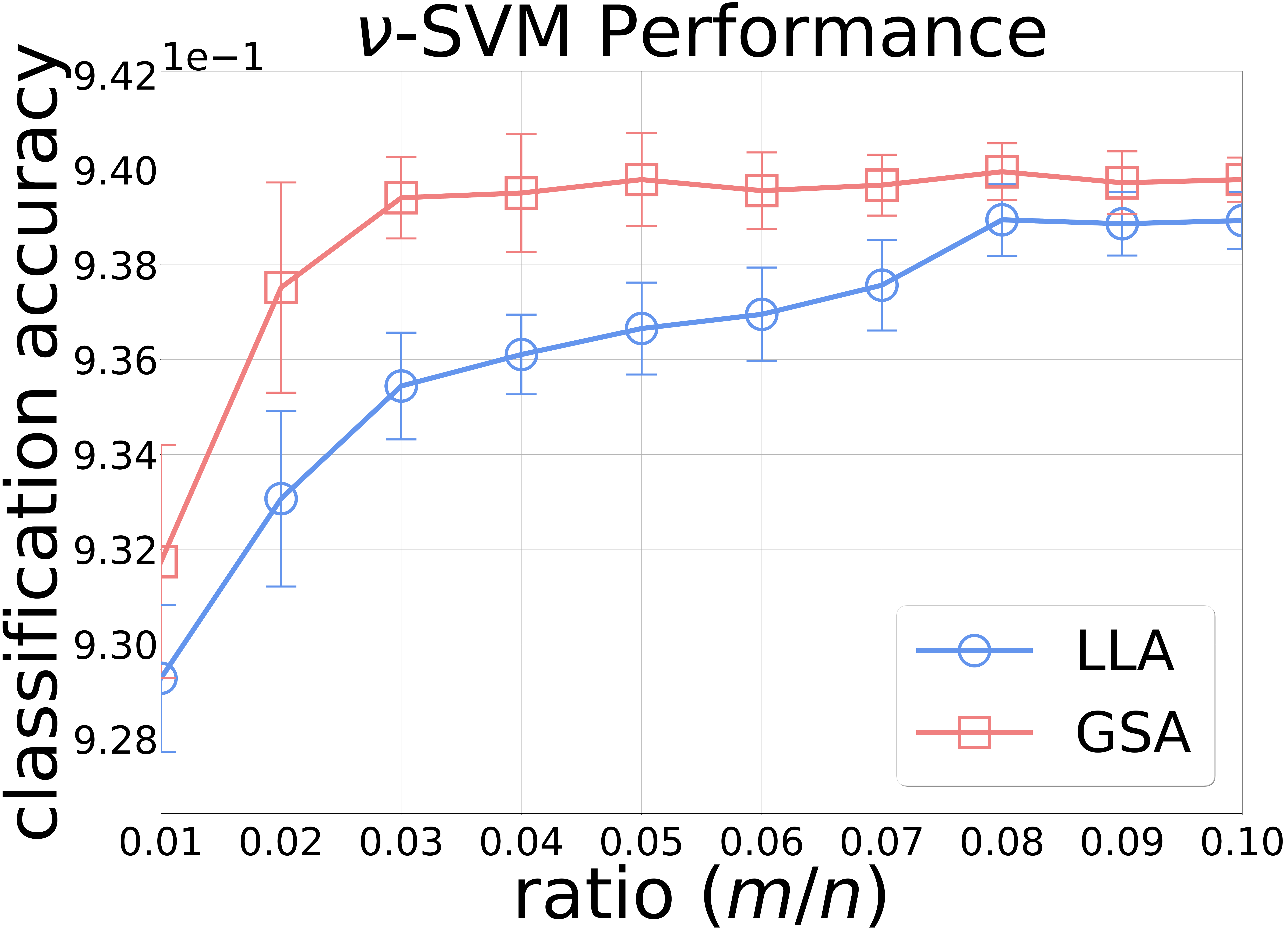}
			& \includegraphics[width=0.225\linewidth]{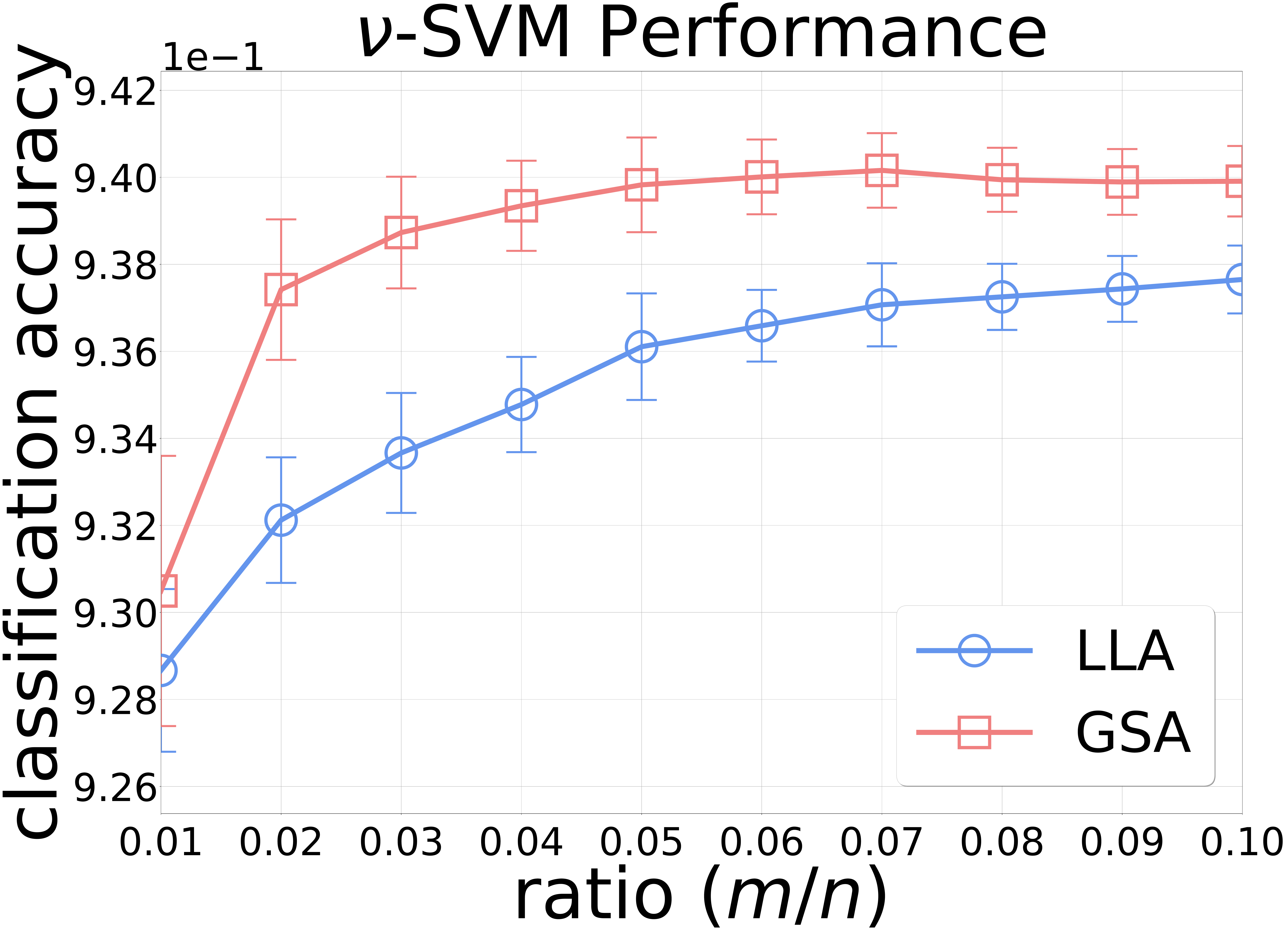}
			& \includegraphics[width=0.225\linewidth]{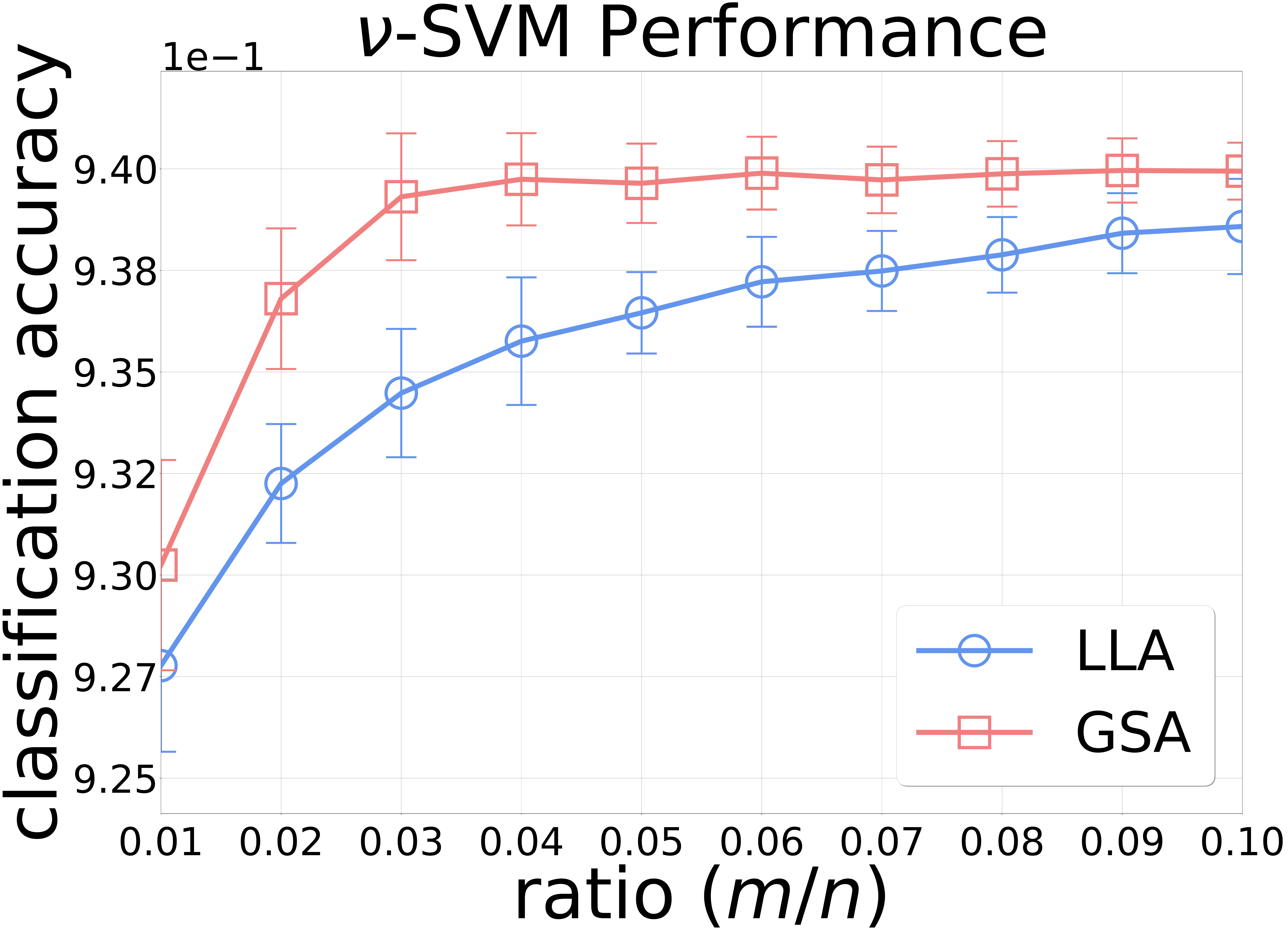}
			& \includegraphics[width=0.225\linewidth]{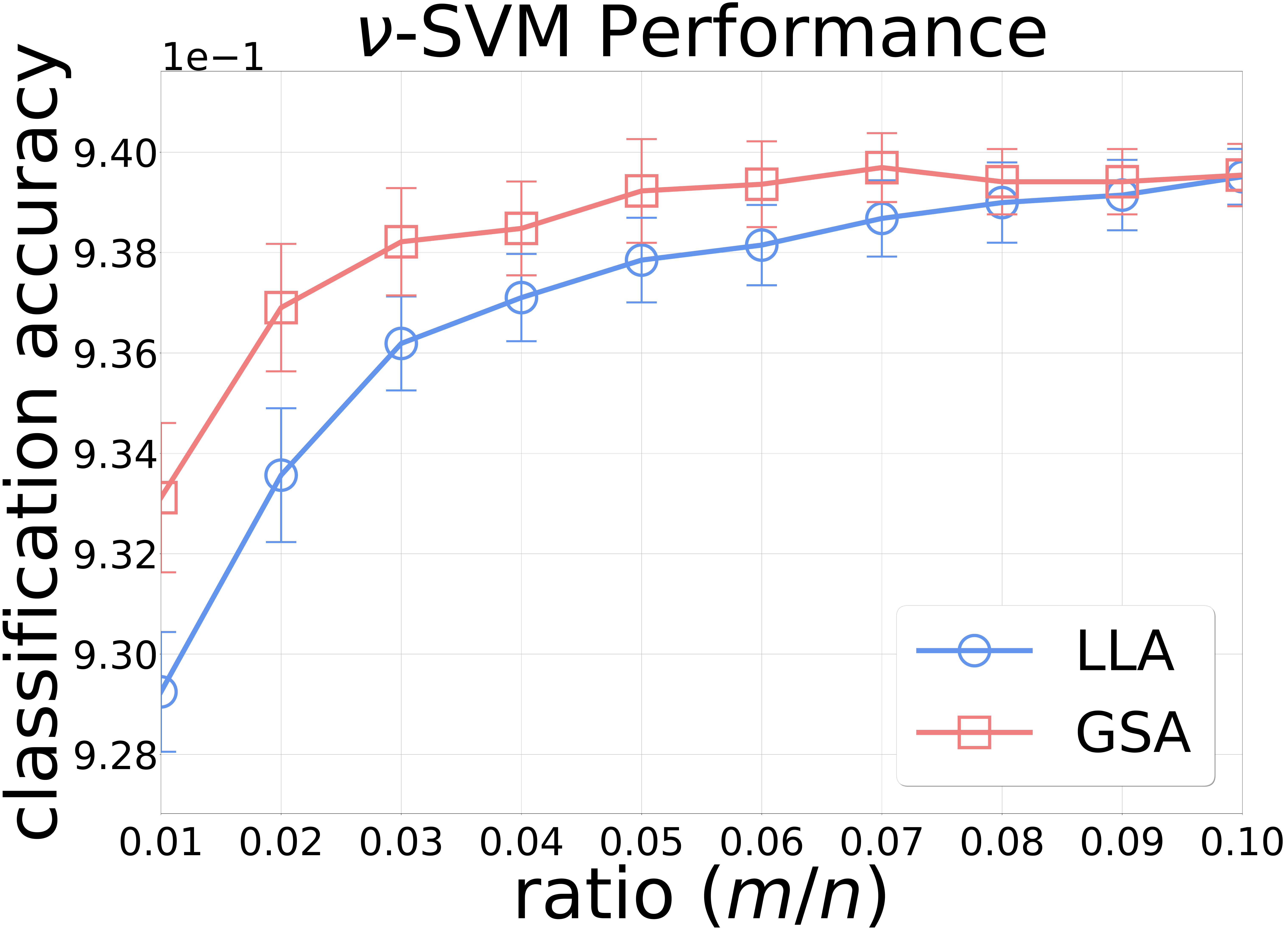}
			\\
			\multirow{2}{*}[0.5ex]{\rotatebox[origin=c]{90}{gisette}}
			& \includegraphics[width=0.225\linewidth]{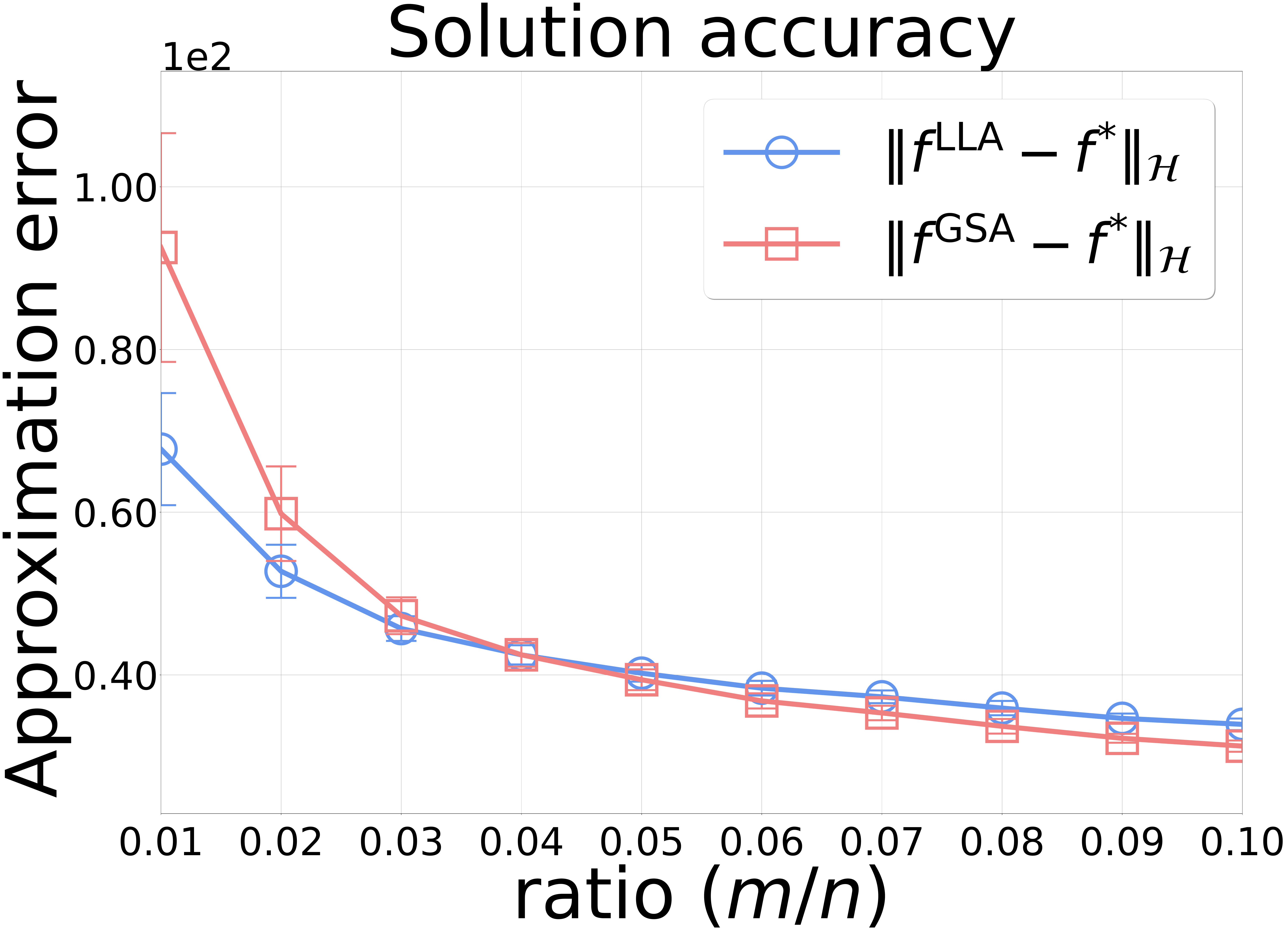}
			& \includegraphics[width=0.225\linewidth]{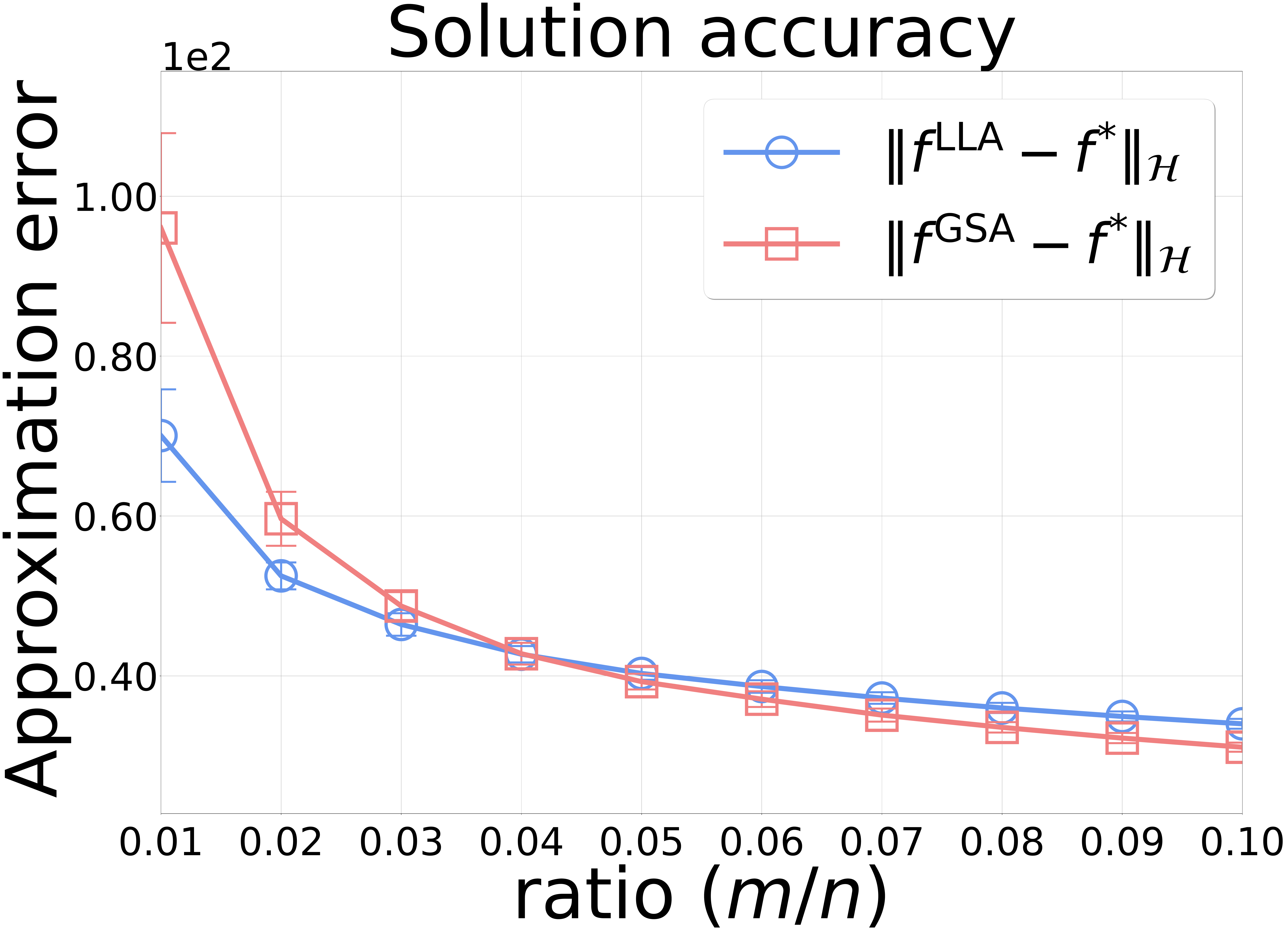}
			& \includegraphics[width=0.225\linewidth]{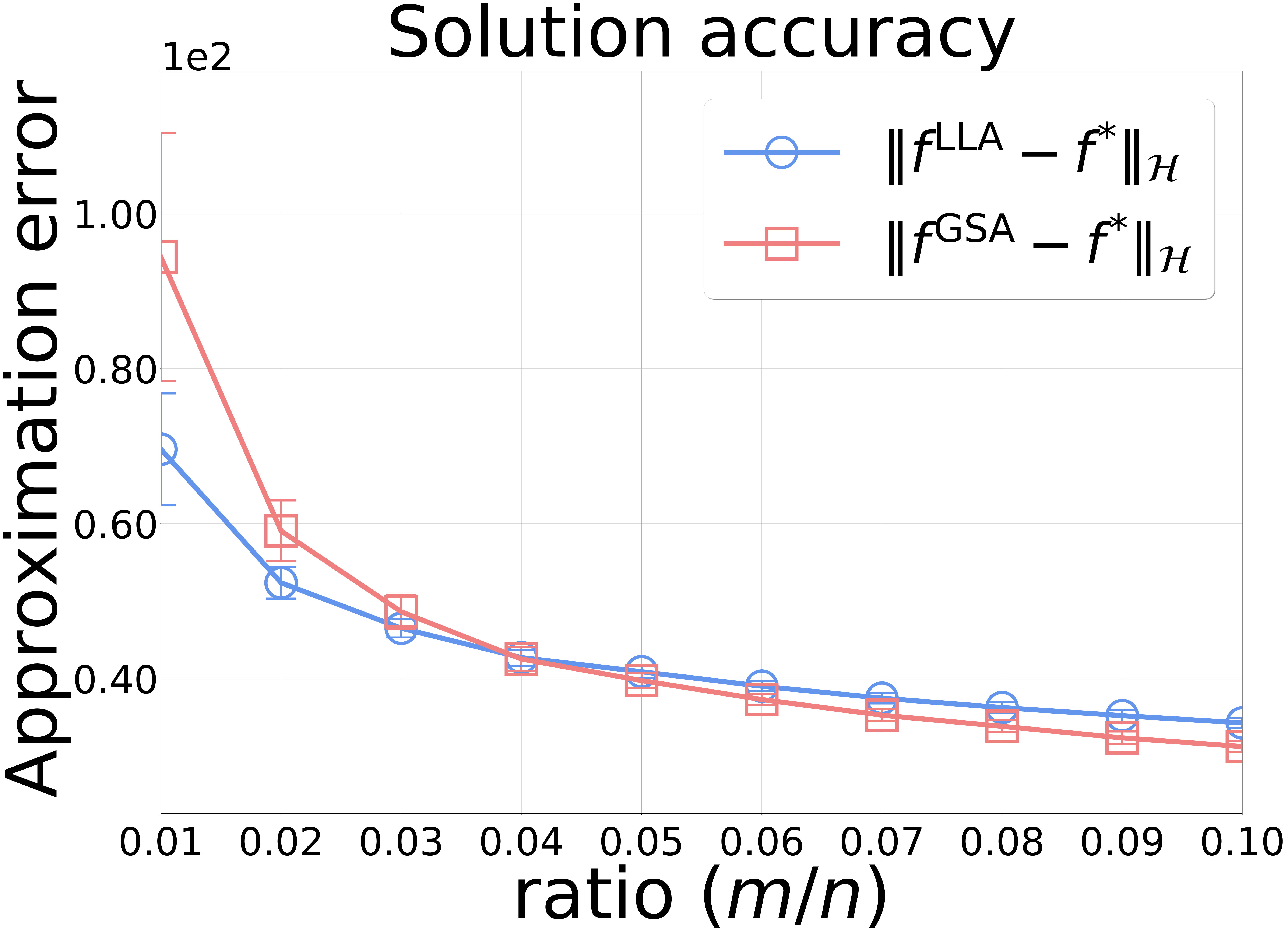}
			& \includegraphics[width=0.225\linewidth]{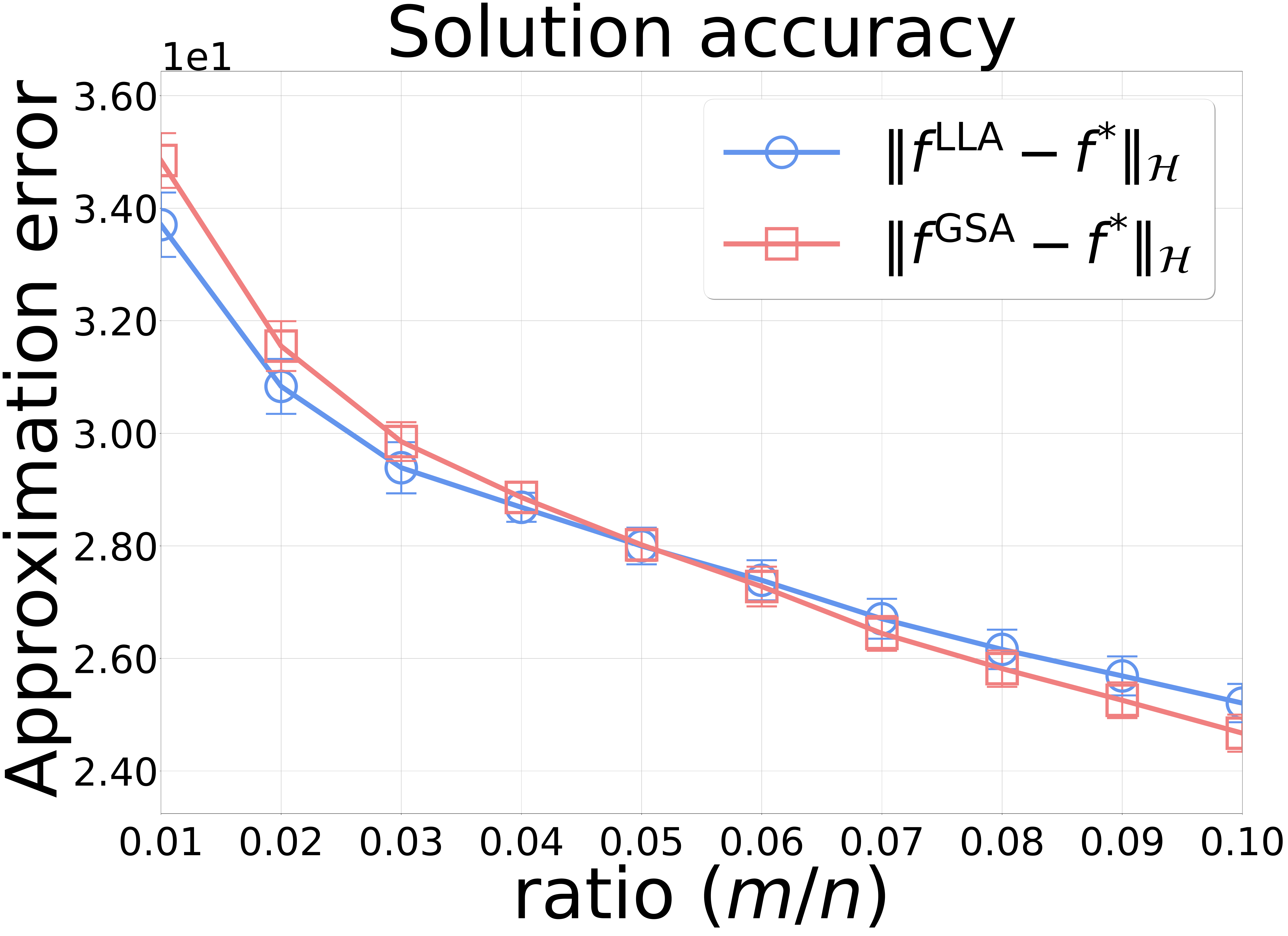} \\
			& \includegraphics[width=0.225\linewidth]{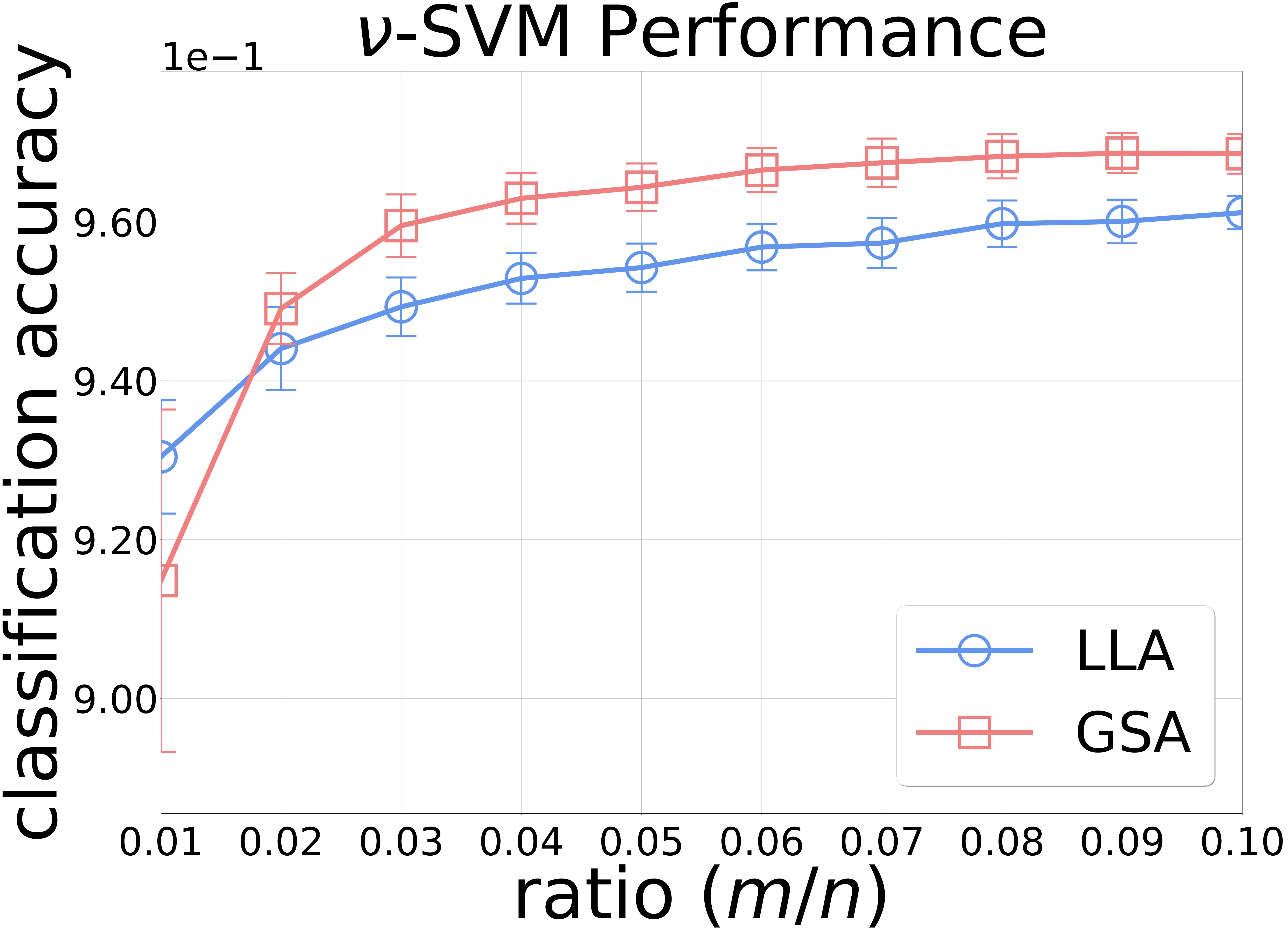}
			& \includegraphics[width=0.225\linewidth]{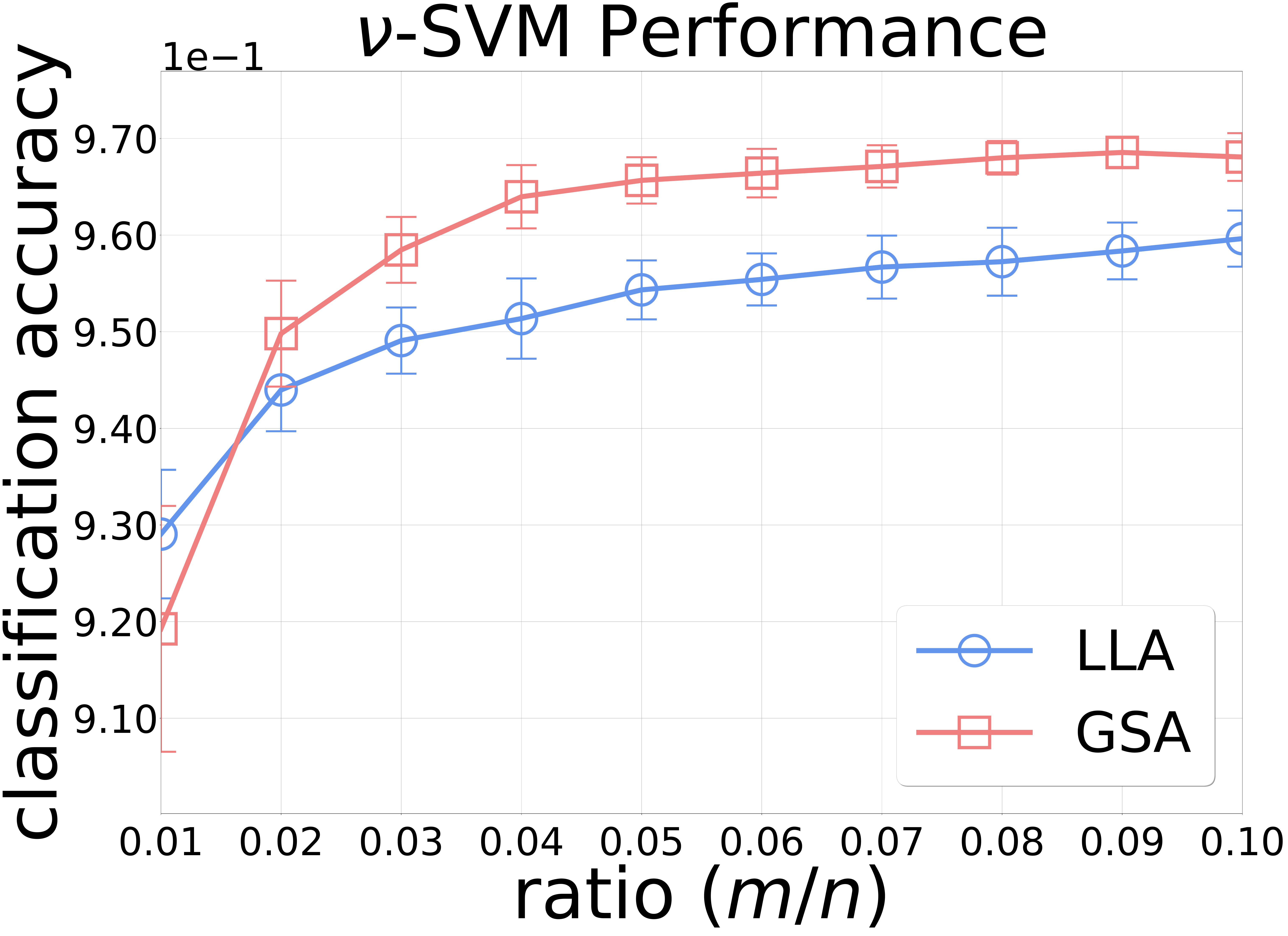}
			& \includegraphics[width=0.225\linewidth]{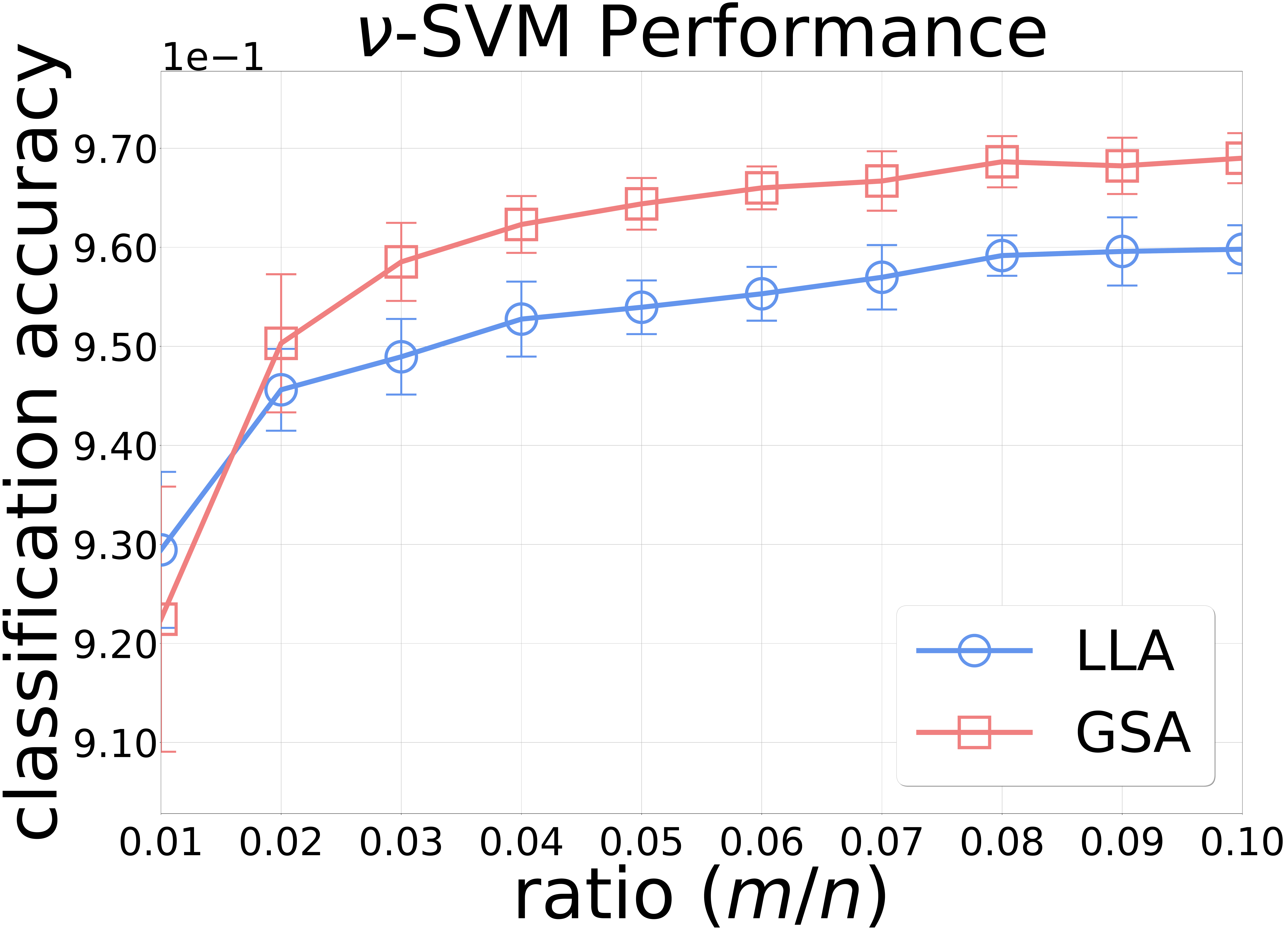}
			& \includegraphics[width=0.225\linewidth]{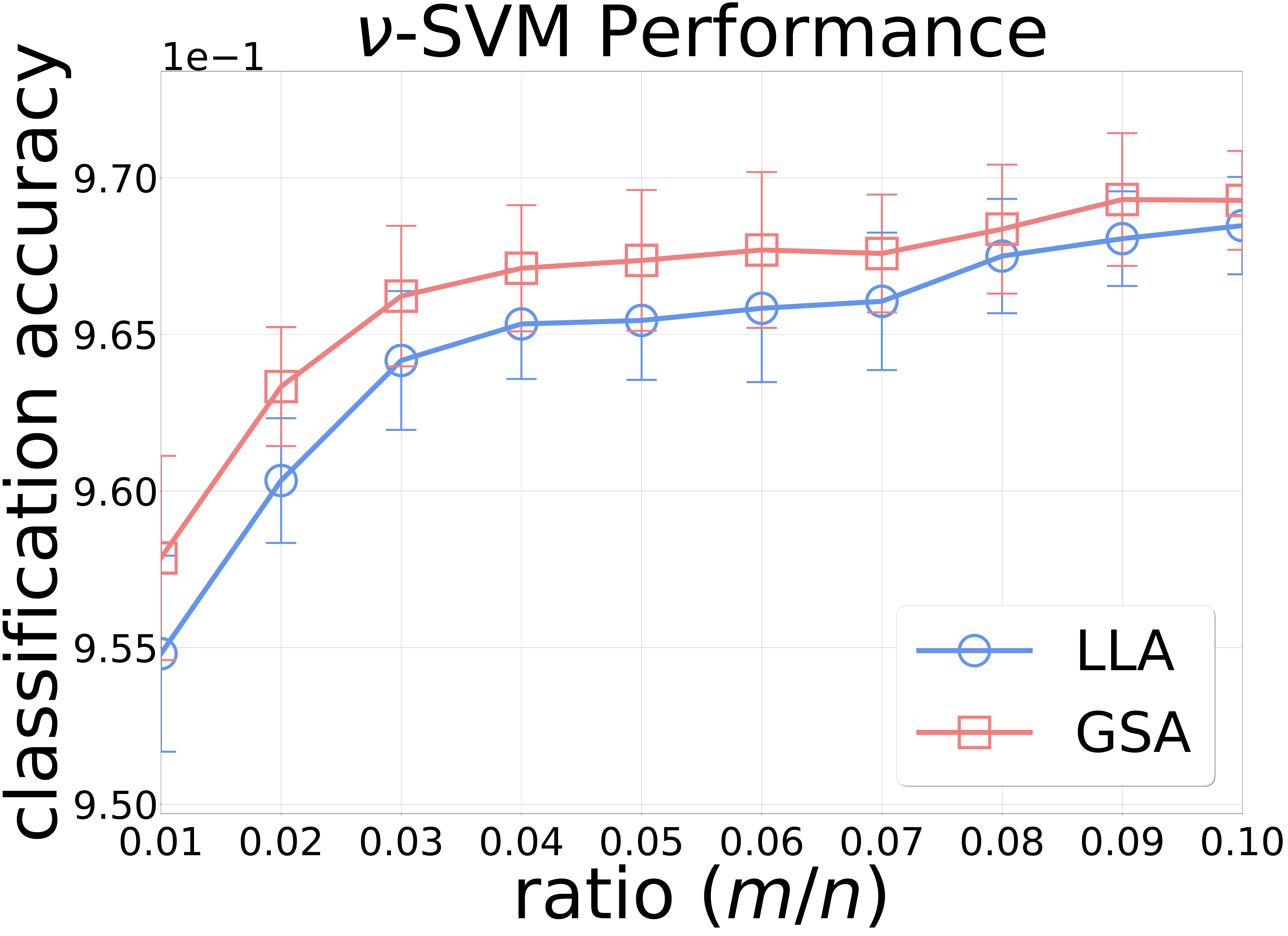}
		\end{tabular}		
		\caption{Comparison between Gram matrix substitution approach (GSA) and low-rank linearization approach (LLA) in terms of approximation error and classidfication accuracy (with $\nu$-SVM). Every two rows correspond to a specific dataset. Each column is related to a certain sampling strategy.}
		\label{fig:result1} 
	\end{figure*}

	\begin{figure*}[h]
		\centering
		\begin{tabular}{@{\hskip 0ex}r@{\hskip 0ex}cccc}
			& {\small Gaussian Sampling} & {\small Uniform Sampling} & {\small Leverage score Sampling} & {\small K-Means Clustering Sampling}
			\\
			\multirow{2}{*}[0.5ex]{\rotatebox[origin=c]{90}{dna}}
			& \includegraphics[width=0.22\linewidth]{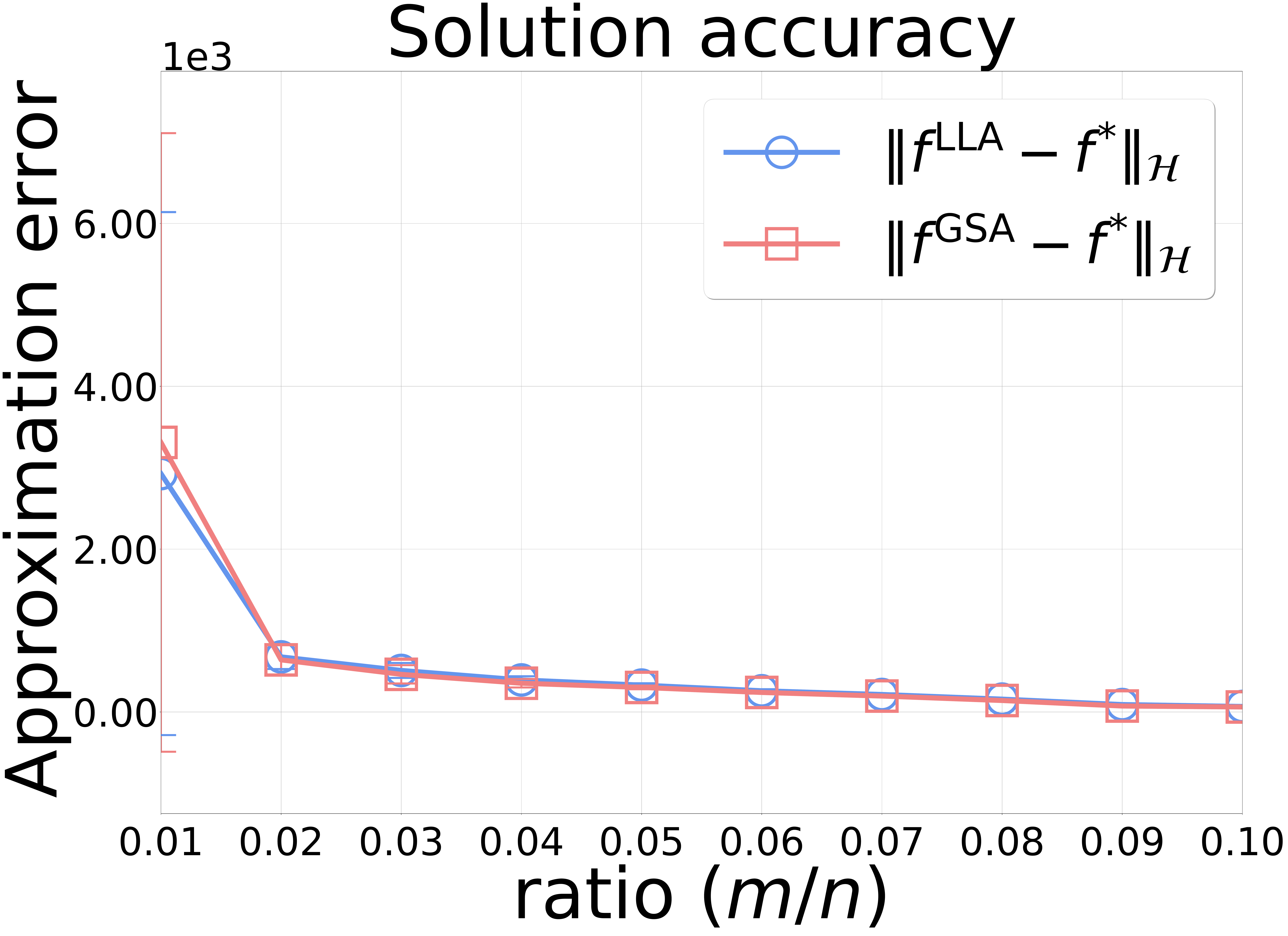}
			& \includegraphics[width=0.22\linewidth]{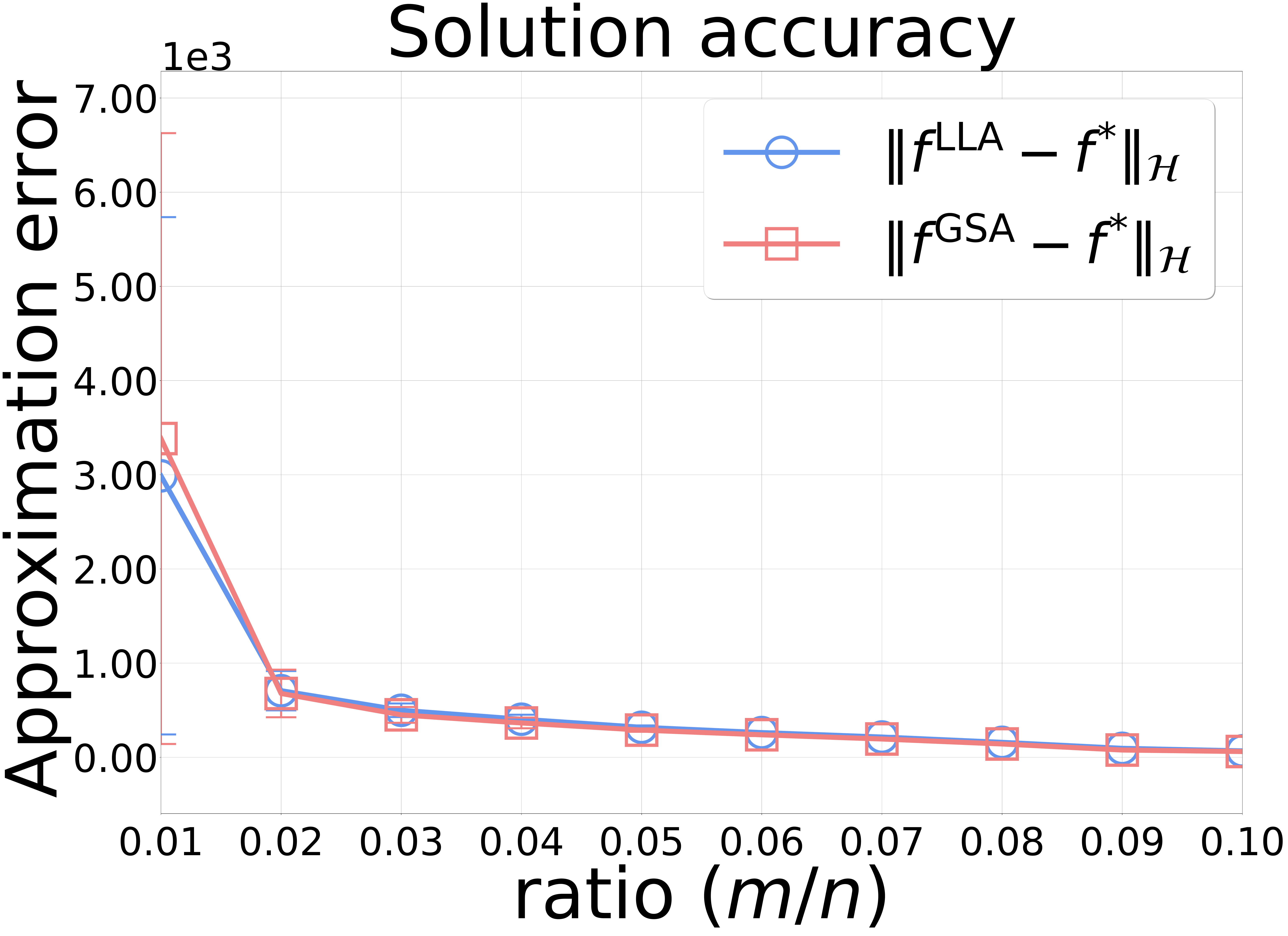}
			& \includegraphics[width=0.22\linewidth]{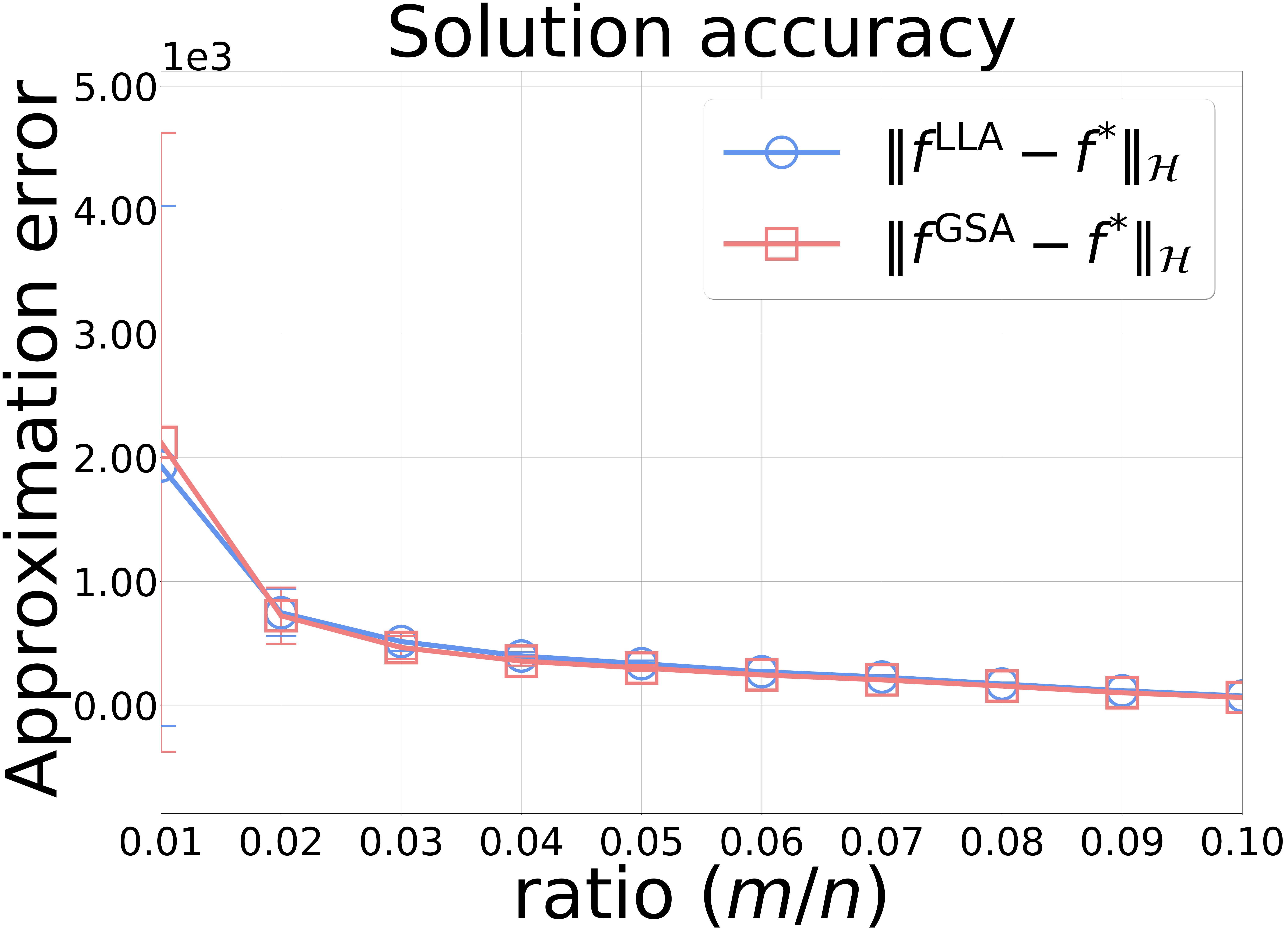}
			& \includegraphics[width=0.22\linewidth]{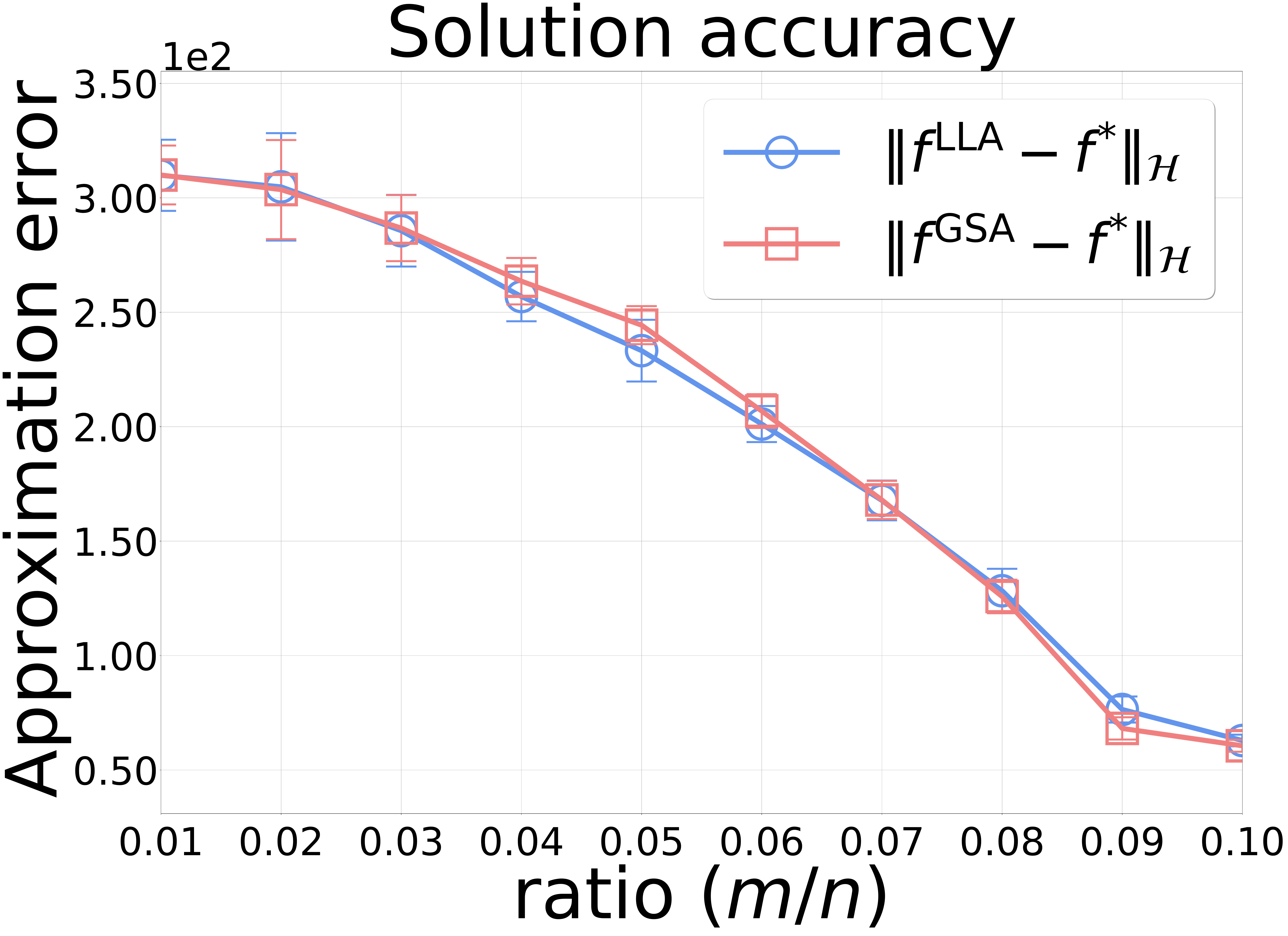} \\
			& \includegraphics[width=0.22\linewidth]{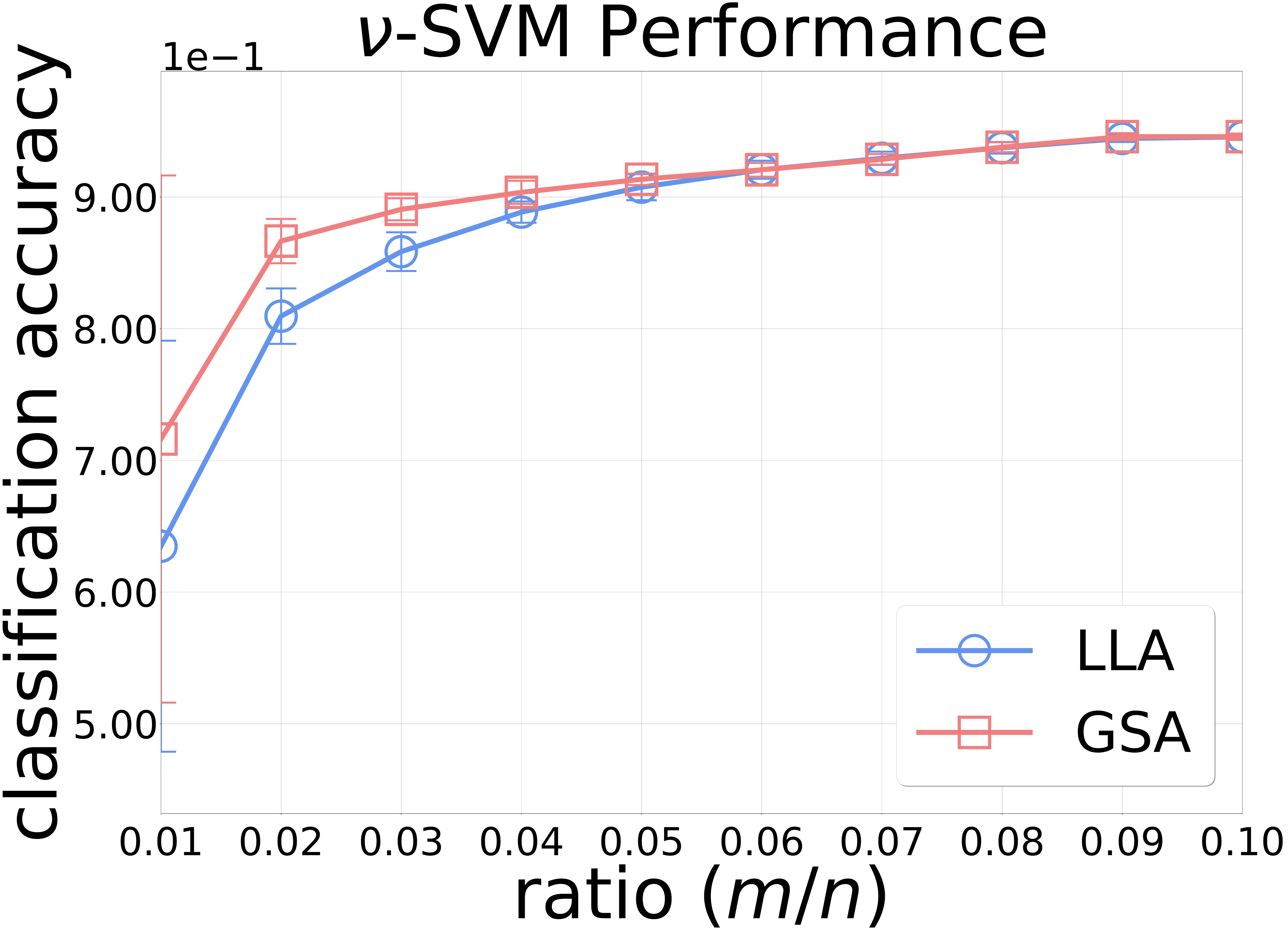}
			& \includegraphics[width=0.22\linewidth]{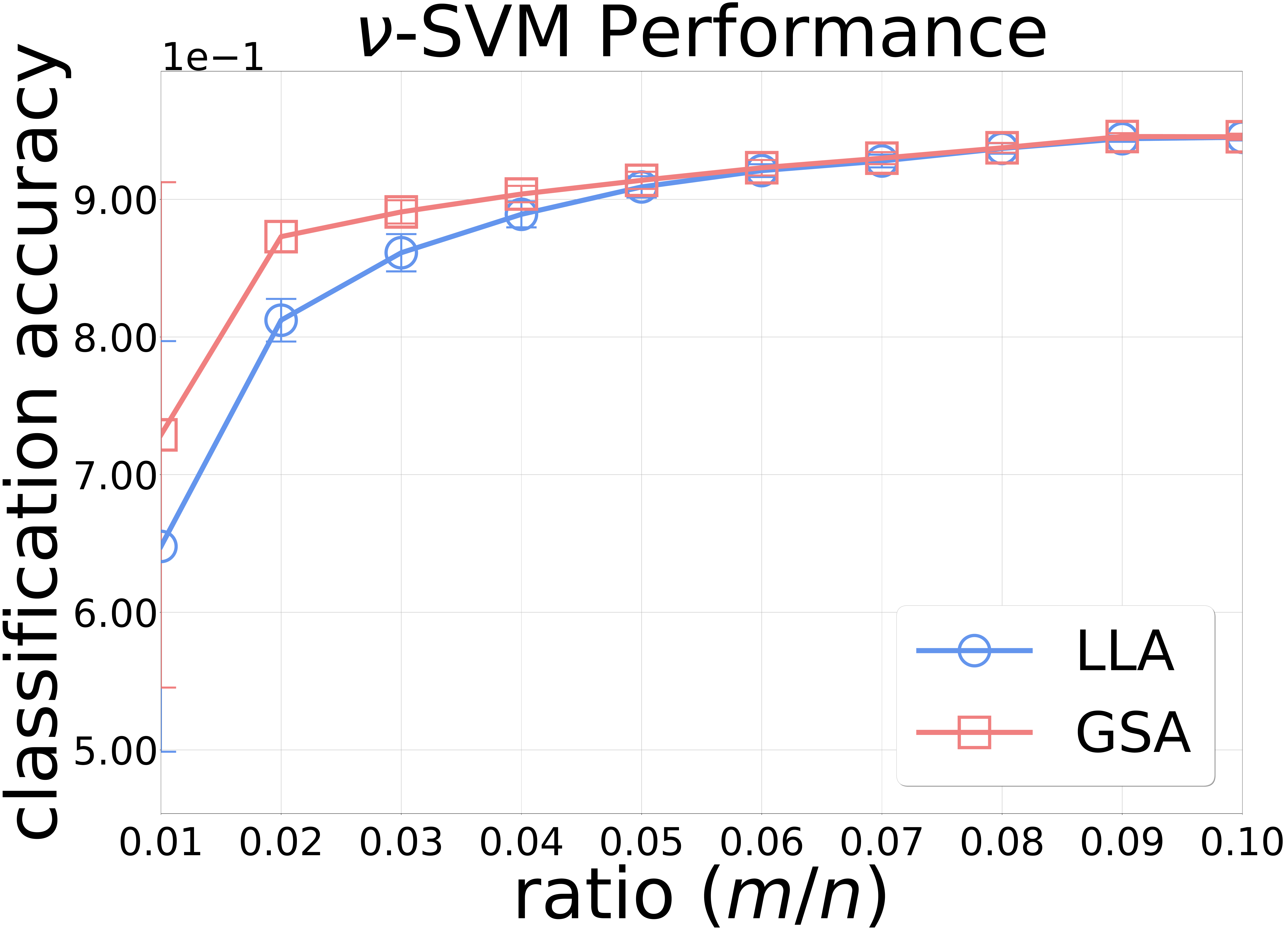}
			& \includegraphics[width=0.22\linewidth]{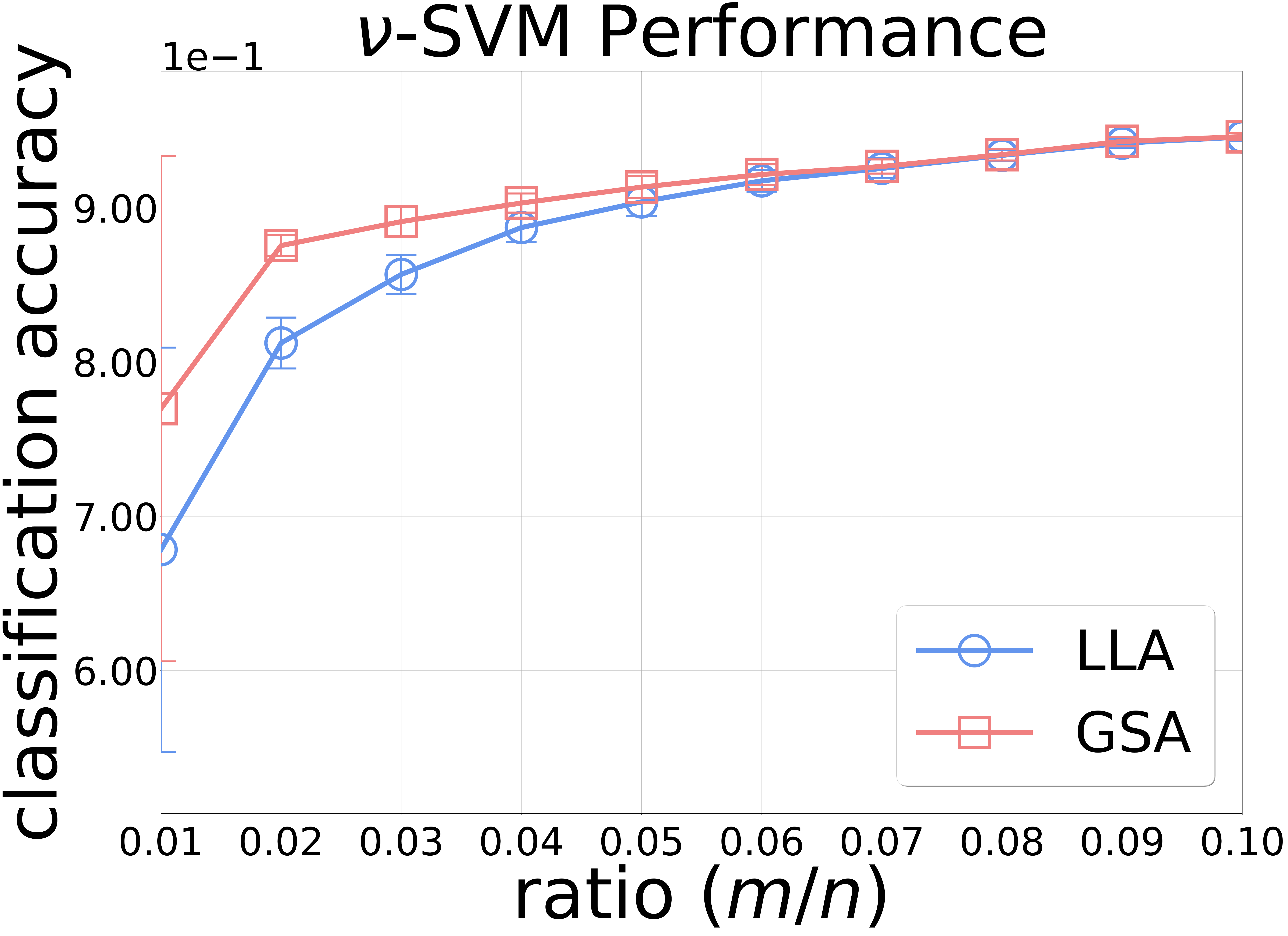}
			& \includegraphics[width=0.22\linewidth]{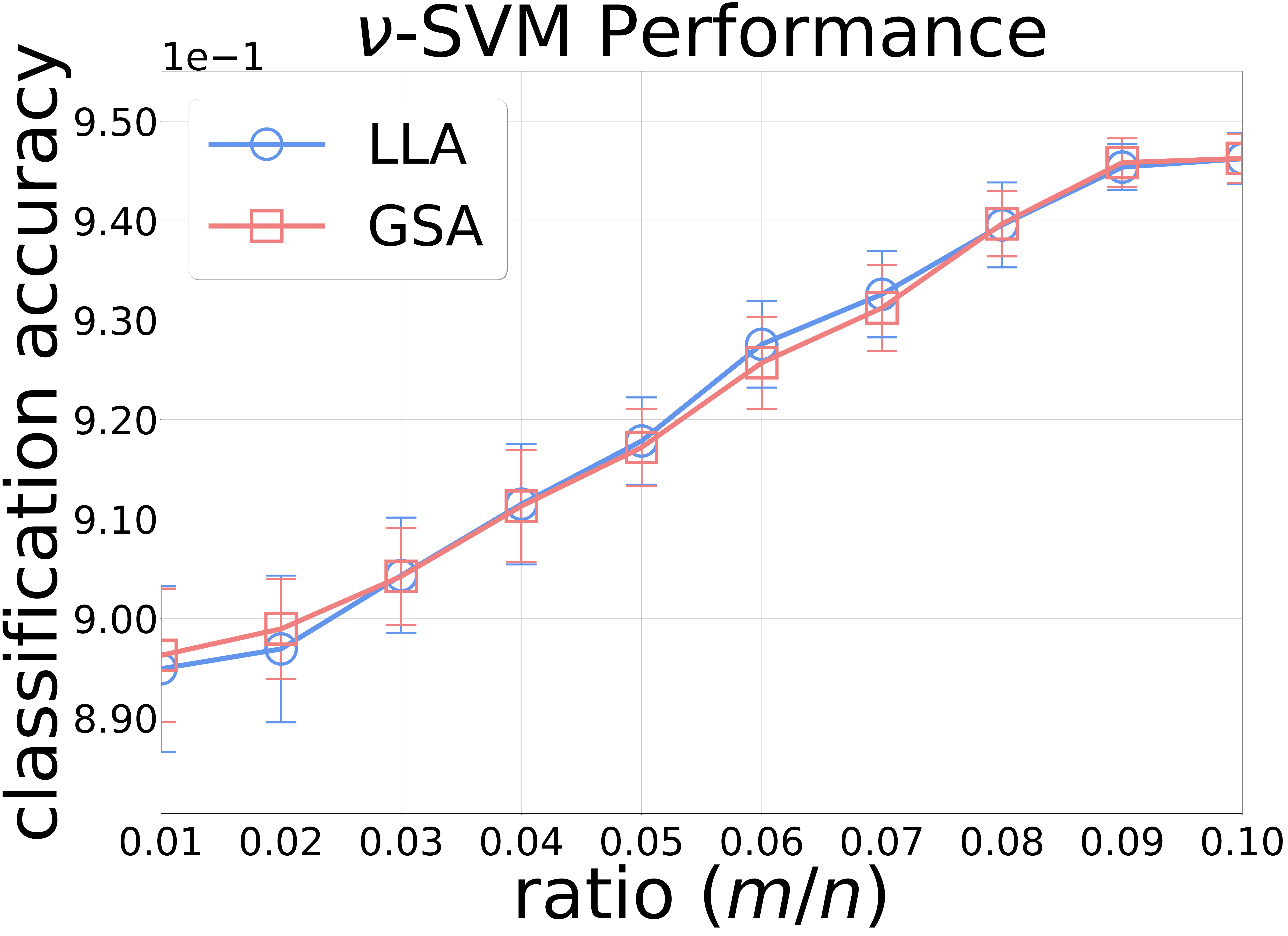}
			\\
			\multirow{2}{*}[0.5ex]{\rotatebox[origin=c]{90}{phishing}}
			& \includegraphics[width=0.22\linewidth]{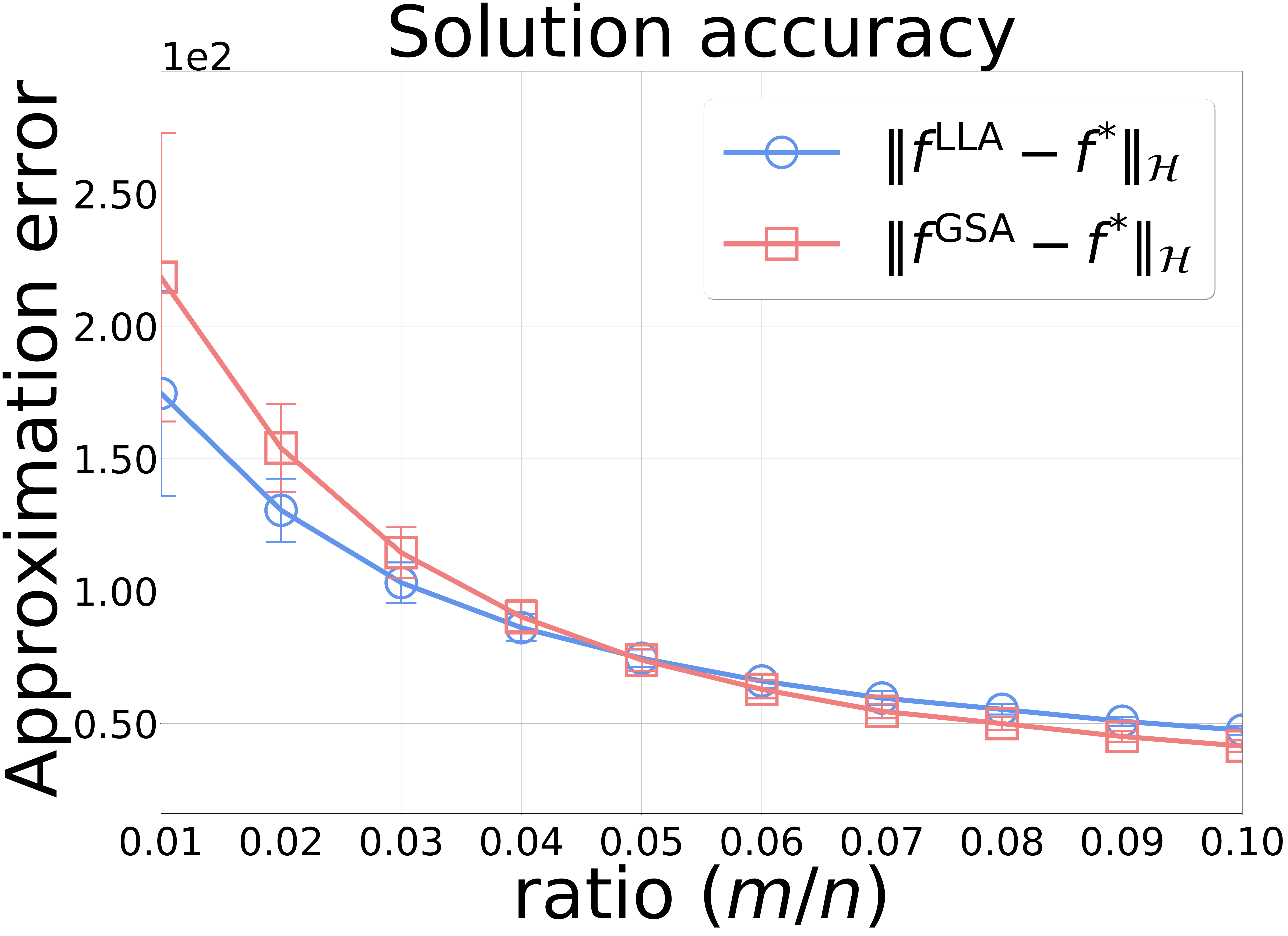}
			& \includegraphics[width=0.22\linewidth]{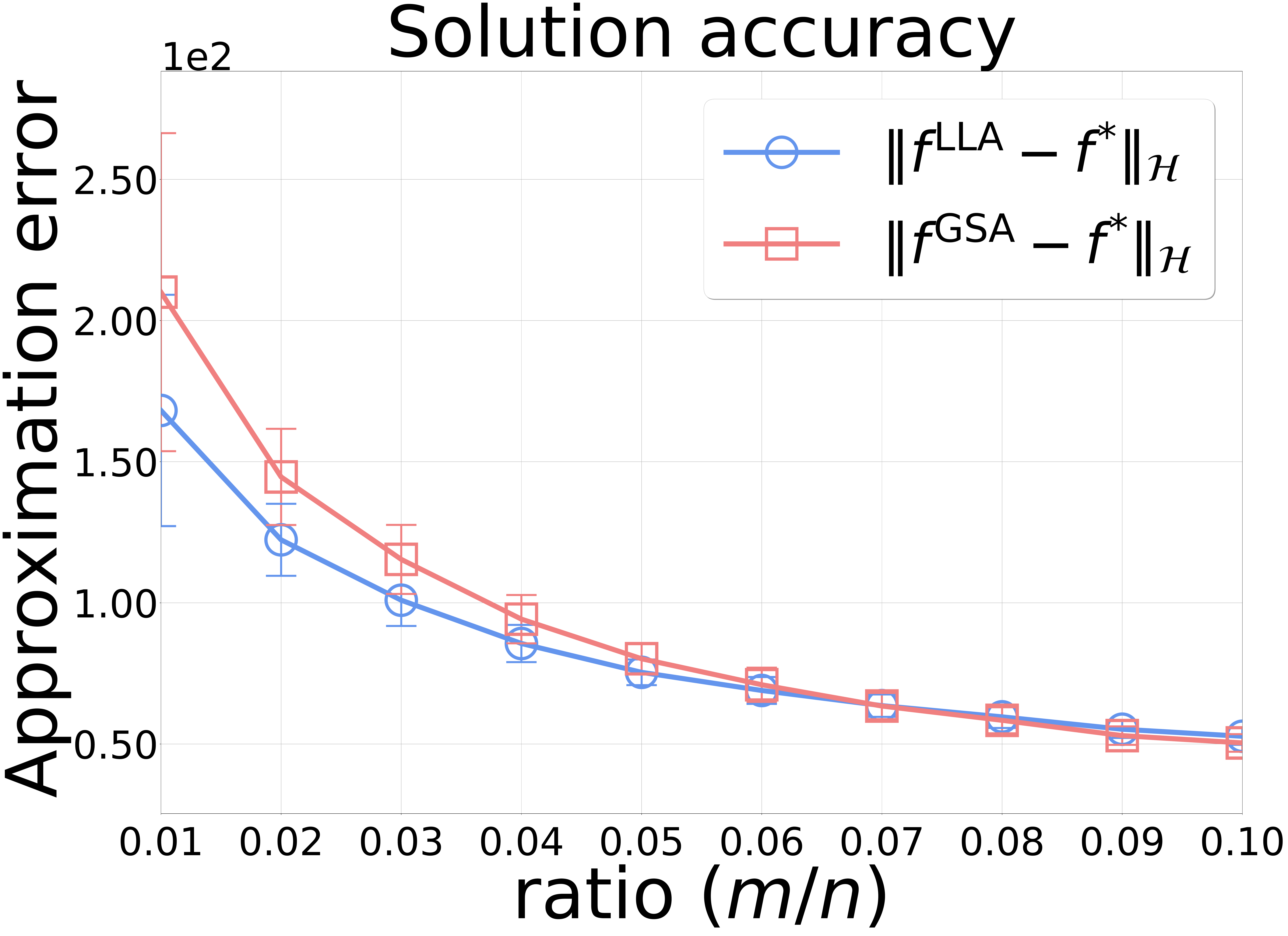}
			& \includegraphics[width=0.22\linewidth]{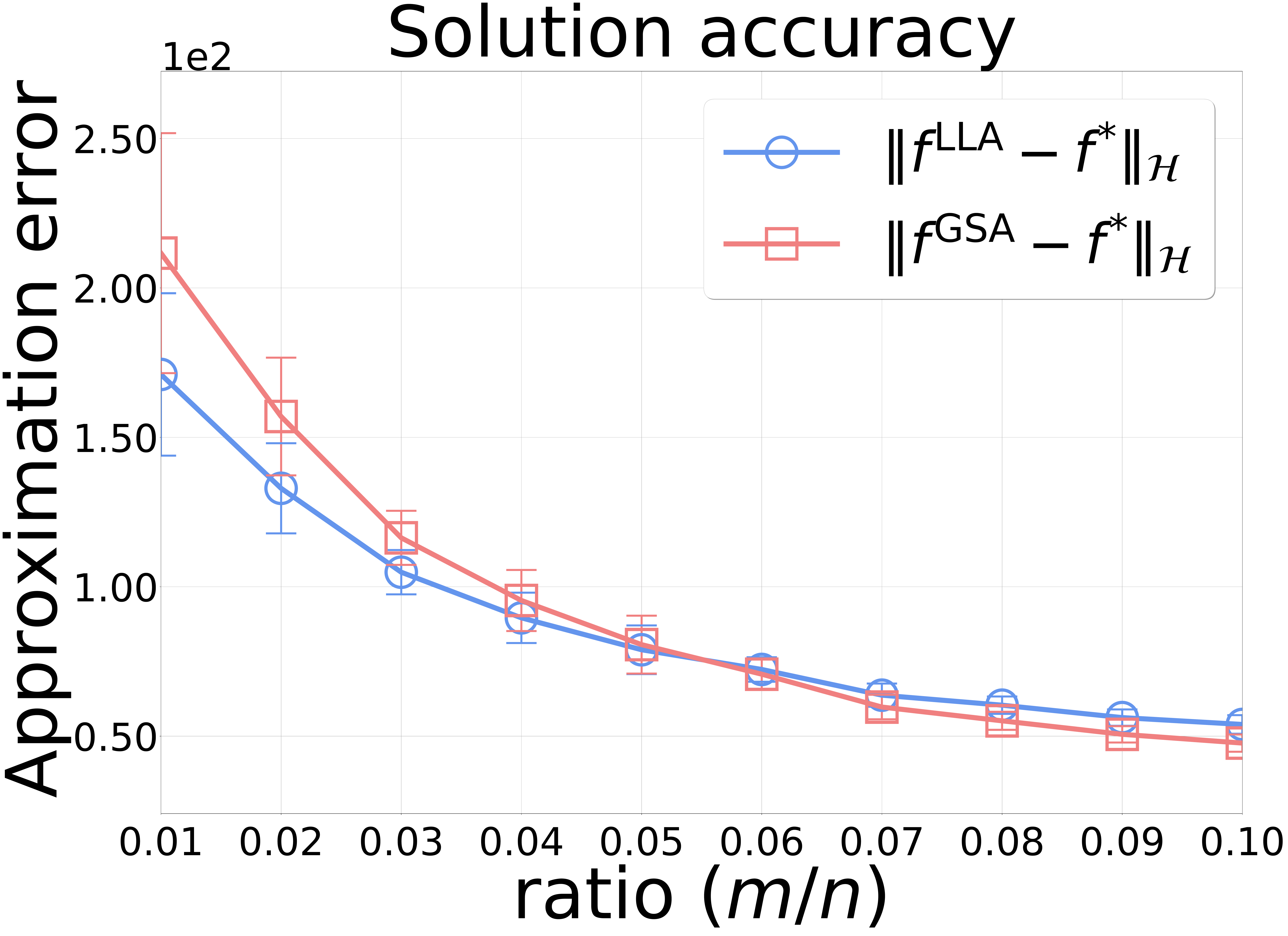}
			& \includegraphics[width=0.22\linewidth]{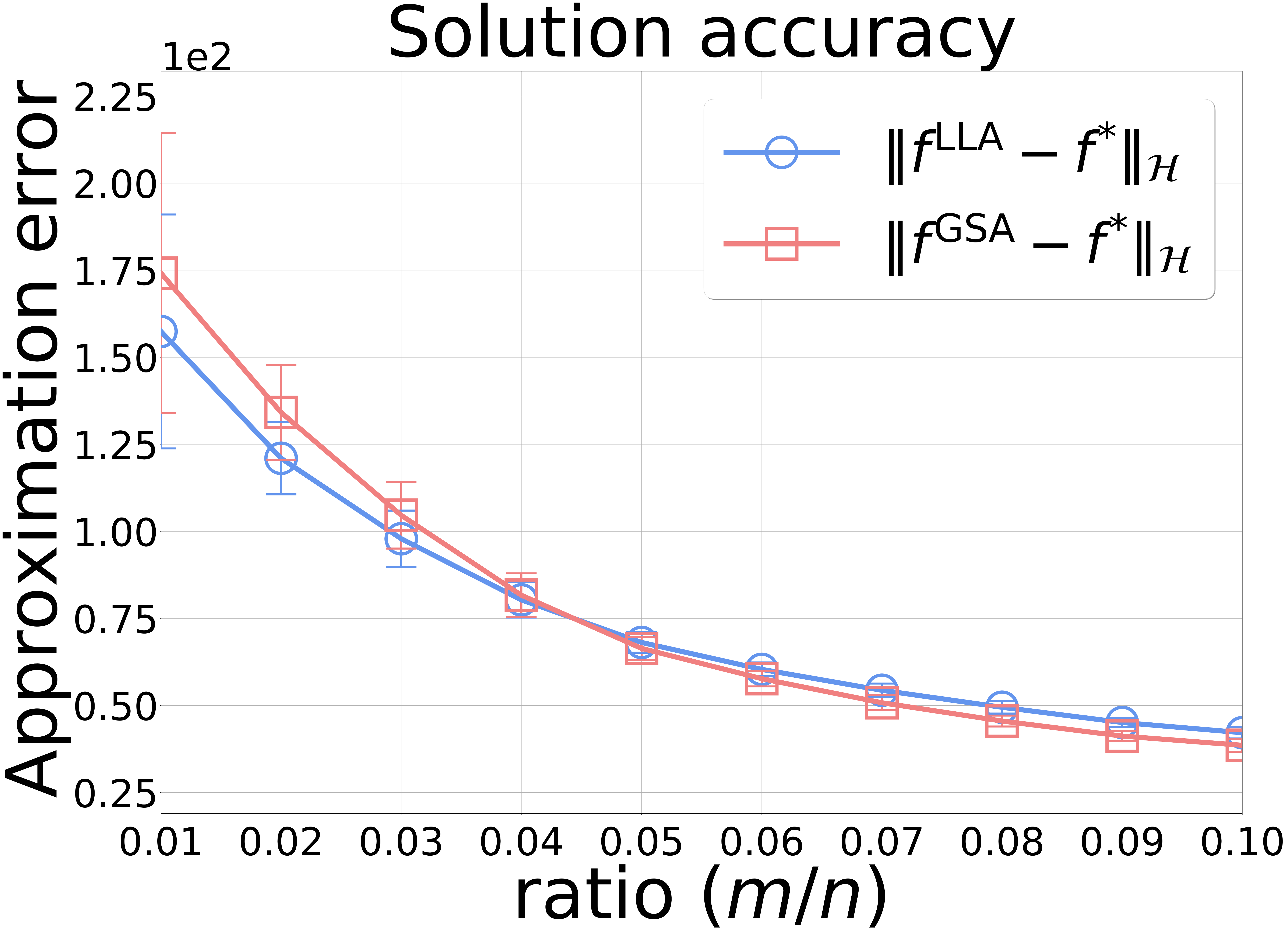} \\
			& \includegraphics[width=0.22\linewidth]{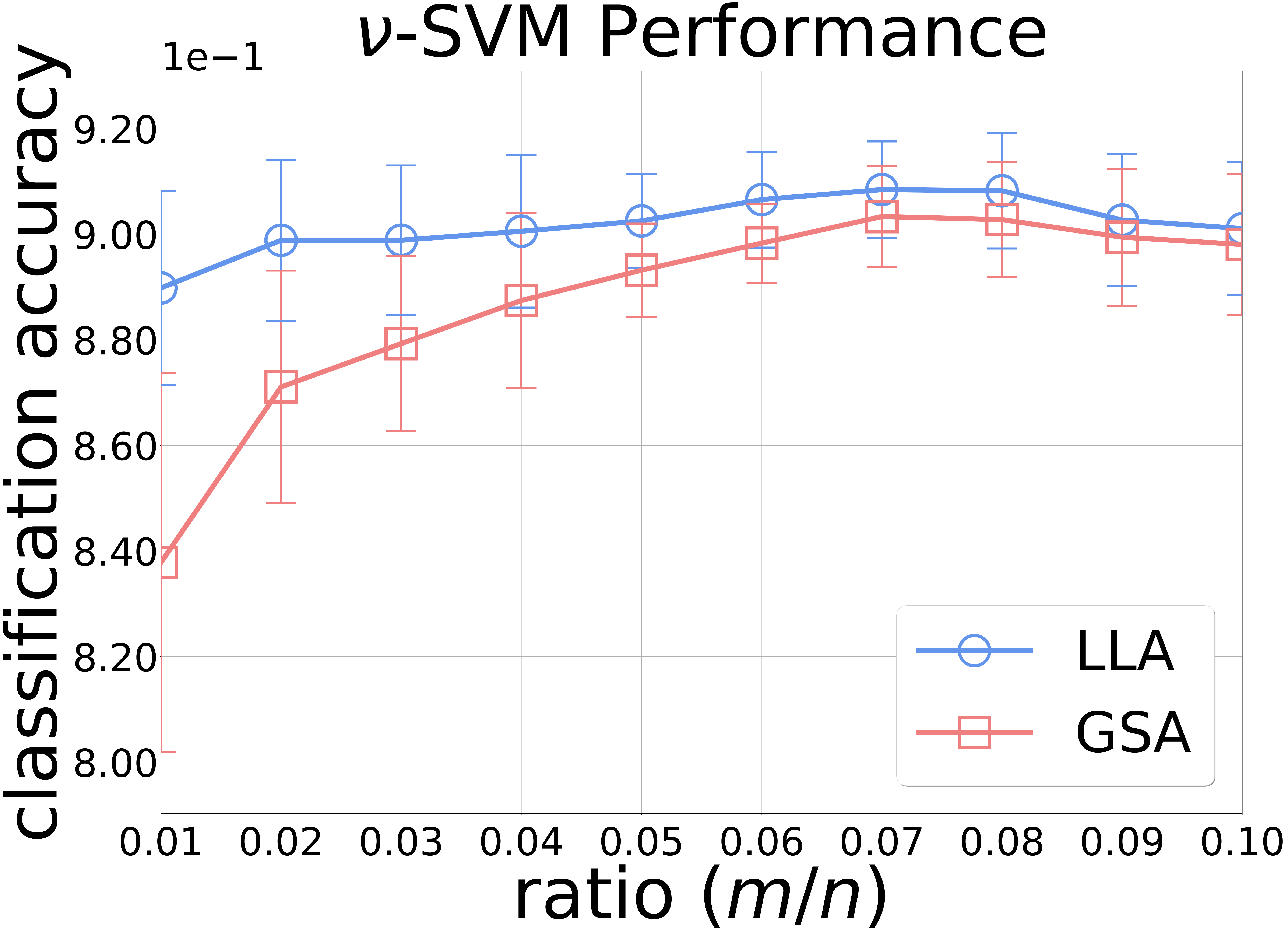}
			& \includegraphics[width=0.22\linewidth]{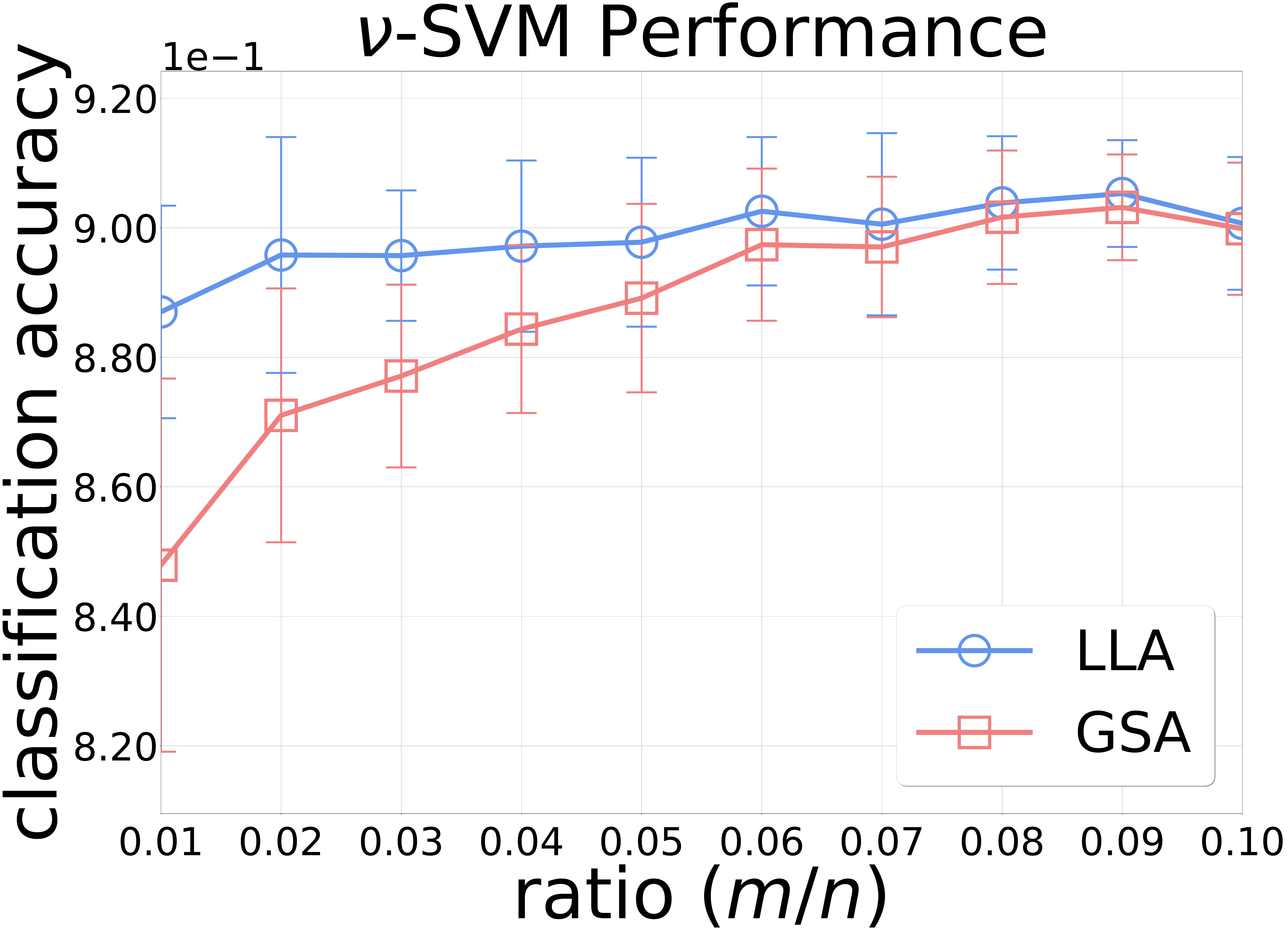}
			& \includegraphics[width=0.22\linewidth]{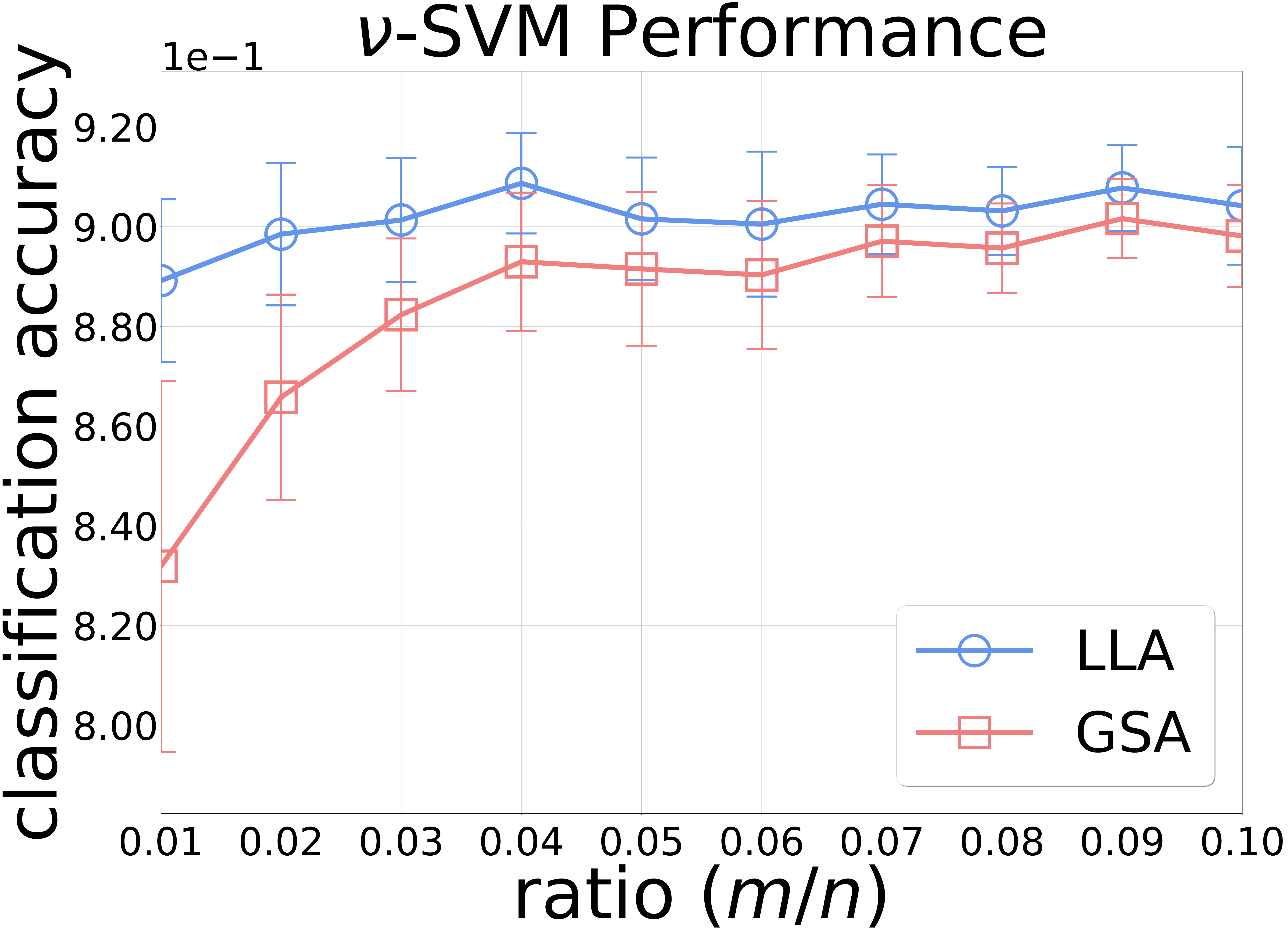}
			& \includegraphics[width=0.22\linewidth]{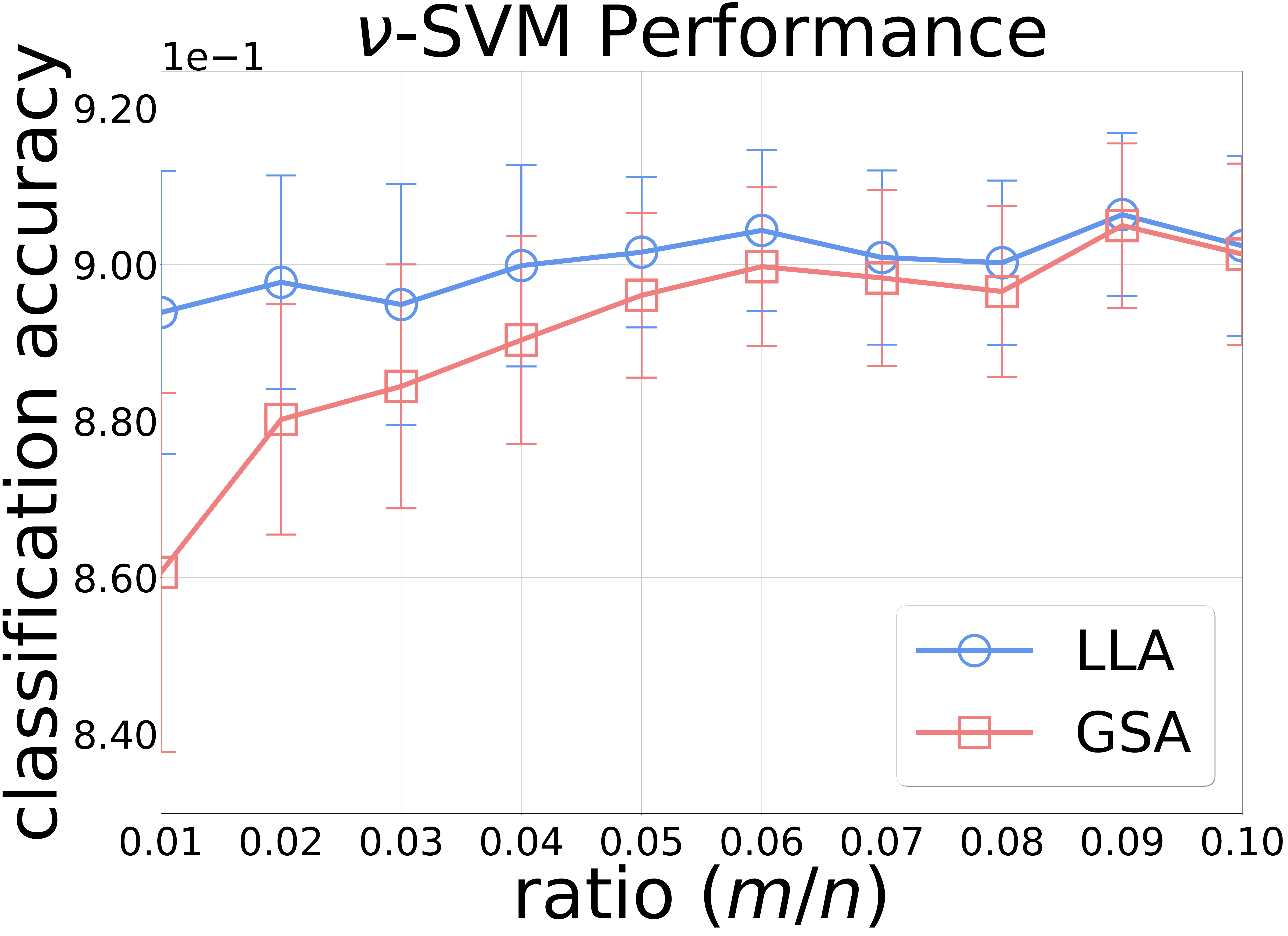}
		\end{tabular}		
		\caption{Comparison between Gram matrix substitution approach (GSA) and low-rank linearization approach (LLA) in terms of approximation error and classidfication accuracy (with $\nu$-SVM). Every two rows correspond to a specific dataset. Each column is related to a certain sampling strategy.}
		\label{fig:result2} 
	\end{figure*}

	\section{Experiments} 
	\label{sec:experiments}
	Bear in mind that the intent of our experiments is to verify that GSA can provide more accurate solutions than LLA (including NCR). 
	Being more accurate does not mean the performance will necessarily be better. 
	But it is expected to be true when the non-approximate optimal solution $ f^{*} $ to the kernel machine~\eqref{op:kernel models} performs well. 
	To the best of our knowledge, when the Nystr\"{o}m method is compared with other scalable techniques, LLA is always employed \cite{lan2019scaling,hsieh2014divide}. 
	Therefore, it is of interest to see the performances of GSA versus LLA.
	Specifically, we use the step provided by Proposition~\ref{prop:w2a} to efficiently optimize GSA, which is presented in Algorithm~\ref{alg:general}.
	
	
	Following previous studies \cite{zhang2010clustered,lan2019scaling,hsieh2014divide}, we focus on classification tasks, and the Gaussian kernel $ \exp(-\| \mathbf{x}-\mathbf{y} \|_{F}^{2} / 2\gamma^{2}) $ is used for all datasets. Four datasets from the LIBSVM archive (\url{https://www.csie.ntu.edu.tw/~cjlin/libsvmtools/datasets/}) are employed, which is listed in Table~\ref{tab:datasets}. 
	In our experiments, \texttt{NuSVC} from \texttt{sklearn} is employed for implementing $ \nu $-SVM. 
	Two metrics are used for evaluation, including 1) approximation error, and 2) classification accuracy.
	All experiments are conducted on a computer with 8 $ \times $ 2.40 GHz Intel(R) Core(TM) i7-4700HQ CPU and 16 GB of RAM. 
	
	\begin{table}[!t]
		\centering
		\caption{Summary of Datasets and Hyperparameters}
		\label{tab:datasets}
		\scriptsize
		\begin{tabular}{ccccccc} \toprule
			\textbf{Dataset} & \#training data & \#test data & \#feature & \#class & $ \gamma $ & $ \nu $ \\ \midrule
			\textbf{usps} & 7,291 & 2,007 & 256 & 10 & 10 & 0.1 \\
			\textbf{gisette} & 4,800 & 1,200 & 5,000 & 2 & 70 & 0.2 \\
			\textbf{phishing} & 8,388 & 2,097 & 9,947 & 4 & 10 & 82\\
			\textbf{dna} & 2,000 & 1,186 & 180 & 3 & 200 & 0.3 \\ \bottomrule
		\end{tabular}	
	\end{table}

	\subsection{Experiment Setting}
	
	Four sampling strategies are used here, including a) Gaussian sampling, b) uniform sampling, c) leverage score sampling, and d) k-means clustering sampling. 
	The details of these sampling strategies can be found in \cite{gittens2016revisiting,zhang2010clustered}. 
	Since the considered sampling strategies involve randomness, for each ratio ($ m/n $) of landmarks to training data, the averaged results with the standard deviations over the first $ 30 $ random seeds are reported.
	For each dataset, the ratio $ m/n $ gradually increases from $ 1\% $ to $ 10\% $ of the size of training data.

	For each dataset, we randomly split the whole dataset into a training set ($ 64\% $), a validation set ($ 16\% $) and a test set ($ 20\% $) if it is not previously divided. 
	We tune the hyperparameters of $ \nu $ and $ \gamma $ based on the training and validation sets.
	The considered ranges for $ \gamma $ and $ \nu $ are $ [10^{-3}, 10^{3}] $ and $ [0.1, 0.9] $, respectively.
	The eventually chosen hyperparameters of $ \nu $ and $ \gamma $ are listed in Table~\ref{tab:datasets}. 
	Besides, the maximum iteration for \texttt{NuSVC} is fixed as $ 1,000 $. 
	The embedded orthogonal basis $ \mathbf{B}_{\mathcal{H}}^{\mathrm{std}} $ when using standard Nystr\"{o}m method to form $ \widetilde{\mathbf{K}} $ is selected throughout our experiments. 
	$ \mathbf{B}_{\mathcal{H}}^{\mathrm{std}} \in \mathcal{H}^{s} $ can be calculated as follows: 1) by spectral decomposition, $ \langle \mathbf{C}_{\mathcal{H}}, \mathbf{C}_{\mathcal{H}} \rangle_{\mathcal{H}} = \mathbf{V}\boldsymbol\Sigma^{2}\mathbf{V}^{T} $, where all diagonal entries in $ \boldsymbol\Sigma $ are positive, and then 2) $ \mathbf{B}_{\mathcal{H}}^{\mathrm{std}} = \mathbf{C}_{\mathcal{H}}\mathbf{V}\boldsymbol\Sigma^{-1} $ is the desired orthogonal basis.
	Here, the dimension of the targeted subspace $ s $ is the dimension of $ \mathrm{span}(\mathbf{C}_{\mathcal{H}}) $.


	\subsection{Results}
	The experimental results on four datasets are reported in Figure~\ref{fig:result1} and Figure~\ref{fig:result2}. 
	From these two figures, we have several interesting observations. 
	\emph{First}, one can see that 
	all approximate solutions get more accurate as the ratio $ m/n $ increases. 
	The reason could be that, as the ratio $ m/n $ of landmarks to training data increases, the Gram matrix approximation error gets smaller; 
	and as indicated by the approximation error bound~\eqref{ieq:ksvm}, $ \| f^{\mathrm{LLA}}-f^{*} \|_{\mathcal{H}} $ is bounded by $ \| \mathbf{K}-\widetilde{\mathbf{K}} \|_{2}^{\frac{1}{4}} $ for KSVM. 
	Therefore, it is expected that the curve of $ \| f^{\mathrm{LLA}}-f^{*} \|_{\mathcal{H}} $ versus the ratio $ m/n $ will go down. 
	\emph{Second}, GSA is significantly more accurate than LLA on the usps dataset.
	On the gisette and phishing datasets, however, LLA can achieve more accurate approximate solutions than GSA when the ratio $ m/n $ is close to $ 1\% $. 
	Meanwhile, LLA is almost as accurate as GSA on the dna dataset. 
	\emph{In addition}, more accurate solutions (i.e., lower approximation errors) do not necessarily lead to better performances in terms of classification accuracy.
	By comparing the performances over the gisette and phishing datasets, we observe that even though GSA provides modestly more accurate solutions on both datasets when the ratio $ m/n $ is close to $ 10\% $, GSA performs better than LLA over gisette dataset but becomes slightly worse over phishing dataset.

	In a nutshell, even though LLA is commonly-used as an exemplar of using Nystr\"{o}m method to scale up kernel machines, we should not forget that GSA is potential to provide more accurate approximate solutions or perform better.

	\section{Conclusion}
	\label{conclusion}
	Motivated by the column inclusion property of Gram matrices, we propose a  subspace projection approach (SPA) as a cornerstone to study the relations among several well-studied approaches (i.e., LLA, NCR and GSA). 
	Specifically, the setting of SPA provides a way to reformulate LLA, which in turn reveals that NCR is a specific case of LLA. 
	When either the selected sampling strategy satisfies the equality $ \mathbf{C}_{\mathcal{H}} = \mathbf{X}_{\mathcal{H}}\mathbf{P} $ or the objective of kernel machine~\eqref{op:kernel models} meets the representer theorem, SPA serves as an alternative perspective for analyzing LLA. 
	The equivalence of LLA and SPA leads to three significant implications.
	\emph{First}, approximation errors for LLA in a general setting can be built up with a little effort.
	\emph{Second}, it reveals that the analytical forms of the approximate solutions computed through LLA and GSA only differ in one term.
	\emph{In addition}, GSA can be implemented as efficiently as LLA by sharing the same training procedure. 
	All the analytical results lead to the conjecture that GSA can provide better solutions than LLA, which is confirmed by our experiments with classification tasks. In a nutshell, the mechanism behind LLA is that it projects all data onto $ \mathrm{span}(\mathbf{B}_{\mathcal{H}}) $ before normal training and test.

	\appendices
	\section{Proofs of Lemma~\ref{lem:relation between K and B} and~\ref{lem:reconstruction error}}
	For Lemma~\ref{lem:relation between K and B}, let $ \widetilde{\mathbf{X}}_{\mathcal{H}} = \mathbf{B}_{\mathcal{H}}\langle \mathbf{B}_{\mathcal{H}}, \mathbf{X}_{\mathcal{H}} \rangle_{\mathcal{H}} $. Since
	\begin{align*}
		\langle \mathbf{X}_{\mathcal{H}}, \widetilde{\mathbf{X}}_{\mathcal{H}} \rangle_{\mathcal{H}} =& \langle \mathbf{X}_{\mathcal{H}}, \mathbf{B}_{\mathcal{H}}\langle \mathbf{B}_{\mathcal{H}}, \mathbf{X}_{\mathcal{H}} \rangle_{\mathcal{H}} \rangle_{\mathcal{H}} \\
		=& \langle \mathbf{X}_{\mathcal{H}}, \mathbf{B}_{\mathcal{H}} \rangle_{\mathcal{H}}\langle \mathbf{B}_{\mathcal{H}}, \mathbf{X}_{\mathcal{H}} \rangle_{\mathcal{H}}\\
		=& \langle \mathbf{X}_{\mathcal{H}}, \mathbf{B}_{\mathcal{H}} \rangle_{\mathcal{H}} \langle \mathbf{B}_{\mathcal{H}}, \mathbf{B}_{\mathcal{H}} \rangle_{\mathcal{H}} \langle \mathbf{B}_{\mathcal{H}}, \mathbf{X}_{\mathcal{H}} \rangle_{\mathcal{H}}\\
		=& \langle \mathbf{B}_{\mathcal{H}}\langle \mathbf{B}_{\mathcal{H}}, \mathbf{X}_{\mathcal{H}} \rangle_{\mathcal{H}}, \mathbf{B}_{\mathcal{H}}\langle \mathbf{B}_{\mathcal{H}}, \mathbf{X}_{\mathcal{H}} \rangle_{\mathcal{H}} \rangle_{\mathcal{H}} \\
		=& \langle \widetilde{\mathbf{X}}_{\mathcal{H}}, \widetilde{\mathbf{X}}_{\mathcal{H}} \rangle_{\mathcal{H}} ,
	\end{align*}
	there is
	\begin{gather*}
		\begin{aligned}
			&\langle \mathbf{X}_{\mathcal{H}}-\widetilde{\mathbf{X}}_{\mathcal{H}}, \mathbf{X}_{\mathcal{H}}-\widetilde{\mathbf{X}}_{\mathcal{H}} \rangle_{\mathcal{H}} \\
			=& \langle \mathbf{X}_{\mathcal{H}}, \mathbf{X}_{\mathcal{H}} \rangle_{\mathcal{H}} + \langle \widetilde{\mathbf{X}}_{\mathcal{H}}, \widetilde{\mathbf{X}}_{\mathcal{H}} \rangle_{\mathcal{H}} - 2 \langle \mathbf{X}_{\mathcal{H}}, \widetilde{\mathbf{X}}_{\mathcal{H}} \rangle_{\mathcal{H}}\\
			=&\mathbf{K} + \widetilde{\mathbf{K}} - 2\widetilde{\mathbf{K}} \\
			=& \mathbf{K}-\widetilde{\mathbf{K}} .
		\end{aligned}
	\end{gather*}
	This equality implies
	\begin{gather*}
		\begin{aligned}
			\left\lVert \mathbf{X}_{\mathcal{H}}-\widetilde{\mathbf{X}}_{\mathcal{H}} \right\rVert_{\mathcal{H}S}^{2}
			= \mathrm{trace}(\mathbf{K} - \widetilde{\mathbf{K}})
			= \left\lVert \mathbf{K} - \widetilde{\mathbf{K}} \right\rVert_{*} . 
		\end{aligned}
	\end{gather*}
	Here, the last equality holds due to that $ \mathbf{K} \succeq \widetilde{\mathbf{K}} $. In other words, $ \mathbf{K}-\widetilde{\mathbf{K}} $ is positive semi-definite.
	
	For the other inequality, it is sufficient to prove that $ \| \mathbf{A}_{\mathcal{H}} \|_{op}^{2} = \| \langle \mathbf{A}_{\mathcal{H}}, \mathbf{A}_{\mathcal{H}} \rangle_{\mathcal{H}} \|_{2} $ for each $ \mathbf{A}_{\mathcal{H}} \in \mathcal{H}^{k} $. For convenience, let $ \mathbf{M} = \langle \mathbf{A}_{\mathcal{H}}, \mathbf{A}_{\mathcal{H}} \rangle_{\mathcal{H}} $. 
	By spectral decomposition, $ \mathbf{M} = \mathbf{B}^{T}\mathbf{B} $.
	According to the definitions of operator norm and spectral norm, there exist $ \mathbf{x},\mathbf{y} \in \mathbb{R}^{k} $ such that $ \| \mathbf{x} \|_{F} = \| \mathbf{y} \|_{F} = 1, \| \mathbf{A}_{\mathcal{H}}\mathbf{x} \|_{\mathcal{H}} = \| \mathbf{A}_{\mathcal{H}} \|_{op} $ and $ \| \mathbf{By} \|_{F} = \| \mathbf{B} \|_{2} $.
	Therefore,
	\begin{align*}
		\| \mathbf{A}_{\mathcal{H}} \|_{op}^{2} & = \| \mathbf{A}_{\mathcal{H}}\mathbf{x} \|_{\mathcal{H}}^{2} = \mathbf{x}^{T}\mathbf{Mx} \leq \| \mathbf{M} \|_{2}\| \mathbf{x} \|_{F}^{2} = \| \mathbf{M} \|_{2} .
	\end{align*}
	On the other hand,
	\begin{align*}
		\| \mathbf{M} \|_{2} &= \| \mathbf{B}^{T}\mathbf{B} \|_{2} = \|  \mathbf{B} \|_{2}^{2} = \| \mathbf{By} \|_{F}^{2} = \mathbf{y}^{T}\mathbf{M}\mathbf{y} \\
		&= \| \mathbf{A}_{\mathcal{H}}\mathbf{y} \|_{\mathcal{H}}^{2} \leq \| \mathbf{A}_{\mathcal{H}} \|_{op}^{2} .
	\end{align*}
	Therefore, we reach the equality $ \| \mathbf{A}_{\mathcal{H}} \|_{op}^{2} = \| \mathbf{M} \|_{2} $. As previously shown, there is $ \langle \mathbf{X}_{\mathcal{H}}-\widetilde{\mathbf{X}}_{\mathcal{H}}, \mathbf{X}_{\mathcal{H}}-\widetilde{\mathbf{X}}_{\mathcal{H}} \rangle_{\mathcal{H}} = \mathbf{K}-\widetilde{\mathbf{K}} $, which leads to
	\begin{gather*}
		\left\lVert \mathbf{X}_{\mathcal{H}}-\widetilde{\mathbf{X}}_{\mathcal{H}} \right\rVert_{op}^{2} = \left\lVert \mathbf{K}-\widetilde{\mathbf{K}} \right\rVert_{2} .
	\end{gather*} \hfill $ \Box $
	
	For Lemma~\ref{lem:reconstruction error},
	let $ \mathbf{p}_{\mathcal{H}} = \Phi(\mathbf{p}), \mathbf{q}_{\mathcal{H}} = \Phi(\mathbf{q}), \mathbf{p}'_{\mathcal{H}} = \mathbf{p}_{\mathcal{H}}-\widetilde{\mathbf{p}}_{\mathcal{H}} $ and $ \mathbf{q}'_{\mathcal{H}} = \mathbf{q}_{\mathcal{H}}-\widetilde{\mathbf{q}}_{\mathcal{H}} $. 
	Note that
	\begin{gather*}
		\langle \widetilde{\mathbf{p}}_{\mathcal{H}}, \mathbf{q}'_{\mathcal{H}} \rangle_{\mathcal{H}} = \langle \widetilde{\mathbf{p}}_{\mathcal{H}}, \mathbf{q}_{\mathcal{H}} \rangle_{\mathcal{H}} - \langle \widetilde{\mathbf{p}}_{\mathcal{H}}, \widetilde{\mathbf{q}}_{\mathcal{H}} \rangle_{\mathcal{H}}  = 0
	\end{gather*}
	since
	\begin{align*}
		\langle \widetilde{\mathbf{p}}_{\mathcal{H}}, \mathbf{q}_{\mathcal{H}} \rangle_{\mathcal{H}} &= \langle \mathbf{B}_{\mathcal{H}}\langle \mathbf{B}_{\mathcal{H}}, \mathbf{p}_{\mathcal{H}} \rangle_{\mathcal{H}}, \mathbf{q}_{\mathcal{H}} \rangle_{\mathcal{H}} \\
		&= \langle \mathbf{q}_{\mathcal{H}}, \mathbf{B}_{\mathcal{H}} \rangle_{\mathcal{H}}\langle \mathbf{B}_{\mathcal{H}}, \mathbf{q}_{\mathcal{H}} \rangle_{\mathcal{H}}\\
		&= \langle \mathbf{q}_{\mathcal{H}}, \mathbf{B}_{\mathcal{H}} \rangle_{\mathcal{H}}\langle \mathbf{B}_{\mathcal{H}}, \mathbf{B}_{\mathcal{H}} \rangle_{\mathcal{H}}\langle \mathbf{B}_{\mathcal{H}}, \mathbf{q}_{\mathcal{H}} \rangle_{\mathcal{H}}\\
		&= \langle \mathbf{B}_{\mathcal{H}}\langle \mathbf{B}_{\mathcal{H}}, \mathbf{p}_{\mathcal{H}} \rangle_{\mathcal{H}}, \mathbf{B}_{\mathcal{H}}\langle \mathbf{B}_{\mathcal{H}}, \mathbf{q}_{\mathcal{H}} \rangle_{\mathcal{H}} \rangle_{\mathcal{H}} \\
		&= \langle \widetilde{\mathbf{p}}_{\mathcal{H}}, \widetilde{\mathbf{q}}_{\mathcal{H}} \rangle_{\mathcal{H}}
	\end{align*}	
	Likewise, $ \langle \widetilde{\mathbf{q}}_{\mathcal{H}}, \mathbf{p}_{\mathcal{H}}' \rangle_{\mathcal{H}} = 0 $.  
	Then,
	\begin{align*}
		&|\langle \widetilde{\mathbf{p}}_{\mathcal{H}}, \widetilde{\mathbf{q}}_{\mathcal{H}} \rangle_{\mathcal{H}} - \langle \mathbf{p}_{\mathcal{H}}, \mathbf{q}_{\mathcal{H}} \rangle_{\mathcal{H}}| \\
		=& | \langle \mathbf{p}_{\mathcal{H}}', \mathbf{q}_{\mathcal{H}}' \rangle_{\mathcal{H}}+\langle \widetilde{\mathbf{p}}_{\mathcal{H}}, \mathbf{q}_{\mathcal{H}}' \rangle_{\mathcal{H}} + \langle \mathbf{p}_{\mathcal{H}}',\widetilde{\mathbf{q}}_{\mathcal{H}} \rangle_{\mathcal{H}} | \\		
		=& |\langle \mathbf{p}'_{\mathcal{H}}, \mathbf{q}'_{\mathcal{H}} \rangle|
		\leq \| \mathbf{p}'_{\mathcal{H}} \|_{\mathcal{H}}\| \mathbf{q}'_{\mathcal{H}} \|_{\mathcal{H}} .
	\end{align*}
	Therefore, the lemma holds since $ \| \mathbf{p}'_{\mathcal{H}} \|_{\mathcal{H}} $ = 0 if and only if $ \mathbf{p}_{\mathcal{H}} \in \mathrm{span}(\mathbf{B}_{\mathcal{H}}) $. \hfill $ \Box $
	
	\section{Proof of Corollary~\ref{cor:NCR2LLA}} \label{app:NCR2LLA}
	To be self-contained, we provide all tools that are needed for our proof here.

	\begin{lemma} \label{lem:T}
		Let $ \mathbf{A}_{\mathcal{H}} \in \mathcal{H}^{p} $ be given, and $ \rho $ denotes the dimension of $ \mathrm{span(\mathbf{A}_{\mathcal{H}})} $. Then, there is an isomorphism $ T_{A} $, which is also an isometry, between $ \mathrm{span}(\mathbf{A}_{\mathcal{H}}) $ and $ \mathbb{R}^{\rho} $.
	\end{lemma}
	
	\emph{Proof.} If the vectors in $ \mathbf{A}_{\mathcal{H}} $ are not linearly dependent, then gradually remove one unnecessary vector each time from $ \mathbf{A}_{\mathcal{H}} $ until the remaining vectors (denoted by $ \hat{\mathbf{A}}_{\mathcal{H}} \in \mathcal{H}^{\rho} $) are linearly independent. Then, $ \mathrm{span}(\mathbf{A}_{\mathcal{H}}) = \mathrm{span}(\hat{\mathbf{A}}_{\mathcal{H}}) $. 
	The linear independence of $ \hat{\mathbf{A}}_{\mathcal{H}} $ guarantees that $ \hat{\mathbf{K}} = \langle \hat{\mathbf{A}}_{\mathcal{H}}, \hat{\mathbf{A}}_{\mathcal{H}} \rangle_{\mathcal{H}} \in \mathbb{R}^{\rho\times\rho} $ is invertible, meaning that it is positive definite. Therefore, by Cholesky decomposition, $ \hat{\mathbf{K}} = \mathbf{B}^{T}\mathbf{B} $ where $ \mathbf{B} \in \mathbb{R}^{\rho\times\rho} $ is invertible. Define a linear map $ T_{A} : \mathrm{span}(\mathbf{A}_{\mathcal{H}}) \mapsto \mathbb{R}^{\rho} $ by setting
	\begin{equation}
		\begin{gathered}
			T_{A}(\sum_{i=1}^{\rho}\alpha_{i}\hat{\mathbf{a}}^{i}_{\mathcal{H}}) = \sum_{i=1}^{\rho}\alpha_{i}\mathbf{b}_{i} .
		\end{gathered}
	\end{equation}
	One can check that $ T_{A} $ is indeed a linear bijection, and thus an isomorphism. Moreover, $ \langle \mathbf{x}_{\mathcal{H}}, \mathbf{y}_{\mathcal{H}} \rangle_{\mathcal{H}} = T_{A}(\mathbf{x}_{\mathcal{H}})^{T}T_{A}(\mathbf{y}_{\mathcal{H}}) $ for all $ \mathbf{x}_{\mathcal{H}},\mathbf{y}_{\mathcal{H}} \in \mathrm{span}(\mathbf{A}_{\mathcal{H}}) $ indicates that $ T_{A} $ is an isometry. \hfill $ \Box $
	
	\begin{lemma}[Compact SVD on $ \mathcal{H} $] \label{lem:svd}
		Let $ \mathbf{A}_{\mathcal{H}} \in \mathcal{H}^{k} $ be given. By spectral decomposition, $ \langle \mathbf{A}_{\mathcal{H}}, \mathbf{A}_{\mathcal{H}} \rangle_{\mathcal{H}} = \mathbf{R}\boldsymbol\Lambda^{2}\mathbf{R}^{T} $ where $ \mathbf{R}^{T}\mathbf{R} = \mathbf{I} $ and all diagonal entries in $ \boldsymbol\Lambda $ are all positive. Let $ \mathbf{Y}_{\mathcal{H}} = \mathbf{A}_{\mathcal{H}}\mathbf{R}\boldsymbol\Lambda^{-1} $, and thus $ \langle \mathbf{Y}_{\mathcal{H}}, \mathbf{Y}_{\mathcal{H}} \rangle_{\mathcal{H}} = \mathbf{I} $. Then, there is $ \mathbf{A}_{\mathcal{H}} = \mathbf{Y}_{\mathcal{H}}\boldsymbol\Lambda\mathbf{R}^{T} $.
	\end{lemma}
	\emph{Proof.} Firstly, construct $ T_{A} $ according to Lemma~\ref{lem:T}. Then, $ T_{A}(\mathbf{A}_{\mathcal{H}})^{T}T_{A}(\mathbf{A}_{\mathcal{H}}) = \langle \mathbf{A}_{\mathcal{H}}, \mathbf{A}_{\mathcal{H}} \rangle_{\mathcal{H}} = \mathbf{R}\boldsymbol\Lambda^{2}\mathbf{R}^{T} $. Let $ \mathbf{Y} = T_{A}(\mathbf{A}_{\mathcal{H}})\mathbf{R}\boldsymbol\Lambda^{-1} $, then $ T_{A}(\mathbf{A}_{\mathcal{H}}) = \mathbf{Y}\boldsymbol\Lambda\mathbf{R}^{T} $ is a compact SVD. 
	Besides, $ T_{A}^{-1}(\mathbf{Y}) = \mathbf{A}_{\mathcal{H}}\mathbf{R}\boldsymbol\Lambda^{-1} = \mathbf{Y}_{\mathcal{H}} $ due to the linearity of $ T_{A}^{-1} $. Note that
	\begin{gather*}
		\mathbf{A}_{\mathcal{H}} = T_{A}^{-1}(\mathbf{Y}\boldsymbol\Lambda\mathbf{R}^{T}) = \mathbf{Y}_{\mathcal{H}}\boldsymbol\Lambda\mathbf{R}^{T},
	\end{gather*}
	the proof is completed. \hfill $ \Box $

	\emph{Proof of corollary~\ref{cor:NCR2LLA}}.
	Note that $ \mathbf{K}_{nm}^{T} = \mathbf{K}_{mn} = \langle \mathbf{C}_{\mathcal{H}}, \mathbf{X}_{\mathcal{H}} \rangle_{\mathcal{H}}, \mathbf{K}_{mm} = \langle \mathbf{C}_{\mathcal{H}}, \mathbf{C}_{\mathcal{H}} \rangle_{\mathcal{H}} $ and $ \mathbf{B}_{\mathcal{H}}^{\mathrm{std}} = \mathbf{C}_{\mathcal{H}}\mathbf{V}\boldsymbol\Sigma^{-1} $ where $ \mathbf{V} $ and $ \boldsymbol\Sigma $ come from the spectral decomposition $ \mathbf{K}_{mm} = \mathbf{V}\boldsymbol\Sigma^{2}\mathbf{V}^{T} $. Here, all diagonal entries in $ \boldsymbol\Sigma $ are positive. According to Lemma~\ref{lem:svd}, there is $ \mathbf{C}_{\mathcal{H}} = \mathbf{B}_{\mathcal{H}}^{\mathrm{std}}\boldsymbol\Sigma\mathbf{V}^{T} $.
	Our goal is to show the equality
	\begin{gather*}
		\mathbf{C}_{\mathcal{H}} (\mathbf{K}_{mn}\mathbf{K}_{nm}+\lambda_{0}\mathbf{K}_{mm})^{\dagger}\mathbf{K}_{mn}\mathbf{y} = \mathbf{B}_{\mathcal{H}}^{\mathrm{std}}\hat{\mathbf{w}}^{\mathrm{KRR}}	
	\end{gather*}
	where
	\begin{gather*}
		\hat{\mathbf{w}}^{\mathrm{KRR}} = (\mathbf{G}\mathbf{G}^{T}+\lambda_{0}\mathbf{I})^{-1}\mathbf{G}\mathbf{y} , \\
		\mathbf{G} = (\mathbf{V}\boldsymbol\Sigma^{-1})^{T}\mathbf{K}_{mn}.
	\end{gather*}
	
	Since
	\begin{gather*}
		\mathbf{B}_{\mathcal{H}}^{\mathrm{std}}\hat{\mathbf{w}}^{\mathrm{KRR}} = \mathbf{C}_{\mathcal{H}}\mathbf{H}(\mathbf{H}^{T}\mathbf{K}_{mn}\mathbf{K}_{nm}\mathbf{H}+\lambda_{0}\mathbf{I})^{-1}\mathbf{H}^{T}\mathbf{K}_{mn}\mathbf{y}
	\end{gather*}
	where $ \mathbf{H} = \mathbf{V}\boldsymbol\Sigma^{-1} $,
	it is sufficient to prove that
	\begin{gather*}
		(\mathbf{K}_{mn}\mathbf{K}_{nm}+\lambda_{0}\mathbf{K}_{mm})^{\dagger} = \mathbf{H}(\mathbf{H}^{T}\mathbf{K}_{mn}\mathbf{K}_{nm}\mathbf{H}+\lambda_{0}\mathbf{I})^{-1}\mathbf{H}^{T} .
	\end{gather*}
	
	Let $ \mathbf{M} = \boldsymbol\Sigma\langle \mathbf{B}_{\mathcal{H}}^{\mathrm{std}}, \mathbf{X}_{\mathcal{H}} \rangle_{\mathcal{H}}\langle \mathbf{X}_{\mathcal{H}}, \mathbf{B}_{\mathcal{H}}^{\mathrm{std}} \rangle_{\mathcal{H}}\boldsymbol\Sigma $, then there is
	\begin{gather*}
		\mathbf{K}_{mn}\mathbf{K}_{nm} = \mathbf{V}\mathbf{M}\mathbf{V}^{T} .
	\end{gather*}
	
	Since
	\begin{gather*}
		\begin{aligned}
			&\mathbf{H}^{T}\mathbf{K}_{mn}\mathbf{K}_{nm}\mathbf{H} + \lambda_{0}\mathbf{I}\\
			=& \mathbf{H}^{T}(\mathbf{K}_{mn}\mathbf{K}_{nm}+\lambda_{0}\mathbf{K}_{mm})\mathbf{H}\\
			=& \mathbf{H}^{T}\mathbf{V}(\mathbf{M}+\lambda_{0}\boldsymbol\Sigma^{2})\mathbf{V}^{T}\mathbf{H} \\
			=& \boldsymbol\Sigma^{-1}(\mathbf{M}+\lambda_{0}\boldsymbol\Sigma^{2})\boldsymbol\Sigma^{-1} ,
		\end{aligned}
	\end{gather*}
	there is 
	\begin{gather*}
		\begin{aligned}
			&\mathbf{H}(\mathbf{H}^{T}\mathbf{K}_{mn}\mathbf{K}_{nm}\mathbf{H}+\lambda_{0}\mathbf{I})^{-1}\mathbf{H}^{T} \\
			=& \mathbf{H}\boldsymbol\Sigma(\mathbf{M}+\lambda_{0}\boldsymbol\Sigma^{2})^{-1}\boldsymbol\Sigma\mathbf{H}^{T} \\
			=& \mathbf{V}(\mathbf{M}+\lambda_{0}\boldsymbol\Sigma^{2})^{-1}\mathbf{V}^{T} \\
			=& (\mathbf{V}(\mathbf{M}+\lambda_{0}\boldsymbol\Sigma^{2})\mathbf{V}^{T})^{\dagger} \\
			=& (\mathbf{K}_{mn}\mathbf{K}_{nm}+\lambda_{0}\mathbf{K}_{mm})^{\dagger} .
		\end{aligned}
	\end{gather*} \hfill $ \Box $
	
	\section{Proof of Proposition~\ref{prop:approximation error}}
	\begin{gather*}
		\begin{aligned}
			& \| f^{\mathrm{LLA}} - f^{*} \|_{\mathcal{H}}^{2}\\
			=& \langle \widetilde{\mathbf{X}}_{\mathcal{H}}\hat{\boldsymbol\alpha}-\mathbf{X}_{\mathcal{H}}\widetilde{\boldsymbol\alpha}, \widetilde{\mathbf{X}}_{\mathcal{H}}\hat{\boldsymbol\alpha}-\mathbf{X}_{\mathcal{H}}\widetilde{\boldsymbol\alpha} \rangle_{\mathcal{H}}\\
			=& \hat{\boldsymbol\alpha}^{T}\widetilde{\mathbf{K}}\hat{\boldsymbol\alpha} + \widetilde{\boldsymbol\alpha}^{T}\mathbf{K}\widetilde{\boldsymbol\alpha} - 2\hat{\boldsymbol\alpha}^{T}\widetilde{\mathbf{K}}\widetilde{\boldsymbol\alpha}\\
			=& (\hat{\boldsymbol\alpha}-\widetilde{\boldsymbol\alpha})^{T}\widetilde{\mathbf{K}}(\hat{\boldsymbol\alpha}-\widetilde{\boldsymbol\alpha}) + \widetilde{\boldsymbol\alpha}^{T}(\mathbf{K}-\widetilde{\mathbf{K}})\widetilde{\boldsymbol\alpha}\\
			\leq& \| \widetilde{\mathbf{K}} \|_{2}\| \hat{\boldsymbol\alpha}-\widetilde{\boldsymbol\alpha} \|_{F}^{2} + \| \mathbf{K}-\widetilde{\mathbf{K}} \|_{2}\| \widetilde{\boldsymbol\alpha} \|_{F}^{2} \\
			\leq& \| \mathbf{K} \|_{2}\| \hat{\boldsymbol\alpha}-\widetilde{\boldsymbol\alpha} \|_{F}^{2} + \| \mathbf{K}-\widetilde{\mathbf{K}} \|_{2}\| \widetilde{\boldsymbol\alpha} \|_{F}^{2}  .
		\end{aligned}
	\end{gather*}
	The last inequality is due to $ \widetilde{\mathbf{K}} \preceq \mathbf{K} \implies \| \widetilde{\mathbf{K}} \|_{2} \leq \| \mathbf{K} \|_{2} $. 
	Taking square root on both sides leads to
	\begin{align*}
		&\| f^{\mathrm{LLA}} - f^{*} \|_{\mathcal{H}} \\
		\leq& \left( \| \mathbf{K} \|_{2}\| \hat{\boldsymbol\alpha}-\widetilde{\boldsymbol\alpha} \|_{F}^{2} + \| \mathbf{K}-\widetilde{\mathbf{K}} \|_{2}\| \widetilde{\boldsymbol\alpha} \|_{F}^{2} \right)^{\frac{1}{2}}\\
		\leq& \left( \| \mathbf{K} \|_{2}\| \hat{\boldsymbol\alpha}-\widetilde{\boldsymbol\alpha} \|_{F}^{2} \right)^{\frac{1}{2}} + \left( \| \mathbf{K}-\widetilde{\mathbf{K}} \|_{2}\| \widetilde{\boldsymbol\alpha} \|_{F}^{2} \right)^{\frac{1}{2}} \\
		=& \| \mathbf{K} \|_{2}^{\frac{1}{2}}\| \hat{\boldsymbol\alpha}-\widetilde{\boldsymbol\alpha} \|_{F} + \| \mathbf{K}-\widetilde{\mathbf{K}} \|_{2}^{\frac{1}{2}}\| \widetilde{\boldsymbol\alpha} \|_{F} .
	\end{align*} \hfill $ \Box $

	\section{Proof of Proposition~\ref{prop:w2a}}
	\begin{lemma}[Theorem 4.11 in~\cite{rudin2006real}] \label{lem:rn}
		Suppose a subspace $ \mathcal{S} \subseteq \mathcal{H} $ is closed, and let $ \mathcal{S}^{\perp} $ be its orthogonal complement $ \{ \mathbf{x}_{\mathcal{H}} \in \mathcal{H} \mid \langle \mathbf{x}_{\mathcal{H}}, \mathbf{s}_{\mathcal{H}} \rangle_{\mathcal{H}} = 0 \text{ for all } \mathbf{s}_{\mathcal{H}} \in \mathcal{S} \} $. Then, there are two linear maps $ r:\mathcal{H}\mapsto \mathcal{S} $ and $ n:\mathcal{H}\mapsto \mathcal{S}^{\perp} $ such that for each $ \mathbf{x}_{\mathcal{H}} \in \mathcal{H} $, $ \mathbf{x}_{\mathcal{H}} = r(\mathbf{x}_{\mathcal{H}}) + n(\mathbf{x}_{\mathcal{H}}) $.
	\end{lemma}

	It will be trivially true if $ \mathbf{B}_{\mathcal{H}}\hat{\mathbf{w}} \in \mathrm{span}(\widetilde{\mathbf{X}}_{\mathcal{H}}) $. Because it implies that the equality~\eqref{eq:xa=bw} holds, and thus Corollary~\ref{cor: LLA2GSA} works.
	Therefore, we suppose that $ \mathbf{B}_{\mathcal{H}}\hat{\mathbf{w}} \not\in \mathrm{span}(\widetilde{\mathbf{X}}_{\mathcal{H}}) $. 
	
	According to Lemma~\ref{lem:T}, $ \mathrm{span}(\widetilde{\mathbf{X}}_{\mathcal{H}}) $ is bijectively isometric to another complete space. This implies $ \mathrm{span}(\widetilde{\mathbf{X}}_{\mathcal{H}}) $ is closed, and thus by Lemma~\ref{lem:rn}, $ \hat{\mathbf{w}} $ can be decomposed as
	\begin{gather*}
		\hat{\mathbf{w}} = \mathbf{w}_{r} + \mathbf{w}_{n}
	\end{gather*}
	such that $ \mathbf{B}_{\mathcal{H}}\mathbf{w}_{r} \in \mathrm{span}(\widetilde{\mathbf{X}}_{\mathcal{H}}), \mathbf{B}_{\mathcal{H}}\mathbf{w}_{n} \in \mathrm{span}(\widetilde{\mathbf{X}}_{\mathcal{H}})^{\perp} $, which means $ \mathbf{B}_{\mathcal{H}}\mathbf{w}_{r} \perp \mathbf{B}_{\mathcal{H}}\mathbf{w}_{n} $. 
	
	The weak equivalence between SPA and LLA indicates that $ \mathbf{B}_{\mathcal{H}}\mathbf{w}_{r} $ is optimal to both SPA and LLA. By Corollary~\ref{cor: LLA2GSA}, $ \mathbf{G}^{\dagger}\mathbf{w}_{r} $ is an optimal solution to the problem~\eqref{op:approximate_solvable models}.
	Note that $ \mathbf{G} = \langle \mathbf{B}_{\mathcal{H}}, \mathbf{X}_{\mathcal{H}} \rangle_{\mathcal{H}} $.
	Suffice it to prove that
	\begin{gather*}
		\mathbf{G}^{\dagger}\hat{\mathbf{w}} = \mathbf{G}^{\dagger}\mathbf{w}_{r} ,
	\end{gather*}
	which is equivalent to say that $ \mathbf{G}^{\dagger}\mathbf{w}_{n} = \mathbf{0} $. 
	This equality will be true if $ \mathbf{w}_{n} \in \mathrm{span}(\widetilde{\mathbf{G}})^{\perp} $. 
	Since $ \widetilde{\mathbf{X}}_{\mathcal{H}} = \mathbf{B}_{\mathcal{H}}\langle \mathbf{B}_{\mathcal{H}}, \mathbf{X}_{\mathcal{H}} \rangle_{\mathcal{H}} = \mathbf{B}_{\mathcal{H}}\mathbf{G} $ and $ \mathbf{B}_{\mathcal{H}}\mathbf{w}_{n} \in \mathrm{span}(\widetilde{\mathbf{X}}_{\mathcal{H}})^{\perp} $, we have
	\begin{gather*}
		\mathbf{0} = \langle \widetilde{\mathbf{X}}_{\mathcal{H}}, \mathbf{B}_{\mathcal{H}}\mathbf{w}_{n} \rangle_{\mathcal{H}} = \mathbf{G}^{T}\langle \mathbf{B}_{\mathcal{H}},\mathbf{B}_{\mathcal{H}} \rangle_{\mathcal{H}}\mathbf{w}_{n} = \mathbf{G}^{T}\mathbf{w}_{n} .
	\end{gather*}
	Therefore, $ \mathbf{w}_{n} \in \mathrm{null}(\mathbf{G}^{T}) = \mathrm{span}(\mathbf{G})^{\perp} $ where $ \mathrm{null}(\mathbf{G}^{T}) =  \{ \mathbf{x}\mid \mathbf{G}^{T}\mathbf{x} = \mathbf{0} \} $. \hfill $ \Box $

	\section{Proof of Proposition~\ref{prop:C=XP}}
	Note that we already have $ \mathrm{span}(\widetilde{\mathbf{X}}_{\mathcal{H}}) \subseteq \mathrm{span}(\mathbf{B}_{\mathcal{H}}) $ since $ \widetilde{\mathbf{X}}_{\mathcal{H}} = \mathbf{B}_{\mathcal{H}}\langle \mathbf{B}_{\mathcal{H}}, \mathbf{X}_{\mathcal{H}} \rangle_{\mathcal{H}} $.
	Suffice it to prove $ \mathrm{span}(\mathbf{B}_{\mathcal{H}}) \subseteq\mathrm{span}(\widetilde{\mathbf{X}}_{\mathcal{H}}) $.
	Since $ \mathbf{B}_{\mathcal{H}} = \mathbf{C}_{\mathcal{H}}\mathbf{A} = \mathbf{X}_{\mathcal{H}}\mathbf{PA} $, we have
	\begin{gather*}
		\mathbf{\widetilde{\mathbf{X}}}_{\mathcal{H}} = \mathbf{B}_{\mathcal{H}}\mathbf{A}^{T}\mathbf{P}^{T}\mathbf{K} .
	\end{gather*}
	The desired result $ \mathrm{span}(\mathbf{B}_{\mathcal{H}}) \subseteq \mathrm{span}(\widetilde{\mathbf{X}}_{\mathcal{H}}) $ will follow if $ \mathbf{A}^{T}\mathbf{P}^{T}\mathbf{K} \in \mathbb{R}^{s\times n} $ is full-row rank, which is guaranteed by the fact that
	\begin{gather*}
		\mathbf{A}^{T}\mathbf{P}^{T}\mathbf{KP}\mathbf{A} = \langle \mathbf{B}_{\mathcal{H}}, \mathbf{B}_{\mathcal{H}} \rangle_{\mathcal{H}} = \mathbf{I} .
	\end{gather*}
	\hfill $ \Box $
	
	\section{Proof of Proposition~\ref{prop:reprst to equiv}}
	Suppose that the objective of kernel machine~\eqref{op:kernel models} satisfies the strict representer theorem, and let $ \hat{\mathbf{w}} $ be an optimal solution to the problem~\eqref{op:solvableLLA}, we will prove the proposition by contradiction. 
	Let $ f = \mathbf{B}_{\mathcal{H}}\hat{\mathbf{w}} $, which is optimal to LLA~\eqref{op:recastLLA}. 
	Similar to discussion in the proof of Proposition~\ref{prop:w2a}, $ f $ can be uniquely decomposed as $ f = f_{r}+f_{n} $ such that $ f_{r} \in \mathrm{span}(\widetilde{\mathbf{X}}_{\mathcal{H}}) $ and $ f_{n} \in \mathrm{span}(\widetilde{\mathbf{X}}_{\mathcal{H}})^{\perp} $. 
	If $ f $ is not optimal to the SPA~\eqref{op:projected kernel models}, there are two cases:
	
	1) If $ f \not\in \mathrm{span}(\widetilde{\mathbf{X}}_{\mathcal{H}}) $, then $ f_{n} \not= \mathbf{0} $. The fact that the kernel machine~\eqref{op:kernel models} satisfies the strict representer theorem indicates that $ \hat{\mathcal{R}}(f_{r}) < \hat{\mathcal{R}}(f) $, a contradiction.
	
	2) If $ f \in \mathrm{span}(\widetilde{\mathbf{X}}_{\mathcal{H}}) $, there will be another solution $ \hat{f} \in \mathrm{span}(\widetilde{\mathbf{X}}_{\mathcal{H}}) \subseteq \mathrm{span}(\mathbf{B}_{\mathcal{H}}) $ such that $ \hat{\mathcal{R}}(\hat{f}) < \hat{\mathcal{R}}(f) $, a contradiction.
	
	Therefore, $ f = \mathbf{B}_{\mathcal{H}}\hat{\mathbf{w}} $ must be optimal to SPA~\eqref{op:projected kernel models}.
	
	Suppose the objective of the kernel machine~\eqref{op:kernel models} satisfies the weak representer theorem. Let $ f \in \mathrm{span}(\mathbf{B}_{\mathcal{H}}) $ be an optimal solution to LLA, and denote its projection onto $ \mathrm{span}(\widetilde{\mathbf{X}}_{\mathcal{H}}) $ by $ \overline{f} $.
	The optimality of $ f $ implies $ \hat{\mathcal{R}}(f) = \hat{\mathcal{R}}(\overline{f}) $, and thus $ \overline{f} $ is optimal to both SPA~\eqref{op:projected kernel models} and LLA~\eqref{op:recastLLA}. \hfill $ \Box $

	\section*{Acknowledgment}
	W.~Li and D.~Zhang were supported in part by the National Key R\&D Program of China (Nos. 2018YFC2001600 and 2018YFC2001602), the National Natural Science Foundation of China (Nos.~61732006, 61876082, and 61861130366), and the Royal Society-Academy of Medical Sciences Newton Advanced Fellowship (No.~NAF$\backslash$R1$\backslash$180371).

	\ifCLASSOPTIONcaptionsoff
	\newpage
	\fi

	\footnotesize
	\bibliographystyle{IEEEtranN}
	\bibliography{mybibfile}

	\if false
	\begin{IEEEbiography}{Michael Shell}
		Biography text here.
	\end{IEEEbiography}
	
	\begin{IEEEbiographynophoto}{John Doe}
		Biography text here.
	\end{IEEEbiographynophoto}
	
	\begin{IEEEbiographynophoto}{Jane Doe}
		Biography text here.
	\end{IEEEbiographynophoto}
	\fi
	
\end{document}